\documentclass[twoside,11pt]{article}

\usepackage{jmlr2e}
\usepackage{graphicx}
\usepackage{adjustbox}
\usepackage{amsmath,amssymb,mathtools}
\usepackage{tikz}
\usepackage{physics}
\usepackage{float}
\usepackage{subfigure}
\usepackage[noend]{algorithmic}
\usepackage{algorithm}
\usepackage{booktabs}
\usepackage{multirow}
\usepackage{dsfont}
\usepackage{wrapfig}
\usepackage{capt-of}
\usepackage{enumitem}
\usepackage{color, colortbl}

\usetikzlibrary{shapes,arrows}
\tikzstyle{arrow}=[draw]
\tikzstyle{block} = [rectangle, draw, text width=6em, text centered, rounded corners, minimum height=4em]
\tikzstyle{line} = [draw, -latex']

\newcommand*\circled[1]{\tikz[baseline=(char.base)]{
            \node[shape=circle,draw,inner sep=0.7pt] (char) {#1};}}
\newtheorem{claim}{Claim}
\definecolor{Gray}{gray}{0.9}

\ShortHeadings{A Near-Optimal Gradient Flow for Learning Neural Energy-Based Models}{Wu and Wei}
\firstpageno{1}

\begin{document}

\title{A Near-Optimal Gradient Flow for Learning Neural Energy-Based Models}

\author{\name Yang Wu \email wuyang36@mail2.sysu.edu.cn \\
       \name PengXu Wei \email weipx3@mail.sysu.edu.cn\\
       \name Liang Lin \email  linliang@ieee.org \\
       \addr School of Computer Science and Engineering\\
       Sun Yet-Sen University\\
       Guangzhou, GuangDong Province, China 
       }
\editor{}
\maketitle

\begin{abstract}
In this paper, we propose a novel numerical scheme to optimize the gradient flows for learning energy-based models (EBMs). 
From a perspective of physical simulation, we redefine the problem of approximating the gradient flow utilizing optimal transport (i.e. Wasserstein) metric. In EBMs, the learning process of stepwise sampling and estimating data distribution performs the functional gradient of minimizing the global relative entropy between the current and target real distribution, which can be treated as dynamic particles moving from disorder to target manifold. 
Previous learning schemes mainly minimize the entropy concerning the consecutive time KL divergence in each learning step. However, they are prone to being stuck in the local KL divergence by projecting non-smooth information within smooth manifold, which is against the optimal transport principle.
To solve this problem, we derive a second-order Wasserstein gradient flow of the global relative entropy from Fokker-Planck equation. Compared with existing schemes, Wasserstein gradient flow is a smoother and near-optimal numerical scheme to approximate real data densities. We also derive this near-proximal scheme and provide its numerical computation equations. 
Our extensive experiments demonstrate the practical superiority and potentials of our proposed scheme on fitting complex distributions and generating high-quality, high-dimensional data with neural EBMs.
  
\end{abstract}


\section{Introduction}
\label{sec:introduction}
Energy-Based Models~(EBMs) as remarkable likelihood models present an appealing property that they amalgamate both discriminative and generative learning into one energy function~\citep{lecun2006tutorial}. Generative learning of EBMs has gained an increasing attention recently, and is used to approximate a large class of real-world data distributions, such as handwritten characters, natural images, and other multi-modal data. EBMs with a {\em neural energy function} $\Phi_{\theta}(\mathbf{x})$ extract the features of multiple scales, and build well-generalizing hierarchical data representations~\citep{xie2016theory}. The parameter estimation of this energy function follows the maximum likelihood estimation~(MLE), and is cast as a learning problem of minimizing the Kullback–Leibler~(KL) divergence between the observed data distribution $p_{\text{data}}$ and the 
estimated distribution $p_{\theta}$. To approximate $p_{\text{data}}$, some Markov chain Monte Carlo~(MCMC) methods---the Gibbs sampler~\citep{casella1992explaining}, Hamiltonian Monte Carlo~\citep{neal2011mcmc} and the stochastic gradient Langevin dynamics~(SGLD)~\citep{welling2011bayesian}---are often used to accomplish the sampling process $\mathbf{x}\sim p_{\theta}$. Those methods can be efficiently implemented by gradient computation via back-propagation through the neural energy function $\Phi_{\theta}(\mathbf{x})$. Among those samplers, SGLD is the most popular choice, and with MLE, the learning process can be described by a cross-updating formula of learning $\theta$ and sampling $\mathbf{x}$ that iteratively minimizes the KL divergence from $p_{\text{data}}$ to $p_{\theta}$. 

In our work, we first investigate the microscopic view of the cross-updating with SGLD and MLE in EBMs, specifically, from a perspective of particle evolution~\cite{cai2019frame}. We consider all the samples $\mathbf{x}$ as a set of {\em dynamic particles} (referred as particles) and the learning process step-wisely transports those particles to match the target distribution $p_{\text{data}}$. The {\em gradient flow} $\mathcal{F}$ performed by MLE is driven by consecutive time KL divergence $D_{\text{KL}}(\rho_{t+1}\|\rho_t)$; namely, we first initiate the particles, and at each time step, the particles are prone to moving to a new location, where the empirical distribution $\rho_{t+1}$ minimizes the KL divergence to the current location empirical distribution $\rho_t$. We refer such a discrete time gradient flow as the {\em KL discrete flow}. This perspective and original SGLD with MLE will be described in Sec.~\ref{sec:setup}. We will next prove the theoretical equivalence between the original and the new perspective in the Sec.~\ref{sec:klflow}. Based on the particle evolution viewpoint, it reveals that the original learning procedure suffers from following problems:
\begin{itemize}
  \item \textbf{KL vanishing.} It is noticeable  that KL divergence is asymmetric, according to its formula. In the case where the empirical distribution $\rho_{t+1}$ is close to zero while $\rho_t$ is non-zero, the effect of $\rho_t$ is disregarded (note that this is just a mini-partial step of minimizing $D_{\text{KL}}(p_{\text{data}}\|p_{\theta})$). Thus, minimizing the local time KL divergence $D_{\text{KL}}(\rho_{t+1}\|\rho_t)$ often causes particles to vanish to zero-points.
  \item \textbf{Undetermined force.} The dynamics of moving particles are driven by SGLD, which relies on randomness for approximation. In most cases, random diffusion is inefficient since it requires large numbers of particles to form a proper approximation.
  \item \textbf{Non-smooth and unbounded transportation.} Minimizing the local time KL divergence $D_{\text{KL}}(\rho_{t+1}\|\rho_t)$ computes the information projection within the data space, of which the low dimensional manifold supports the moving particles. The non-smooth orthogonal projection is unbounded, and thus will typically over-estimate the support of $\rho_{t+1}$ and provide useless gradients for training.
\end{itemize}

We aim to tackle above issues by improving the learning and sampling process of EBMs with the {\em optimal transport} (OT) methods. In the last few decades, OT refers to the study of transportation resource allocation problems and has emerged as a foundational tool to analyze diverse problems at the interface among variational analysis, partial differential equations and probabilistic models~\citep{Villani2003}. Currently, OT enjoys applications in very diverse fields such as computer vision~\citep{Salimans2018,Rubner2000}, reinforcement learning~\citep{scetbon2021equitable}, astrophysics~\citep{frisch2002monge}, statistics~\citep{Papadakis2014}, economy~\citep{Chiappori2009} and etc. The major application interest of OT cost is the symmetric property, since OT provides a smooth and optimal way of transporting particles. Particularly, the minimum OT cost of all particles is the Wasserstein distance. Following the initial work of Jordan, Kinderlehrer and Otto~\citep{jordan1998variational}, it is possible to perform gradient flow of $D_{\text{KL}}(p_{\text{data}}\|p_{\theta})$ w.r.t discrete time in Wasserstein distance, which gives the name of this method, the JKO scheme. We refer such discrete time gradient flow as {\em JKO discrete flow}. Then in Sec.~\ref{sec:ourmethod}, we show that under the ideal circumstances, solving KL discrete flow ends up with the same stationary solution with solving JKO discrete flow, even these two processes are independent. Motivated by this, we interpret KL discrete flow with JKO discrete flow and find that in fact it is to deal with an OT problem. The nice properties of JKO discrete flow largely avoid the aforementioned problems of KL discrete flow. 

Although the Wasserstein distance is a useful tool to investigate transport equations, for its application to dynamics of {\em measures}\footnote{A measure is a generalization of the concepts of length, area, and volume. In this paper we treat probability distribution as one special case of measure.}, it still has a limitation of being well-defined only if the estimated two measures have the same mass. But a more flexible formulation of the Wasserstein distance (namely a {\em dynamical} JKO discrete flow), proposed by Benamou and Brenier~\citep{benamou2000computational} solves this problem. We will in detail prove the equivalence of the dynamical and normal JKO discrete flow in Sec.~\ref{sec:jkoflow} and discuss the disadvantages and advantages over those two forms in our settings. The dynamical JKO discrete flow in principle requires the empirical measure curve $\rho_t$ to be in a set of Lipschitz curves, which is equivalent to restricting the neural energy function $\Phi_\theta$ to be Lipschitz continuous. This idea is quite similar to adding the spectral normalization to restrict the Lipschitz constant of the discriminator function in GAN models~\citep{miyato2018spectral}. 
Note that $L$-Lipschitz $\Phi_\theta$ yields a bounded velocity as well as the valid gradients for learning neural parameters and transporting particles. Based on this, a new dynamic equation characterizing the movements of particles can be directly derived from the dynamical JKO discrete flow, and we show it is an approximation of the stochastic gradient descent~(SGD). We elaborate this deduction in Sec.~\ref{sec:methodderive}. The proposed numerical scheme can be carried out under deep learning frameworks to well handle the high dimensional data situations. To this end, our contributions can be summarized as follows:
\begin{itemize}
    \item \textbf{Asymptotic analysis.} We first revisit the learning of EBMs in KL discrete flow from a particle evolution perspective. In this progress, concretely, we show the MLE learning of $\theta$ and SGLD sampling of $\mathbf{x}$ together implicitly solve the saddle-point of KL discrete flow. We also empirically analyze and verify the disadvantages of KL discrete flow. 
    
    \item \textbf{Gradient flow optimization.} We show the equivalence between KL discrete flow and JKO discrete flow. To further optimize the JKO scheme, we introduce an Eulerian point of view to revisit the OT problem that the dynamical JKO discrete flow has the bounded velocity so that optimal trajectories are smooth and Lipschitz continuous. To this end, we derive a novel near-optimal numerical approximation scheme for leaning EBMs, which corresponds to the dynamical JKO discrete flow. We refer to this scheme for EBMs throughout this work as OT-EBM. Note that theoretically, our method can be applied to other EBM-like generative models. 
    
    \item \textbf{Deterministic dynamic particles.} Our derived particle evolution equation corresponding to the dynamical JKO discrete flow can be related to the SGD solver, which is an unbiased estimator. SGD estimator not only provides a deterministic force for the dynamic particles to make the evolution of the particles predictable, but also presents future research directions for the particle evolution equation to be further improved by other gradient descent variants such as Adam~\citep{kingma2014adam}.
\end{itemize}

%
\section{Problem Setup}
\label{sec:setup}
\textbf{Energy Function.} Let $p_{\theta}$ be the probability density for $\mathbf{x}\in\mathbb{R}^d$ modeled by an energy function $\Phi_{\theta}(\mathbf{x})$, namely 
\begin{equation}
    \label{eq:ebm}
    p_{\theta}(\mathbf{x}) = \frac{1}{Z_{\theta}}e^{-\Phi_{\theta}(\mathbf{x})}, 
\end{equation}
where $Z_\theta=\int_\mathbf{x} \exp\left(-\Phi_{\theta}(\mathbf{x})\right)d\mathbf{x}$ is the normalizing constant. We treat the energy function $\Phi_{\theta}$ as a neural network composed of $l$ affine and non-linear activation functions:
\begin{equation}
    \label{eq:neuralphi}
    \Phi_{\theta}(\mathbf{x}) = h_l\left(\omega_l\left(h_{l-1}\left(\ldots\left(\omega_1\mathbf{x}+b_1\right)\ldots\right)\right)+b_l\right),
\end{equation}
 where $h(x)=\max(0,x)$, and $\theta=\left(\omega_1,b_1,\ldots,\omega_l,b_l\right)$ denoting the weight matrices and bias vectors of the neural network. \\

\noindent\textbf{Particle Perspective.} 
Let $\{\mathbf{x}_t^i\}_{i=1}^{n}$ be a set of Brownian particles\footnote{Particles with random or Brownian motion, will be called dynamic particles later in this paper.} with kinetic energy initiated from a Gaussian reference distribution $q\left(\mathbf{x}\right)=\frac{1}{(2\pi\sigma^2)^{\vert \mathbf{x}\vert/2}} \exp\left(-\frac{1}{2\sigma^2}| \mathbf{x}|^2\right)$ with mean $0$ and variance $\sigma^{2}$. The empirical measure of these particles at each time step $t$ can be described by $\rho_t=\frac{1}{n}\sum_{i=1}^{n}\delta_{\mathbf{x}_t^i}$, and $\rho_t$ converges to $p_{\theta}$ when $n\rightarrow\infty$ and $t\rightarrow\infty$ ($\rho_{\infty}$), where $\delta$ is the Dirac delta function. In this sense, we denote the updated $\theta$ at each time step with foot notation $t$ and assume that $\mathbb{E}_{\rho_t}[\Phi_{\theta_t}] \rightarrow \mathbb{E}_{p_{\theta}}[\Phi_{\theta}]$. 
The moving pattern of dynamic particles $\{\mathbf{x}_t^i\}_{i=1}^n$ is determined by the distribution parameters $\theta_t$, which is estimated stepwisely with time $t$, and thus the parameter estimation process can be regarded as an evolution of dynamic particles~\citep{cai2019frame}. 
Once the energy decreases to $0$, the particles reach a stable status and maintain static. In fact, stepwisely updating $\theta$ is to maximize the entropy functional $\mathcal{H}(p_{\theta})=-\int p_{\theta}(\mathbf{x})\log p_{\theta}(\mathbf{x})d\mathbf{x}$ which subjects to $\mathbb{E}_{p_{\theta}}[\Phi_{\theta}] = \mathbb{E}_{p_{\text{data}}}[\Phi_{\theta}]$ (MLE). In particle perspective, this updating process can be regarded as the dynamic of $\{\theta_t\}$ guiding stochastic movements of $\{\mathbf{x}_t^i\}_{i=1}^{n}$ to $p_{\theta}$, which also includes an evolution of the density function $\{\rho_t\}$. As a result, if $\{\mathbf{x}_t^i\}_{i=1}^{n}$ and $\{\theta_t\}$ converge to their optimal $\tilde{\mathbf{x}}$ and $\theta^*$ respectively, we will obtain $\tilde{\mathbf{x}}\sim \rho_{\infty}\approx p_{\theta^*}\approx p_{\text{data}}$ as $t\rightarrow\infty$. Overall, typical particle perspective of learning EBM is estimating the parameter $\theta$ step-by-step through MLE transferring particles $\{\mathbf{x}^i_{t-1}\}^n_{i=1}$ to $\{\mathbf{x}^i_{t}\}^n_{i=1}$ until the empirical measure converges to $\rho_{\infty}$. Note when $t\rightarrow\infty$, $\rho_t$ is still an empirical measure considering the ideal approximation of $\rho_{\infty}$ is the real data distribution $p_{data}$ that is factually inaccessible.
\\

\noindent\textbf{Evaluation.} Let $\mathbf{X}\sim p_{data}$ be the real data samples. To measure the convergence of learning, we first consider the following mean absolute energy error~(MAEE) between $\mathbb{E}_{\rho_t}\left[\Phi_{\theta_t}\left(\mathbf{x}_t\right)\right]$ and $\mathbb{E}_{p_{\text{data}}}\left[\Phi_{\theta_t}\left(\mathbf{X}\right) \right]$:
\begin{equation}
\label{eq:maee_distance}
R = \left\lvert\frac{1}{n}\sum_{i=1}^{n} \Phi_{\theta_t}\left(\mathbf{x}_t^i\right) - \frac{1}{m}\sum_{i=1}^{m} \Phi_{\theta_t}\left(\mathbf{X}^i\right) \right\rvert.
\end{equation}
Apparently, the smaller the MAEE is, the closer the particles $\mathbf{x}$ are transported to $\mathbf{X}$.
 Yet Eq.~\eqref{eq:maee_distance} will misjudge a scenario that $\forall \mathbf{x}\in\mathbb{R}^d$, $\Phi_{\theta_t}$ vanishes to $0$ (theoretically impossible) since in this scenario $R=0$. Thus, we further analyze the learning status by inspecting on the vector field of $\Phi_{\theta_t}\left(\mathbf{x}\right)$, i.e., $\nabla_{\mathbf{x}}\Phi_{\theta_t}\left(\mathbf{x}\right)$, indicating how far and where the current force defined by the model is pushing the particles to. The vector field is also perpendicular to the learned probabilistic manifold and more convenient to observe whether the model vanishes or not.

Furthermore, to measure the generative property of EBM, we also evaluate the final learned model with two metrics, which are widely used on visual data generation---Fr\'echet Inception Distance (FID)~\citep{heusel2017gans} and Inception Score (IS). 
Specifically, FID measures the inner context structure similarity between the generated data and the real data, while IS estimates the diversity of the generation. Other evaluation approaches that measure the divergence between empirical probability distribution and the real data distribution will be 
complemented later in the experiment Sec.~\ref{sec:experiment}.

\section{Related Work}
\label{sec:pre}
\textbf{Germination.} The essence of EBMs is to construct a mapping from the input space to a scalar, which is called ``energy function''. Ever since~\citep{ackley1985learning}, plenty of applications and theory studies on energy-based models have been published, such as~\citep{hinton2002training} and~\citep{lecun2006tutorial}. Generally, learning an EBM is a process that the energy function parameter $\theta$ is trained by minimizing the KL divergence from data distribution $p_{\text{data}}$ to model distribution $p_{\theta}$, i.e.,  $D_{\text{KL}}\left(p_{\text{data}}\|p_{\theta}\right)$. This learning approach is exactly based on the idea of the well-known maximum likelihood estimation (MLE), which yields gradients of $\theta$ in the form
\begin{equation}
\begin{split}
\label{eq:mle}
    \nabla_{\theta}D_{\text{KL}}(p_{\text{data}}\|p_{\theta}) &= -\frac{1}{n}\sum_{i=1}^{n}\nabla_{\theta}\log p_{\theta}(\mathbf{x}^i) \\ 
    &=  \nabla_{\theta}\mathbb{E}_{\mathbf{x}\sim p_{\text{data}}}[\Phi_{\theta}(\mathbf{x})] - \nabla_{\theta}\mathbb{E}_{\mathbf{x}\sim p_{\theta}}[\Phi_{\theta}(\mathbf{x})].
\end{split}
\end{equation}
This method draws samples using MCMC to estimate the gradient, which was widely adopted by early works of EBMs. In the special case of a restricted Boltzmann machine~\citep{hinton2002training}, estimating $\theta$ by minimizing the contrastive divergence is to simplify and boost the training.\\

\noindent\textbf{Blossom.} Recent progresses on EBMs exploit deep neural networks~\citep{simonyan2014very,he2016deep,du2019implicit,song2019generative} as the energy function to learn high-dimensional data. Training large-scale EBMs is extremely difficult due to a large amount of data and the growing parameter capacity. The expensive computation also makes the partition function $Z_\theta$ become harder to reliably estimate. Recent works~\citep{dai2014generative, lu2015learning, xie2016theory, du2019implicit} ease this pain by using a sampler based on SGLD~\citep{welling2011bayesian} as follows:
\begin{equation}
    \label{eq:sgld}
    \mathbf{x}_{t+1} = \mathbf{x}_t -  \nabla_{\mathbf{x}_{t}}\Phi_{\theta}(\mathbf{x}_{t}) + \sqrt{2} \mathbf{z}_t,
\end{equation}
where $\mathbf{z}_t\sim\mathcal{N}(0,I)$ is standard Gaussian and $\mathbf{x}_0\sim q(\mathbf{x})$. Although Eq.~\eqref{eq:sgld} is widely applied, Liu et al.~\citep{liu2016stein} have pointed out that the non-smooth perturbation of the derivative of KL divergence in SGLD might impair the convergence. They mitigate this weakness by iteratively transporting a set of particles to match the given distribution with Stein’s identity, which in general shares the similar viewpoint with us. Other researchers also notice that the metric matters in learning. For example, instead of using KL divergence, Montavon et al.~\citep{montavon2016wasserstein} train the restricted Boltzman machine by minimizing the Wasserstein distance between empirical distribution and real data distribution. 
Recently, the Wasserstein distance is also employed to address high-dimensional data issue~\citep{lin2020projection}. \\

\noindent\textbf{Bottleneck.} Despite a great success of the sampling-based learning, it is still impractical to involve millions of dimensional data, since the well-known sampling algorithms still take exponential time to converge. 
This problem can be circumvented by applying a straightforward generator network to generate samples, which is first proposed in generative adversarial networks (GANs)~\citep{goodfellow2014}. The generator network is utilized to incorporate with an EBM to improve the model potential. For example,~\citep{xie2018cooperative} trained the generator and EBM cooperatively and~\citep{kim2016deep,wang2017learning} trained them adversarially. Score matching~\citep{hyvarinen2005estimation} also found its way to avoid heavy calculations during learning the EBMs, and matched the derivative of the model density with the derivative of the data density. This approach has recently been applied to high-dimensional images~\citep{song2019generative}.

%

\section{Gradient Flow in MLE}
\label{sec:klflow}
Based on the problem setup in Sec.~\ref{sec:setup}, in the following, we present how to reformulate the conventional EBM that uses MLE by the gradient flow (or particle) perspective.
Concretely, we treat the discrete-time evolution of empirical probability density $\rho_t$ of MLE as the KL discrete flow, making the particles to dynamically adhere to SGLD progress. For simplicity, we denote all the gradient flows with $\mathcal{F}$, and the foot notation to specify the different transferring metrics, for example, $\mathcal{F}_{\text{KL}}$ as KL discrete flow.

\subsection{KL Discrete Flow}
\label{sec:subklflow}
We first uncover this time-discrete and iterative variational learning scheme with respect to KL divergence for conventional MLE learning in EBMs. This scheme adheres to a flow functional
\begin{equation}
    \label{eq:klproximal}
    \mathcal{F}_{\text{KL}}(\rho_0; \rho) = \min_{\mu\in\mathcal{T}(\rho_0,\rho)} D_{\text{KL}}(\mu\|\rho_0) + \tau\mathbb{E}_{\rho}[\Phi_\theta],
\end{equation}
 where $\tau$ is the time interval between $\rho$ and $\rho_0$; energy parameter $\theta$ will also be omitted in later context for simplicity; $\mathcal{T}(\rho_0,\rho)$ denotes the collection of all the measures that have been stepwisely estimated before the ending $\rho$ and after the initial $\rho_0$. $D_{\text{KL}}$ in the Eq.~\eqref{eq:klproximal} measures the distance between $\mu$ and $\rho_0$. According to \citep{goldfeld2020convergence}, the KL divergence $D_{\text{KL}}$ between empirical measure $\mu$ and $\upsilon$ is defined as $D_{\text{KL}}(\mu||\upsilon)=\int\log(\frac{d\mu}{d\upsilon})d\mu$, namely the entropy difference $\mathcal{H}(\upsilon)-\mathcal{H}(\mu)$.

 We use $\operatorname*{argmin}_{\rho}\mathcal{F}_{\text{KL}}(\rho;\rho_0)$ as a proximal operator that seeks to find the optimal $\rho$ that satisfies Eq.~\eqref{eq:klproximal}, considering the gradient flow $\mathcal{F}_{\text{KL}}$ consisting of the KL divergence and the energy function $\Phi$. To this end, it is assumed that the learning process starts from a fixed discrete density $\rho_0$.  Accordingly, the final learning $\Phi$ with $\mathcal{F}_{\text{KL}}$ is obtaining the empirical measure $\rho_t$:
\begin{equation}
    \label{eq:kldflow}
     \rho_{t+1} = \operatorname*{argmin}_{\rho}\mathcal{F}_{\text{KL}}(\rho;\rho_t), \ \ \forall t\geq 0, \quad \rho_0=\frac{1}{n}\sum^n_{i=1}q_{\mathbf{x}_0^i}, \quad  t\rightarrow\infty,
\end{equation}
where the empirical distribution sequence $\rho_t$ starts with a certain Gaussian reference distribution $q$, and to be consistent with the discrete form of $\rho_t$, we write $\rho_0=\frac{1}{n}\sum^n_{i=1}q_{\mathbf{x}_0^i}$. This equation also implies that the evolutionary particle distribution $\rho_t$ within consecutive discrete-time is restricted by KL divergence; namely, $\{\mathbf{x}^{i}_t\}_{i=1}^{n}$ are drifted by the gradient of \eqref{eq:klproximal} and diffused by increasing their entropy (decreasing the KL divergence) to search the parameter space. 

Next, based on the particle perspective that Eq.~\eqref{eq:klproximal} represents a system of $n$ independent Brownian particles $\{\mathbf{x}^{i}_t\}_{i=1}^{n}$, the positions of $\mathbf{x}^{i}_t$ in $\mathbb{R}^{d}$ are given by a Wiener process~(or Brownian motion, a stochastic process of moving particles) satisfying the following {\em stochastic differential equation}~(SDE):
\begin{equation}
    \label{eq:sde}
    d\mathbf{x}_{t} = -\nabla\Phi(\mathbf{x}_t)dt + \sqrt{2}dW_{t},
\end{equation}
where $-\nabla\Phi(\mathbf{x}_t)dt$ on the RHS is the drift term and $W_t$ denotes diffusion term (named after Wiener process). Eq. \eqref{eq:sde} is mathematically sound and is the intuitive notion of SDE especially when SDE is ordinary differential equations plus noise. Note that the Euler discretization of Eq.~\eqref{eq:sde} is exactly the same with SGLD in Eq.~\eqref{eq:sgld}. 

To this end, it seems feasible to interpret the conventional EBM theory from MLE perspective into the gradient flow perspective. Before formally proving that, we use the following theorem that combines Eq.~\eqref{eq:klproximal},~\eqref{eq:kldflow} and~\eqref{eq:sgld}, to relate with the EBM distribution in Eq.~\eqref{eq:ebm}:
\begin{theorem}
  \label{thm:1}
  If the i.i.d. particles $\{\mathbf{x}_t^{i}\}_{i=1}^{n}$ are with common moment generating function $\mathbb{E}\left[e^{-\Phi(\mathbf{x})}\right]^{\frac{1}{c}}<\infty$ ($c$ could be any integer between $[1,d]$), the empirical measure $\rho_{t}$ of these particles satisfies the large deviation principle (LDP)~\citep{touchette2011basic} with a rate functional in the form of $\mathcal{F}_{\text{KL}}$.
\end{theorem}
Theorem~\ref{thm:1} implies that the particle-based model refers to an evolving state where there is a periodic goal for the energy changing. When the system reaches the equilibrium, the states with lower energies are more likely to occur. Those large amount of states produced by the iteration process conform to LDP if the probability of visiting a non-typical state is exponentially small. Specifically, LDP can be briefly described as: a random variable $S_k$ or its probabilistic distribution function $p_{S_k}$ satisfies LDP if the following limit exists:
\begin{equation}
  \lim_{k \to \infty}-\frac{1}{k}\ln p_{S_k}(s)=I(s),   
\end{equation}
and gives rise to a rate function $I(s)$, which is not
everywhere zero. In our case, $p_{S_k}(s)=p_{\theta_t}(\mathbf{x})$, and the distribution sequence $p$ varies by the optimization time $t$; $I(s)$ is identical to the energy function $\Phi(\mathbf{x})$, $k\rightarrow\infty$ when $t\rightarrow\infty$.
In the context of large deviation theory, the logarithmic moment generating function  $\mathbb{E}\left[e^{-\Phi(\mathbf{x})}\right]^{\frac{1}{c}}$is also called the free energy function, and thus a simple and natural model for such a system at the equilibrium is exactly given by an EBM. Detailed proof of Theorem~\ref{thm:1} is provided in Appendix~\ref{append:proof}.

\subsection{Learning EBM via KL Discrete Flow}
The connotation of the KL discrete flow $\mathcal{F}_{\text{KL}}$ describes a system of Brownian particles with evolutionary probability distributions, which also inspires a new perspective on interpreting the learning of EBMs in the discrete states. Following the setup in Sec.~\ref{sec:setup}, we would maximize the entropy functional $\mathcal{H}(\rho_t)$ that subjects to $\mathbb{E}_{\rho_t}[\Phi] = \mathbb{E}_{p_{\text{data}}}[\Phi]$, which is equivalent to minimizing the KL divergence $D_{\text{KL}}(\rho_t\|\rho_{t-1})$. Next we show how to mathematically equate the evolving particles expression $\mathcal{F}_{\text{KL}}$ with Eq.~\eqref{eq:mle}.

 To rewrite the original maximum entropy problem in gradient flow form, we consider for any conjugate distributions $p$ and $q$ in the space of Borel probability measures $\mathcal{P}(\mathbb{R}^d)$ on any given subset of $\mathbb{R}^d$, there exists $f\in \mathcal{P}$ that $p$ and $q$ satisfies a linear constraint $p=fq$. Since the only probability family that satisfies such condition is the exponential family, and that is exactly where the EBM density $p$, its conjugate Gaussian prior $q$ and the likelihood function $\frac{1}{Z_{\theta}}e^{-\Phi_{\theta}(\mathbf{x})}$ belongs to. Then we can construct a linear constraint space $\mathcal{P}_{\text{lin}}$:
\begin{equation*}
\begin{split}
   \mathcal{P}_{\mathrm{lin}} = \bigg\{p\in\mathcal{P}:\exists f\geq 0~s.t.~p=fq, \int_{\mathbb{R}^d}  pd\mathbf{x}=1; 
  \mathbb{E}_{p}[\Phi(\mathbf{x})]=\mathbb{E}_{p_{\text{data}}}[\Phi]; f, q\in\mathcal{P}\bigg\},
\end{split}
\end{equation*}
which is a space of probability densities (sum up to one) satisfying a linear constraint (constraining the density to be in the exponential form). The product of Borel measure in the definition of $\mathcal{P}_{lin}$ is a special case based on the original product definition and will be further explained in Appendix~\ref{append:product}.

With Theorem~\ref{thm:1} and linear space $\mathcal{P}_{\textrm{\text{lin}}}$, we can conclude a following corollary indicating that iteratively solving the KL discrete flow optimization problem in Eq. \eqref{eq:kldflow} yields an identical solution with that of MLE.
\begin{corollary}
  \label{coro:klflow}
  Given a list of Brownian particles $\{\mathbf{x}_t^{i}\}_{i=1}^{n}$, of which the corresponding empirical measure is $\rho_t$, starting from $\rho_0=\frac{1}{n}\sum q_{\mathbf{x}}$, for $t>0$, if $\rho_t$ evolves according to the following constrained optimization problem:
  \begin{equation}
      \begin{split}
          \label{eq:klconstrain}
          \rho_{t+1} = &\operatorname*{argmin}_{\rho}\min_{\mu\in\mathcal{T}(\rho,\rho_t)}D_{\text{KL}}(\mu\|\rho_{t}), \\
          \text{s.t.} \quad \mathbb{E}_{\rho}[\Phi] = & \mathbb{E}_{p_{\text{data}}}[\Phi], \quad \rho_0=\frac{1}{n}\sum^n_{i=1}q_{\mathbf{x}_0^i}, \quad\forall\rho\in\mathcal{P}_{\mathrm{lin}},
      \end{split}
  \end{equation}
  then as $t\rightarrow\infty$, the steady state $\rho_{\infty}$ is a weak approximation of $p_{\theta^*}$ as well as $p_{\text{data}}$.
\end{corollary}
Corollary~\ref{coro:klflow} inherently describes an evolving process of density $\rho_t$ iterating from $q$ in $\mathcal{P}_{\mathrm{lin}}$. This MLE optimization problem solved in a Lagrangian form by rewriting Eq.~\eqref{eq:klconstrain} into the paramerterized Lagrangian functional $\mathcal{F}_{\text{KL}}^{l}$ that incorporates the Lagrangian multiplier $\theta$ into $\Phi$:
\begin{equation}
  \begin{split}
      \label{eq:klflow_lag}
      & \mathcal{F}_{\text{KL}}^{l}(\rho;\rho_t) =\operatorname*{min}_{\rho}\operatorname*{max}_{\theta
      } \bigg\{D_{\text{KL}}(\rho\|\rho_{t}) + \mathbb{E}_{\rho}[\Phi_{\theta}] - \mathbb{E}_{p_{\text{data}}}[\Phi_{\theta}]\bigg\}.
  \end{split}
\end{equation}
Once we have $\mathcal{F}_{\text{KL}}^{l}$, the earlier optimization problem \eqref{eq:klproximal} is now rephrased as a saddle-point solving problem because of Corollary \ref{coro:klflow}. Recall that \eqref{eq:klproximal} in Sec.~\ref{sec:subklflow}, at discrete time $t$, minimizing the equivalent \eqref{eq:klflow_lag} w.r.t. $\rho$ also implies an SGLD on $\{\mathbf{x}_t^i\}_{i=1}^{n}$. If we fix $\theta$, the sampling scheme of dynamic particles in \eqref{eq:sgld} makes the empirical measure follow the KL discrete flow $\mathcal{F}_{\text{KL}}^{l}$, and the flow will fluctuate when $\theta$ varies. The parameter $\theta$ is updated by calculating $\nabla_{\theta}\mathcal{F}_{\text{KL}}^{l}$, implying that $\theta$ can dynamically be transformed into the desired through a transition map~(if regarding all the learned distributions as transformed reference distributions stem from $q$). The sampling-based learning process of EBMs driven by $\mathcal{F}_{\text{KL}}^{l}$ can be written in
\begin{equation}
  \label{eq:standard_learning}
  \left\{
  \begin{aligned}
      \mathbf{x}_{t+1} &= \mathbf{x}_t -  \nabla_{\mathbf{x}_{t}}\Phi_{\theta_t}(\mathbf{x}_{t}) + \sqrt{2}\mathbf{z}_t \\
      {\theta}_{t+1} & = {\theta}_{t} + \nabla_{\theta_t}\mathbb{E}_{\mathbf{X}\sim p_{\text{data}}}[\Phi_{\theta_t}(\mathbf{X})] -  \nabla_{\theta_t}\mathbb{E}_{\mathbf{x}_t\sim\rho_{t}}[\Phi_{\theta_t}(\mathbf{x}_t)].
  \end{aligned}
  \right.
\end{equation}
The above numerical computation equations describe exactly the same process of MLE in Eq.~\eqref{eq:mle} and SGLD in Eq.~\eqref{eq:sgld}, in a more general way. 
Namely, to this end, we have converted the learning of an EBM by MLE in \eqref{eq:mle} into optimizing a Lagrangian form KL discrete flow $\mathcal{F}_{\text{KL}}^{l}$ defined in \eqref{eq:klflow_lag}.

\subsection{Disadvantages Uncovered by KL Discrete Flow}
\label{sec:problem}
In theory, learning EBMs with MLE and SGLD expects that the cross-updating procedure ceases until $\rho_t$ converges to $p_{\theta^*}$ and $p_{\text{data}}$. At this moment, $\mathbf{x}_t$ approaches to the observed data point $\mathbf{X}$ as well. As mentioned in Sec.~\ref{sec:introduction}, minimizing the local time KL divergence is equivalent to projecting from $\rho_{t}$ to $\mathcal{P}_{\text{lin}}$ to find the optimal $\rho_{t+1}$. While in practice, this merely happens when the time interval $\tau$ is infinitely small, and thus it is impossible to obtain the optimal state (or the global minimum). It is the unbounded orthogonal property of this projection that exposes the original learning procedure to two risks as follows. 
\begin{itemize}
    \item The given searching space has the risk of causing local minimum and non-convergence;
    \item When it is back-propagating on $\mathbf{x}$ and $\theta$, there might be possibly incorrect~(or unconstrained) gradient values.
\end{itemize}
The formed global evolutionary paths of $\mathbf{x}_t$ and $\rho_t$ thereby are non-smooth, and the aforementioned defects would be amplified and end up with a failure when applied to higher-dimensional spaces. 

In EBM, its learning process can be regarded as an energy dissipating from high energy (chaotic) to a relatively low energy (ordered) status. The proposed KL discrete flow perspective uncovers a unstable nature of this dissipation process. Moreover, the instability is a hidden cause of mode collapse. Zero energy ($\Phi_{\theta_t}(\mathbf{x}_t)=0$) is the sign that the model crumbles and outputs $0$ for all inputs. When the energy goes negative, it implies that the final stage particles are much further away from the observed data points. We have identified both phenomena in the experiments (Sec.~\ref{sbsec:toyexample}).

\begin{figure*}[ht]
  \begin{center}
  \begin{adjustbox}{width=1\linewidth}
  \tikzset{every picture/.style={line width=0.75pt}} 
  \begin{tikzpicture}[x=0.75pt,y=0.75pt,yscale=-1,xscale=1]

  \draw   (66,59.47) .. controls (66,56.63) and (68.3,54.33) .. (71.13,54.33) -- (107.87,54.33) .. controls (110.7,54.33) and (113,56.63) .. (113,59.47) -- (113,74.87) .. controls (113,77.7) and (110.7,80) .. (107.87,80) -- (71.13,80) .. controls (68.3,80) and (66,77.7) .. (66,74.87) -- cycle ;
  \draw    (12,42.33) -- (655,41.33) ;
  \draw    (113,68) -- (179,68) ;
  \draw [shift={(181,68)}, rotate = 180] [fill={rgb, 255:red, 0; green, 0; blue, 0 }  ][line width=0.08]  [draw opacity=0] (12,-3) -- (0,0) -- (12,3) -- cycle    ;
  \draw    (89.5,79.17) -- (89.5,96.83) ;
  \draw [shift={(89.5,98.83)}, rotate = 270] [fill={rgb, 255:red, 0; green, 0; blue, 0 }  ][line width=0.08]  [draw opacity=0] (12,-3) -- (0,0) -- (12,3) -- cycle    ;
  \draw   (50,98) -- (134,98) -- (134,183.33) -- (50,183.33) -- cycle ;
  \draw   (180,60.47) .. controls (180,57.63) and (182.3,55.33) .. (185.13,55.33) -- (288.87,55.33) .. controls (291.7,55.33) and (294,57.63) .. (294,60.47) -- (294,75.87) .. controls (294,78.7) and (291.7,81) .. (288.87,81) -- (185.13,81) .. controls (182.3,81) and (180,78.7) .. (180,75.87) -- cycle ;
  \draw    (235,80.33) -- (235,107.33) ;
  \draw [shift={(235,109.33)}, rotate = 270] [fill={rgb, 255:red, 0; green, 0; blue, 0 }  ][line width=0.08]  [draw opacity=0] (12,-3) -- (0,0) -- (12,3) -- cycle    ;
  \draw   (204,109) -- (266,109) -- (266,162.33) -- (204,162.33) -- cycle ;
  \draw    (134,137) -- (202,137.32) ;
  \draw [shift={(204,137.33)}, rotate = 180.27] [fill={rgb, 255:red, 0; green, 0; blue, 0 }  ][line width=0.08]  [draw opacity=0] (12,-3) -- (0,0) -- (12,3) -- cycle    ;
  \draw    (294,68) -- (359,68) ;
  \draw [shift={(361,68)}, rotate = 180] [fill={rgb, 255:red, 0; green, 0; blue, 0 }  ][line width=0.08]  [draw opacity=0] (12,-3) -- (0,0) -- (12,3) -- cycle    ;
  \draw   (361,60.47) .. controls (361,57.63) and (363.3,55.33) .. (366.13,55.33) -- (469.87,55.33) .. controls (472.7,55.33) and (475,57.63) .. (475,60.47) -- (475,75.87) .. controls (475,78.7) and (472.7,81) .. (469.87,81) -- (366.13,81) .. controls (363.3,81) and (361,78.7) .. (361,75.87) -- cycle ;
  \draw    (416,80.33) -- (416,107.33) ;
  \draw [shift={(416,109.33)}, rotate = 270] [fill={rgb, 255:red, 0; green, 0; blue, 0 }  ][line width=0.08]  [draw opacity=0] (12,-3) -- (0,0) -- (12,3) -- cycle    ;
  \draw   (385,109) -- (447,109) -- (447,162.33) -- (385,162.33) -- cycle ;
  \draw    (266,137) -- (384,137) ;
  \draw [shift={(386,137)}, rotate = 180] [fill={rgb, 255:red, 0; green, 0; blue, 0 }  ][line width=0.08]  [draw opacity=0] (12,-3) -- (0,0) -- (12,3) -- cycle    ;
  \draw    (475,68.47) -- (540,68.47) ;
  \draw [shift={(542,68.47)}, rotate = 180] [fill={rgb, 255:red, 0; green, 0; blue, 0 }  ][line width=0.08]  [draw opacity=0] (12,-3) -- (0,0) -- (12,3) -- cycle    ;
  \draw   (543,57.47) .. controls (543,54.63) and (545.3,50.33) .. (548.13,50.33) -- (651.87,50.33) .. controls (654.7,50.33) and (657,54.63) .. (657,57.47) -- (657,72.87) .. controls (657,75.7) and (654.7,78) .. (651.87,80) -- (548.13,80) .. controls (545.3,80) and (543,75.7) .. (543,72.87) -- cycle ;
  \draw    (598,80) -- (598,108) ;
  \draw [shift={(598,110)}, rotate = 270] [fill={rgb, 255:red, 0; green, 0; blue, 0 }  ][line width=0.08]  [draw opacity=0] (12,-3) -- (0,0) -- (12,3) -- cycle    ;
  \draw   (566,109) -- (628,109) -- (628,162.33) -- (566,162.33) -- cycle ;
  \draw    (447,137) -- (565,137) ;
  \draw [shift={(567,137)}, rotate = 180] [fill={rgb, 255:red, 0; green, 0; blue, 0 }  ][line width=0.08]  [draw opacity=0] (12,-3) -- (0,0) -- (12,3) -- cycle    ;
  \draw    (90,184.33) -- (90,211.33) ;
  \draw [shift={(90,213.33)}, rotate = 270] [fill={rgb, 255:red, 0; green, 0; blue, 0 }  ][line width=0.08]  [draw opacity=0] (12,-3) -- (0,0) -- (12,3) -- cycle    ;
  \draw   (60,212) -- (122,212) -- (122,235) -- (60,235) -- cycle ;
  \draw    (234,162) -- (234,210) ;
  \draw [shift={(234,212)}, rotate = 270] [fill={rgb, 255:red, 0; green, 0; blue, 0 }  ][line width=0.08]  [draw opacity=0] (12,-3) -- (0,0) -- (12,3) -- cycle    ;
  \draw   (204,212) -- (266,212) -- (266,235) -- (204,235) -- cycle ;
  \draw    (123,224) -- (202,224) ;
  \draw [shift={(204,224)}, rotate = 180] [fill={rgb, 255:red, 0; green, 0; blue, 0 }  ][line width=0.08]  [draw opacity=0] (12,-3) -- (0,0) -- (12,3) -- cycle    ;
  \draw    (266,224) -- (382,224) ;
  \draw [shift={(384,224)}, rotate = 180] [fill={rgb, 255:red, 0; green, 0; blue, 0 }  ][line width=0.08]  [draw opacity=0] (12,-3) -- (0,0) -- (12,3) -- cycle    ;
  \draw   (384,209) -- (446,209) -- (446,236) -- (384,236) -- cycle ;
  \draw   (413.68,162.88) -- (413.72,189.97) -- (361.78,190.04) ;
  \draw    (256,190) -- (361.78,190.04) ;
  \draw    (256,190) -- (256,211) ;
  \draw [shift={(256,213)}, rotate = 270] [fill={rgb, 255:red, 0; green, 0; blue, 0 }  ][line width=0.08]  [draw opacity=0] (12,-3) -- (0,0) -- (12,3) -- cycle    ;
  \draw   (597.68,162.88) -- (597.71,189.97) -- (545.77,189.97) ;
  \draw    (428,189.97) -- (545.77,189.97) ;
  \draw    (428,189.97) -- (428,209) ;
  \draw [shift={(428,209)}, rotate = 270] [fill={rgb, 255:red, 0; green, 0; blue, 0 }  ][line width=0.08]  [draw opacity=0] (12,-3) -- (0,0) -- (12,3) -- cycle    ;
  \draw    (446,223) -- (562,223) ;
  \draw [shift={(564,223)}, rotate = 180] [fill={rgb, 255:red, 0; green, 0; blue, 0 }  ][line width=0.08]  [draw opacity=0] (12,-3) -- (0,0) -- (12,3) -- cycle    ;
  \draw   (564,210) -- (626,210) -- (626,235) -- (564,235) -- cycle ;

  \draw (52,17.4) node [anchor=north west][inner sep=0.75pt]    {1) $p_{\theta _{t}}\rightarrow p_{data} $; 2) $\mathbf{x}_{t} \sim p_{\theta _{t}}$};
  \draw (7,19) node [anchor=north west][inner sep=0.75pt]   [align=left] {\textbf{Goal:}};
  \draw (8,60) node [anchor=north west][inner sep=0.75pt]  [font=\footnotesize] [align=left] {Method:};
  \draw (52,144.4) node [anchor=north west][inner sep=0.75pt]  [font=\footnotesize]  {$ \begin{array}{l}
  \mathbb{E}_{p_{\theta }}[ \Phi (\mathbf{x})] \ =\\
  \mathbb{E}_{p_{data}}[ \Phi (\mathbf{x})] \ 
  \end{array}$};
  \draw (57,105) node [anchor=north west][inner sep=0.75pt]  [font=\footnotesize] [align=left] {\begin{minipage}[lt]{48.53pt}\setlength\topsep{0pt}
  Learning by
  \begin{center}
  sampling s.t.
  \end{center}

  \end{minipage}};
  \draw (212,115) node [anchor=north west][inner sep=0.75pt]  [font=\footnotesize] [align=left] {optimize };
  \draw (207,134.4) node [anchor=north west][inner sep=0.75pt]  [font=\footnotesize]  {$\mathcal{F}_{\text{KL}}^{l}( \rho ;\rho _{t})$};
  \draw (188.13,61.33) node [anchor=north west][inner sep=0.75pt]  [font=\footnotesize] [align=left] {KL discrete flow};
  \draw (74.13,61.33) node [anchor=north west][inner sep=0.75pt]  [font=\footnotesize] [align=left] {MLE};
  \draw (138,121) node [anchor=north west][inner sep=0.75pt]  [font=\footnotesize] [align=left] {Corollary 2};
  \draw (369.13,61.33) node [anchor=north west][inner sep=0.75pt]  [font=\footnotesize] [align=left] {JKO discrete flow};
  \draw (393,115) node [anchor=north west][inner sep=0.75pt]  [font=\footnotesize] [align=left] {optimize };
  \draw (388,134.4) node [anchor=north west][inner sep=0.75pt]  [font=\footnotesize]  {$\mathcal{F}_{w^{2}}( \rho ;\rho _{t})$};
  \draw (295,121) node [anchor=north west][inner sep=0.75pt]  [font=\footnotesize] [align=left] {Theorem 3};
  \draw (560.13,52  .33) node [anchor=north west][inner sep=0.75pt]  [font=\footnotesize] [align=center] {Near-Optimal \\ discrete flow};
  \draw (574,115) node [anchor=north west][inner sep=0.75pt]  [font=\footnotesize] [align=left] {optimize };
  \draw (567,134.4) node [anchor=north west][inner sep=0.75pt]  [font=\footnotesize]  {$\mathcal{F}_{W^2}^{l}( \rho ;\rho _{t})$};
  \draw (474,121) node [anchor=north west][inner sep=0.75pt]  [font=\footnotesize] [align=left] {Claim 1};
  \draw (116,53) node [anchor=north west][inner sep=0.75pt]  [font=\footnotesize] [align=left] {equivalent};
  \draw (297,53) node [anchor=north west][inner sep=0.75pt]  [font=\footnotesize] [align=left] {equivalent};
  \draw (474,53.33) node [anchor=north west][inner sep=0.75pt]  [font=\footnotesize] [align=left] {approximate };
  \draw (11,128) node [anchor=north west][inner sep=0.75pt]  [font=\footnotesize] [align=left] {Idea:};
  \draw (70,217) node [anchor=north west][inner sep=0.75pt]  [font=\footnotesize] [align=left] {\eqref{eq:mle}; \eqref{eq:sgld}};
  \draw (4,210) node [anchor=north west][inner sep=0.75pt]  [font=\footnotesize] [align=left] {\begin{minipage}[lt]{36.75pt}\setlength\topsep{0pt}
  \begin{center}
  Learning:
  \end{center}

  \end{minipage}};
  \draw (221,218) node [anchor=north west][inner sep=0.75pt]  [font=\footnotesize] [align=left] {\eqref{eq:standard_learning}};
  \draw (133,209) node [anchor=north west][inner sep=0.75pt]  [font=\footnotesize] [align=left] {equivalent };
  \draw (387,209) node [anchor=north west][inner sep=0.75pt]  [font=\footnotesize] [align=left] {\begin{minipage}[lt]{33.11pt}\setlength\topsep{0pt}
  \begin{center}
  \eqref{eq:new_sde_alpha2};\eqref{eq:new_learning}b
  \end{center}

  \end{minipage}};
  \draw (313,176.4) node [anchor=north west][inner sep=0.75pt]  [font=\footnotesize]  {$\tau \rightarrow 0$};
  \draw (270,209) node [anchor=north west][inner sep=0.75pt]  [font=\footnotesize] [align=left] {2 order approximate };
  \draw (460,175) node [anchor=north west][inner sep=0.75pt]  [font=\footnotesize]  {Functional derivative};
  \draw (581,216) node [anchor=north west][inner sep=0.75pt]  [font=\footnotesize] [align=left] {\eqref{eq:new_learning}};
  \draw (463,209) node [anchor=north west][inner sep=0.75pt]  [font=\footnotesize] [align=left] {Theorem 4 };
  \end{tikzpicture}
\end{adjustbox}
\caption{This figure depicts the general philosophy of deducing the near-optimal learning EBM method. The goal is to learn the ideal $p_{\theta_t}$ that converges to the real data distribution $p_{data}$ and obtain the realistic synthesis $\mathbf{x}$. The first line in this theoretical graph displays an evolution of the learning EBM in this paper, summarizing different perspectives and optimization purposes. The second line specifies the objective functions in the first line. The bottom presents the numerical computing equations. Specifically,
`b' stands for the bottom equation. Each $\mathcal{F}$ stands for different discrete gradient flow. \label{fig:deducing}}
\end{center}

\end{figure*}

\section{A Near Optimal Gradient Flow}
\label{sec:ourmethod}
In this section, we first introduce the JKO discrete flow based on optimal transport in a dynamic form from the particle perspective. Second, we show the relative equivalence between the proposed KL and JKO discrete flow. Since JKO still has the disadvantage of high computation consumption, to solve this issue, we further propose a near-optimal discrete flow to approximate JKO discrete flow. The deduction logic is summarized in the Figure~\ref{fig:deducing}, where all the theoretical relations among gradient flows are also presented.

\subsection{Optimal Transport and Prerequisites}
\label{sec:wd} 
Computing the Wasserstein metrics among measures  guides the learning direction of EBM is similar to computing KL divergence in KL discrete flow. However, JKO discrete flow is not another simple version of KL discrete flow with its KL divergence substituted by Wasserstein distance. In fact, this substitution requires series of preconditions. In this section we first introduce the Wasserstein metric for measuring the learning space that applied in JKO discrete flow and two preconditions for further proof on its equivalence with KL discrete flow.

For the random variable $\mathbf{x}_{t_1}$ in the Euclidean space at time $t_1$, the cost with $\mathbf{x}_{t}$ at time $t_2$ is $c(\mathbf{x}_{t_1}, \mathbf{x}_{t_2})=|\mathbf{x}_{t_2}-\mathbf{x}_{t_1}|^r$ and the corresponding optimal transport problem with $r-$order Wasserstein distance is defined as:
\begin{equation}
\label{eq:wmetric}
\begin{split}
    w^r(\mu_{1}, \mu_{2}):=\min\bigg\{\int_{\mathbb{R}^d\times\mathbb{R}^d}|\mathbf{x}_{t_1} -\mathbf{x}_{t_2} |^r d\pi(\mathbf{x}_{t_1}, \mathbf{x}_{t_2}):
    \pi\in\Pi(\mu_{1}, \mu_{2})\bigg\}.
\end{split}
\end{equation}
$\Pi$ here is the collection of optimal coupling. Before directly applying this equation, we here present its dynamic form within a time interval $[t, t+1]$. This well-defined Wasserstein metric shares the same mass as the original one.
Benamou and Brenier in~\citep{benamou2000computational} claim that the space of probability measures endowed with Wasserstein distances turns out to be a geodesic space. Thus, with the dynamic form of Wasserstein metric, the corresponding optimal transport problem above can be reformulated into the following:
\begin{equation}
\label{eq:bb_wmetric}
  \begin{split}
    W^r(\mu_{1}, \mu_{2}) := \min \bigg\{\int_{\text{Lip}([0,1];\mathbb{R}^d)}\int_t^{t+1}|\nu_{t}|^{r}dtd\rho :
    \rho_{t} = \mu_{1}, \rho_{t+1}=\mu_{2}\bigg\}.
  \end{split}
\end{equation}
 $\text{Lip}([0,1];\mathbb{R}^d)$ is the set of Lipshcitz curves in $\mathbb{R}^d$, parametrized over the interval $[0,1]$. A Lipschitz curve has a bounded velocity $L$, i.e. measure $\nu_t$ is Lipschitz continuous,
\begin{equation}
    \label{eq:boundv}
    |\nu_t|\leq L,\quad \forall t\geq 0.
\end{equation}
The minimum in Eq.~\eqref{eq:bb_wmetric} is taken among all pairs $(\rho_t,\nu_t)$, with $\rho_t$ being a Lipshcitz curve of measures and $\nu_t$ being obtained from the time-independent vector field. According to~\citep{ambrosio2008gradient}, for every Lipschitz or absolutely continuous curve $\rho_t$, there exists a map $\rho_t\cdot\nu_t$ from $[0,1]$ to the space of vector-valued measures, representing the flux $\div(\rho_{t}\cdot\nu_{t})$. By the divergence theorem, the opposite divergence of this flux is equal to the volume rate of fluid crossing the surface $\partial_{t}\rho_{t}$, forming the \textit{continuity equation} condition:
\begin{equation}
\label{eq:continuity}
    \partial_{t}\rho_{t} + \div(\rho_{t}\cdot\nu_{t})=0.
\end{equation}

Eq.~\eqref{eq:wmetric} fails to describe the time-evolutionary property of the transportation and merely implies the whole process depending on the optimal coupling $\pi$. The mass of Eq.~\eqref{eq:wmetric} located at $\mathbf{x}_1$ is moved to $\mathbf{x}_2$ along the manifold connecting time-evolutionary data points. On the contrary, the Benamou-Brenier form of Wasserstein metric in Eq.~\eqref{eq:bb_wmetric} is dynamic and many trajectories of transportation are admissible. They are driven by the tangent space of the manifold associated with Lipschitz curves. Therefore, Eq.~\eqref{eq:bb_wmetric} is the expected projection that specifies smooth and bounded gradient flow along the data manifold.

To obtain the JKO discrete flow by considering the Wasserstein distance in a basic discrete flow, here is another precondition within our consideration.
Following the initial work of~\citep{jordan1998variational}, we can discretize a non-linear partial differential equation (PDE) as the  gradient flow of an integral functional $F$ using implicit gradient step w.r.t. Wasserstein distance. Let the PDE be the continuity equation in \eqref{eq:continuity}, and the integral functional be
\begin{equation}
\label{eq:f}
    F(\rho) = -\mathcal{H}(\rho)+\mathbb{E}_{\rho}[\Phi] = \int_{\mathbb{R}^d}\rho\log\rho d\mathbf{x}+\int_{\mathbb{R}^d}\Phi\rho d\mathbf{x}.
\end{equation}
Through this functional, we can bridge the optimal transport problem and the gradient flow learning process by using a \textit{Fokker-Planck equation}:
\begin{equation}
  \label{eq:fp}
  \partial_{t} \rho = \div(\rho \nabla \Phi ) + \Delta \rho,
\end{equation}
where $\Delta$ denotes the second derivative of $\rho$ w.r.t. $\mathbf{x}$ and $\div$ means the divergence. This equation holds in our deduction because of the variational condition $\nu = -\nabla \frac{\delta F}{\delta\rho}$ as follows:
\begin{equation*}
    \nu = -\nabla \frac{\delta F}{\delta\rho} = -( \nabla\frac{\delta\rho\log\rho}{\delta\rho} + \nabla\frac{\delta\Phi\rho}{\delta\rho})= -(\nabla\log\rho + \nabla\Phi).
\end{equation*}
Here $\frac{\delta F}{\delta\rho}$ is the Euler-Lagrange first variation of integral function $F$. If substituting the above derivative into \eqref{eq:continuity}, then we can easily obtain the Fokker-Planck equation \eqref{eq:fp} in our situation. For the detailed deduction, please see Appendix~\ref{append:fp_proof}.

\subsection{The JKO Discrete Flow}
\label{sec:jkoflow} 
Based on the prerequisites mentioned above, herein we will present the JKO dicrete flow.
Again following~\citep{jordan1998variational}, the Fokker-Planck diffusion of distributions is recovered when minimizing entropy functional w.r.t. $2$-order Wasserstein metric $w^2$. Therefore, we can define our flow functional w.r.t. $w^2$ as
\begin{equation}
    \label{eq:jkoproximal}
    \mathcal{F}_{w^2}(\rho;\varrho) =  \frac{1}{2} w^{2}(\rho, \varrho) + \tau F(\rho).
\end{equation}
Starting from some fixed density $\rho_0$, similar to KL discrete flow in \eqref{eq:kldflow}, the solution of \eqref{eq:fp} can be approximated by the time-discrete sequence defined recursively by the \textit{JKO discrete flow} using $\mathcal{F}_{w^2}$:
\begin{equation}
    \label{eq:jkodflow}
     \rho_{t+1} = \operatorname*{argmin}_{\rho}\mathcal{F}_{w^2}(\rho;\rho_t), \ \ \forall t\geq 0, \quad \rho_0=\frac{1}{n}\sum^n_{i=1}q_{\mathbf{x}_0^i}.
\end{equation}
Jordan et al. also develop the well-known JKO scheme for numerically approximating the Fokker-Planck equation in discrete time~\citep{jordan1998variational}. According to \citep{gardiner1985handbook}, solutions of the Fokker-Planck equation \eqref{eq:fp} provide the probability density for empirical measure of the particles. These solutions are also inherently related to the SDE \eqref{eq:sde}. If $\Phi$ satisfies appropriate growth conditions, then the stationary solution of \eqref{eq:fp} is unique and it can be written in the form of the EBM distribution \eqref{eq:ebm}. However, Eq. \eqref{eq:fp} also brings $\mathcal{F}_{w^2}$ a property of high computation consumption due to $ \Delta \rho$, and  thus there still exists space for improvements. Based on these thoughts, we will present another new learning scheme of EBM in later paragraphs.

\subsection{Near-Optimal Discrete Flow}
\label{sec:methodderive}
Simply using $\mathcal{F}_{w^2}$ as a substitution of $\mathcal{F}_{\text{KL}}$ in practice is not advisable due to the static property of $w^2$. From our particle perspective, in this section, we apply the dynamic form of $w^2$, i.e., $W^2$, and introduce another near-optimal discrete flow that is more practical for approximating the optimal JKO discrete flow.

As specified in Sec.~\ref{sec:jkoflow}, both the KL discrete flow in Eq. \eqref{eq:kldflow} and JKO discrete flow in Eq. \eqref{eq:jkodflow} lead to the identical stationary distribution and the same SDE that describes the movements of the particles. Therefore, if the convergence condition of both $\mathcal{F}_{w^2}$ and $\mathcal{F}_{\text{KL}}$ are satisfied, it would be `equivalent' to optimizing either of them. We can prove this equivalence by the following theorem:
\begin{theorem}
  \label{thm:klkjolink}
  \textnormal{\citep[Theorem 2.2]{erbar2015large}} Fix $\rho_t\in\mathcal{P}(\mathbb{R}^d)$, we have the following formula in the sense of $\Gamma$-convergence.
  \begin{equation}
    \label{eq:gammaconverge}
  \begin{split}
        \min_{\rho\in\mathcal{T}(\rho_{t},\rho_{t+\tau})}\frac{1}{\tau} D_{\mathrm{KL}}(\rho\|\rho_t) - \frac{1}{4\tau}w^2(\rho_{t};\rho_{t+\tau}) \xrightarrow[\tau\rightarrow 0]{\Gamma} 
        \frac{1}{2}F(\rho_{t+\tau}) - \frac{1}{2} F(\rho_t).
    \end{split}
  \end{equation}
  
\end{theorem}
Based on the above theorem and by rearranging the terms with \eqref{eq:klproximal} and  \eqref{eq:jkoproximal}, we can reformulated \eqref{eq:gammaconverge} into
\begin{equation}
\label{eq:kljkolink}
    \mathcal{F}_{\text{KL}}(\rho_t;\rho_{t+\tau}) \approx \frac{1}{2}\mathcal{F}_{w^2}(\rho_t;\rho_{t+\tau})\quad \text{as}\ \tau\rightarrow 0.
\end{equation}
 The detailed deduction of this approximation is provided in Appendix \ref{append:gamma}. Eq.~\eqref{eq:kljkolink} reveals an appealing and previously-unexplored relationship between the KL discrete flow and the JKO discrete flow. Both $\mathcal{F}_{\text{KL}}$ and $\mathcal{F}_{w^2}$ describe a single time step of length $\tau$: $\mathcal{F}_{\text{KL}}$ characterizes the fluctuations of the particle system during the time length $\tau$, and $\mathcal{F}_{w^2}$ characterizes a single time step of length $\tau$ in the time-discrete approximation of Eq. \eqref{eq:fp}. Together with Theorem~\ref{coro:klflow} and Theorem~\ref{thm:klkjolink}, the JKO discrete flow $\mathcal{F}_{w^2}$ results in the same sampling-based learning scheme with the updating formula \eqref{eq:standard_learning} of $\mathcal{F}_{\text{KL}}$, if the time-step is infinitely small. 

 However, it is impractical for $\tau$ to be relatively small since we have to trade-off between time-cost and accuracy during sampling. Without this strict condition, the most significant benefit of interpreting $\mathcal{F}_{\text{KL}}$ into $\mathcal{F}_{w^2}$ is that we can make use of the dynamic characteristic of Wasserstein distance $w^2$ in Eq.~\eqref{eq:bb_wmetric}. As discussed in Sec.~\ref{sec:wd}, $w^2$ only depends on the optimal coupling of two distributions, and the minimum of $w^2$ is taken among all pairs of masses to find the optimal pair to form the only optimal trajectory. But there still exists the drawback that the larger $\tau$ yields a broader searching region, namely, the optimal trajectory will be highly biased. Compared with $w^2$, $W^2$ is driven by dynamical Lipschitz curves and many trajectories of transportation are allowable, as long as it is on the manifold. Since the velocity is bounded, a large $\tau$ will not affect the short-time behaviour of $\mathbf{x}$. The Benamou-Brenier form of the flow functional is written as
\begin{equation}
    \label{eq:bbjkoproximal}
    \begin{split}
    \mathcal{F}_{W^2} (\rho;\varrho) = \frac{1}{8}&W^{2}(\rho, \varrho) + \tau \tilde{F}(\rho, \varrho),\\
    \tilde{F}(\rho, \varrho) = F(\rho) & +  \frac{3}{8}(1-\alpha)\int_{\text{Lip}([0,1];\mathbb{R}^d)}\vert\nabla\Phi\vert^{2}\rho d\mathbf{x} \\
    & + \frac{3}{8}\alpha\int_{\text{Lip}([0,1];\mathbb{R}^d)}\vert\nabla\Phi\vert^{2}\varrho d\mathbf{x}, \quad 0 \leq\alpha\leq 1.
    \end{split}
\end{equation}
So far, we have already proved that solving Eq.~\eqref{eq:mle} is equally solving $\mathcal{F}_{\text{KL}}$ by Theorem~\ref{thm:1}, and $\mathcal{F}_{\text{KL}}\approx\mathcal{F}_{w^2}$ by Theorem~\ref{thm:klkjolink}, and next we will show that $\mathcal{F}_{w^2} \approx\mathcal{F}_{W^2}$.
\begin{claim}
  \label{clm:equal}
  $\mathcal{F}_{w^2} \approx\mathcal{F}_{W^2}$.
  \end{claim}
  \begin{proof}
  To show the above claim holds, we expand the geodesic Wasserstein metric $W^2\left(\rho_{t_0}, \rho_{t_1}\right)$ between $\rho_{t_0}$ and $\rho_{t_1}$ as follows:
  \begin{equation}
  \label{eq:interpolate}
  \begin{split}
    W^2(\rho_{t_0}, \rho_{t_1}) &:= \min \int_{\text{Lip}([0,1];\mathbb{R}^d)}\int_{t_0}^{t_1}|\nabla\Phi|^2 dt d\rho_{t}\\
    &\approx \left(t-t_{0}\right)\int_{\text{Lip}([0,1];\mathbb{R}^d)}\vert\nabla\Phi\vert^{2}d\rho_{t_{1}} +\left(t_{1}-t\right)\int_{\text{Lip}([0,1];\mathbb{R}^d)}\vert\nabla\Phi\vert^{2}d\rho_{t_{0}} \\
     &= \tau\left(1-\alpha\right)\int_{\text{Lip}([0,1];\mathbb{R}^d)}\vert\nabla\Phi\vert^{2}\rho_{t_{1}}d\mathbf{x}+\tau\alpha\int_{\text{Lip}([0,1];\mathbb{R}^d)}\vert\nabla\Phi\vert^{2}\rho_{t_{0}}d\mathbf{x},
    \end{split}
  \end{equation}
The second line of \eqref{eq:interpolate} uses the second mean value theorem for definite integrals, and thus we can approximate the integral via two randomly interpolated rectangles. Note that by \eqref{eq:interpolate}, the functional derivative of $W^{2}(\rho_{t_{1}}, \rho_{t_{0}})$ w.r.t. $\rho_{t_{1}}$ or $\rho_{t_{0}}$ is proportional to $|\nabla\Phi|^{2}$, namely
  \begin{equation}
      \label{eq:wderive}
    \frac{\delta W^{2}(\rho_{t_{0}}, \rho_{t_{1}})}{\delta\rho_{t_{1}}} \propto |\nabla\Phi|^{2},
  \end{equation}
  which is similar to the result of Proposition 8.5.6 in~\citep{ambrosio2008gradient}, implying that our expansion is valid. Next, substituting \eqref{eq:interpolate} into \eqref{eq:bbjkoproximal}, we have 
  \begin{equation*}
  \begin{split}
      \mathcal{F}_{W^2} &\approx \frac{1}{8}W^{2}(\rho, \varrho) + \tau\left(F(\rho)+\frac{3}{8\tau}W^{2}(\rho, \varrho)\right) \\
      &= \frac{1}{2}W^{2}(\rho, \varrho) + \tau F(\rho) \\
      &= \frac{1}{2}w^{2}(\rho, \varrho) + \tau F(\rho) = \mathcal{F}_{w^2}.
  \end{split}
  \end{equation*}
  \end{proof}

 Assuming $\Phi$ to be at least twice differentiable, by Eq.~\eqref{eq:wderive}, the partial functional derivative of $\tilde{F}(\rho;\varrho)$ w.r.t $\rho$ is
  \begin{equation}
      \label{eq:bbwderive}
      \tilde{\nu} = \nabla\frac{\delta\tilde{F}}{\delta\rho} = \nabla\log\rho + \nabla\Phi + \frac{3(1-\alpha)}{8}\nabla|\nabla\Phi|^2.
  \end{equation}
  We can treat $\tilde{F}$ in \eqref{eq:bbjkoproximal} as the variational condition of the continuity equation, substitute \eqref{eq:bbwderive} into \eqref{eq:continuity}, and then obtain a new Fokker-Planck equation that considers the evolutionary density to be in a set of Lipschitz curves:
  \begin{equation}
    \label{eq:new_fokker}
     \begin{split}
    \partial_{t} \rho = 
    \div\left(\rho \left(\nabla \Phi +   \frac{3(1-\alpha)}{8}\nabla \vert \nabla \Phi \vert^{2}\right)\right) + \Delta \rho,\quad \text{s.t. }|\nabla\Phi|\leq L.
    \end{split}
  \end{equation}
  In other words, if the neural energy function $\Phi$ defined in \eqref{eq:neuralphi} is $L$-Lipschitz continuous, the corresponding optimal transport density curve will be the $L$-Lipshcitz curve.  

  \subsection{Proximal Numerical Learning and Sampling Scheme}
  We have shown that solving the Benamou-Brenier JKO discrete flow $\mathcal{F}_{W^2}$ yields a near-optimal solution to the problem defined in Sec.~\ref{sec:setup}, owing to the Lipschitz characterization. Thus, till now we only need to show the proximal sampling scheme of the proposed gradient flow. 
  
  We first unfold the SDE related with \eqref{eq:new_fokker} to
  \begin{equation}
    \label{eq:new_sde}
    d\mathbf{x}_{t} = -\nabla{\Phi(\mathbf{x}_{t})}dt - \frac{3(1-\alpha)}{8} \nabla\vert\nabla\Phi(\mathbf{x}_{t})\vert^{2}dt + \sqrt{2}d\mathcal{W}_t,
  \end{equation}  
  where $\mathcal{W}_t$ stands for the standard  Wiener process. This equation implies that the discrete time sampling involves the calculation of the second derivative; therefore, the computation consumption will be huge especially when the network is deep, not to mention that the sampling is time-consuming. Thus, we expect to find a proper approximation of \eqref{eq:new_sde}. Recall that Eq.~\eqref{eq:sgld}, if the stochastic gradient descent $\nabla\Phi$ serves as an unbiased estimator, then we can remove noise part $\mathbf{z}$ in the SGD iteration and write it in a simpler form, 
  \begin{equation}
      \label{eq:sgd}
      \mathbf{x}_{t+1} = \mathbf{x}_{t} - \nabla\Phi(\mathbf{x}_{t}).
  \end{equation}
  The following theorem from \citep{li2017stochastic} enables us to approximate \eqref{eq:sgd} by \eqref{eq:new_sde}.
  \begin{theorem}
  \label{thm:approxsgd}
    \textnormal{\citep[Theorem 1]{li2017stochastic}} Assume $\Phi$ is Lipschitz continuous, at least twice differentiable and has at most linear asymptotic growth. Then,
      \begin{enumerate}
          \item[(i)] The stochastic process $\mathbf{x}_t$ satisfying SGLD \eqref{eq:sde} is an order 1 weak approximation of SGD in \eqref{eq:sgd}.
          \item[(ii)] The stochastic process $\mathbf{x}_t$ satisfying
          \begin{equation}
              \label{eq:new_sde_alpha2}
              d\mathbf{x}_{t} = -\nabla{\Phi(\mathbf{x}_{t})}dt - \frac{1}{4} \nabla\vert\nabla\Phi(\mathbf{x}_{t})\vert^{2}dt + \sqrt{2}d\mathcal{W}_t.
          \end{equation}
          is a 2-order weak approximation of SGD in \eqref{eq:sgd}.
      \end{enumerate}
  \end{theorem}

  
  
  
       

Theorem~\ref{thm:approxsgd} reveals that SGD implicitly minimizes Benamou-Brenier JKO discrete flow with $2$-order approximation, if $\alpha=\frac{1}{3}$ in \eqref{eq:new_sde}. The second-order approximation, known as the convexity correction, is important for the equilibrium functions, and always yields the better solution that the first-order approximates. Besides, employing SGD directly in the conventional EBM learning and sampling scheme (Eq.~\eqref{eq:standard_learning}) is improper, since there is no Lipschitz continuous restriction to let the SGLD satisfy the condition (i) in Theorem~\ref{thm:approxsgd}. Similar to the parameterized KL flow $\mathcal{F}_{\text{KL}}^{l}$ defined in Eq.~\eqref{eq:klflow_lag}, we propose a parameterized Benamou-Brenier JKO flow $\mathcal{F}_{W^2}^{l}$ with the $L$-Lipshcitz constraint:
\begin{equation}
  \label{eq:parajkoflow}
  \begin{split}
      \mathcal{F}_{W^2}^{l}(\rho;\rho_t) =  \operatorname*{min}_{\rho}\operatorname*{max}_{\theta} \Bigg\{W^2(\rho,\rho_t)  &+ \mathbb{E}_{p_{\text{data}}}[\Phi_\theta] - \mathbb{E}_{\rho}[\Phi_\theta] 
       - \mathcal{H}(\rho) \\
       &+\frac{\tau}{8}\mathbb{E}_{\rho_t}\left[\Big||\nabla\Phi_{\theta}|-L\Big|^2\right] 
      +\frac{\tau}{4}\mathbb{E}_{\rho}\left[\Big||\nabla\Phi_{\theta}|-L\Big|^2\right] \Bigg\}.
\end{split}
\end{equation}
To satisfy the Lipschitz continuous condition (assumed in Eq. \eqref{eq:new_fokker}), $\mathcal{F}^l_{W^2}$ has been added up with two $l_2$ normalization items, ensuring that $|\nabla\Phi|\leq L$.
Described by $\mathcal{F}_{W^2}^{l}$, the dynamics of the empirical measure $\rho_t$ is minimized and approximated through the particles driven by the SGD. $\theta$ is updated by calculating $\nabla_{\theta}\mathcal{F}_{W^2}^{l}$, which involves the consecutive time particles. At last, we conclude the proximal numerical learning and sampling scheme which is summarized as cross-updating formulations as follows:
\begin{equation}
  \label{eq:new_learning}
  \left\{
  \begin{aligned}
      \mathbf{x}_{t+1} & = \mathbf{x}_{t} - \nabla_{\mathbf{x}_t}\Phi_{\theta_t}(\mathbf{x}_{t}); \\
      \theta_{t+1} & = \theta_{t} +  \nabla_{\theta_t}\mathbb{E}_{\mathbf{X}\sim p_{\text{data}}}[\Phi_{\theta_t}(\mathbf{X})] 
       - \nabla_{\theta_t}\mathbb{E}_{\mathbf{x}_{t+1}\sim\rho_{t+1}}[\Phi_{\theta_t}(\mathbf{x}_{t+1})] \\
      & + \frac{\tau}{8}\nabla_{\theta_t}\mathbb{E}_{\mathbf{x}_t\sim\rho_{t}}\left[\Big||\nabla_{\mathbf{x}_t}\Phi_{\theta_t}(\mathbf{x}_t)|-L\Big|^{2}\right] 
       + \frac{\tau}{4}\nabla_{\theta_t}\mathbb{E}_{\mathbf{x}_{t+1}\sim\rho_{t+1}}\left[\Big||\nabla_{\mathbf{x}_{t+1}}\Phi_{\theta_t}(\mathbf{x}_{t+1})|-L\Big|^{2}\right].
  \end{aligned}
  \right.
\end{equation}

\begin{algorithm}[t]
  \renewcommand{\algorithmicrequire}{\textbf{Input:}}
  \renewcommand{\algorithmicensure}{\textbf{Output:}}
 \caption{Our Near-Optimal Learning and Sampling Scheme}
 \label{alg:OT-EBM}
 \begin{algorithmic}[1]
     \REQUIRE  Observed data $\{\mathbf{X}\}$;\ the neural network $\Phi_{\theta_0}$; \ sampling step-size $K$; \ learning time length $T$
     \ENSURE Sampled data $\{\mathbf{x}\}$; learned parameter $\theta$
     \STATE  Initialize $\theta_0$ and $\mathbf{x}_{0} \leftarrow \mathcal{N}(0,I)$
     \FOR{$t=0$ {\bfseries to} $T-1$}
     \STATE Update $\mathbf{x}_{t+1}$ through $K$ times SGD sampling in \eqref{eq:new_learning} using $\mathbf{x}_{t}$ and $\Phi_{\theta_t}$
     \STATE Update $\theta_{t+1}$ through \eqref{eq:new_learning} using $\mathbf{x}_{t}$, $\mathbf{x}_{t+1}$, $\mathbf{X}$ and $\Phi_{\theta_t}$, based on Adam optimizer
     \ENDFOR
 \end{algorithmic}
\end{algorithm}

\noindent\textbf{Summary}. We name the EBM learned by Eq.~\eqref{eq:new_learning} as OT-EBM, considering its OT property.  Given the entire deduction above, we provide an overview of the whole deduction procedure in Figure~\ref{fig:deducing}. The algorithm details are provided in Algorithm~\ref{alg:OT-EBM}. Eq. \eqref{eq:new_learning} indicates that the gradient of $\theta$ determined in the Wasserstein manner is being added with some soft gradient norm constraints between the last two iterations. Such gradients norm has the following advantages compared with the original iterative process Eq.~\eqref{eq:standard_learning}. 
\begin{enumerate}
  \item[(i)] The norm $|\nabla_{\mathbf{x}}\Phi_{\theta}(\mathbf{x})|^2$ serves as the constant speed geodesic connecting $\rho_{t}$ and $\rho_{t+1}$ in the manifold spanned by $p_{\theta}$ and $p_{\text{data}}$, to make particles transported optimally.
  \item[(ii)] The norm can also be interpreted as a soft force counteracting the original gradient and prevents the entire learning process from an explosion.
  \item[(iii)] Compared to the original iterative process (Eq.~\eqref{eq:standard_learning})                    , which only looks one step ahead, OT-EBM is a two-step lookahead algorithm, which encourages a particularly smooth form of transportation and promises a performance improvement. 
  \item[(iv)] The $L$-Lipschitz continuity of neural energy $\Phi$ is seamlessly incorporated into the learning process of $\theta$. Namely, the risk of model collapse is reduced.
  \item[(v)] The updating iteration of $\mathbf{x}$ is now deterministic and refrains from random noises, and can be conveniently carried on ready-made SGD solvers.
\end{enumerate}

\begin{figure*}[t]
    \centering
  \subfigure[Gaussian Mixture Modeling]{
  \begin{minipage}[t]{\linewidth}
  \centering
  \includegraphics[width=15.5cm]{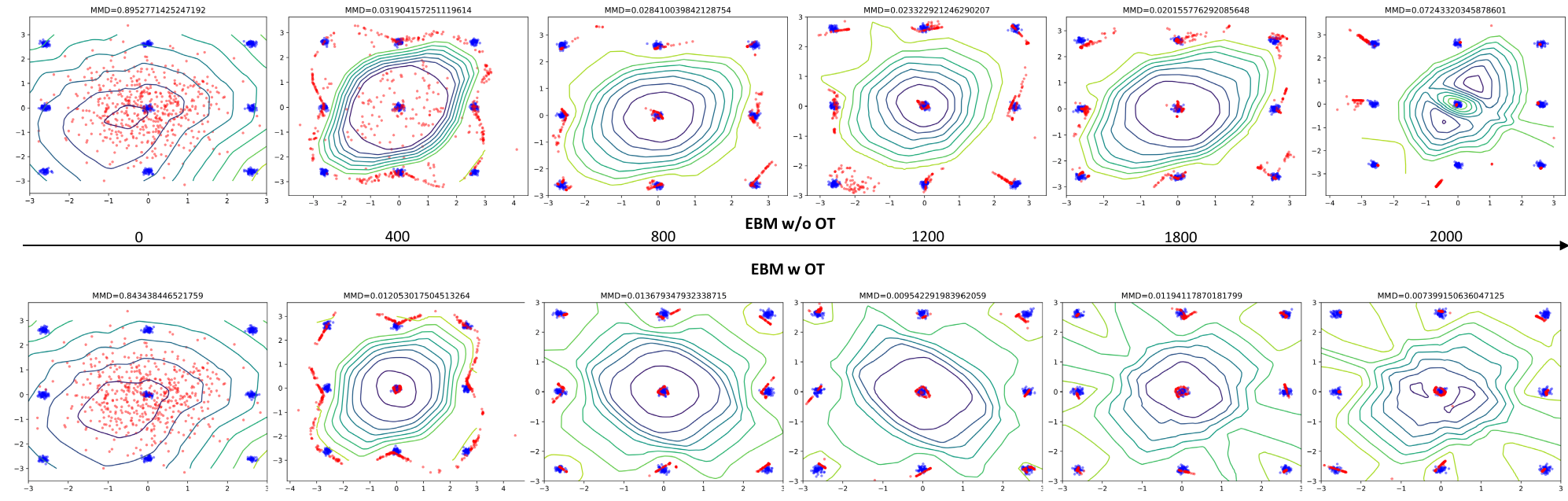}
  \end{minipage}
  }
  \subfigure[Swissroll Modeling]{
  \begin{minipage}[t]{\linewidth}
  \centering
  \includegraphics[width=15.5cm]{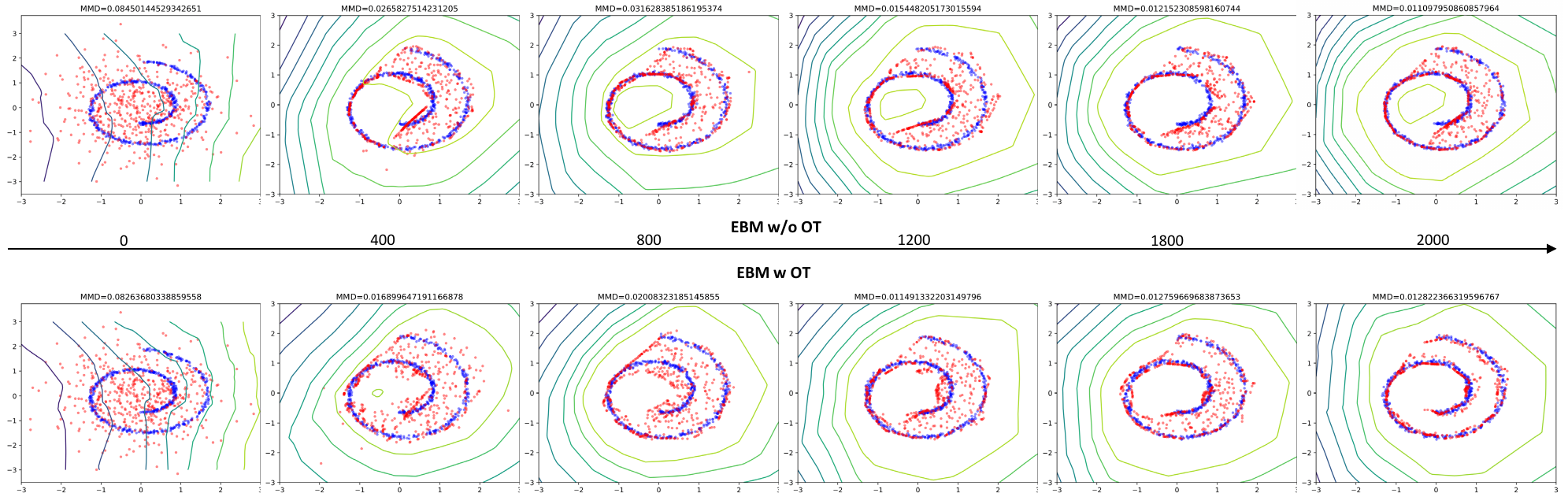}
  \end{minipage}
  }
    \caption{The visualization on the learned manifolds transforming by iteration. Larger MMD means inferior fitting. The red spots are from the generated distribution, and the blue ones conform to the target distribution. The iridescent lines stand for the output neural function contours. \label{fig:toyexmaple_gaussian}}
    \vspace{-30pt}
\end{figure*}


\section{Empirical Results}
\label{sec:experiment}
We evaluate the proposed method from following aspects:
\begin{itemize}
    \item Comparing the performance of EBM and OT-EBM on artificially generated samples that follow certain given distributions.
    \item Comparing EBM and OT-EBM on generation with few samples in high dimensions. In this part we also provide a comparison on EBMs and the popular GAN model.
    \item Expanding the real distribution to the large scale, where we apply both EBMs on widely used human face and classification datasets.
    \item Exploring OT-EBM on some generation tasks, such as in-painting, high-resolution picture, and even graph inference.
\end{itemize}
\subsection{Toy Example Modeling}
\label{sbsec:toyexample}
We first evaluate OT-EBM and EBM by artificially designed toy examples. The real data are generated from a 2-dimensional Gaussian mixture model (GMM) and a Swissroll distribution in aggregation. 
As shown in Figure~\ref{fig:toyexmaple_gaussian} that displays the whole process of sampling and estimation, our goal is to transport red particles to the ideal sample locations (blue) as close as possible by modeling them with a neural network~(energy function) $\Phi_{\theta}$. One thing that matters is the network should learn a valid vector field so that it can drive arbitrarily located particles to their next-step ideal state in the data space. Both EBM and OT-EBM are with similar setups, including the neural network structure and hyperparameters. The network is composed of five fully connected layers (with bias weighted). The learning rate of network parameter $\theta$ is $10^{-3}$, that of dynamical particles $\mathbf{x}$ is $1.0$, $\tau$ is set to $20$ and the optimizer is Adam~\citep{kingma2014adam}, and the Lipschitz constraint $L$ is $1.0$ by default.\\

\begin{figure*}[t]
  \subfigure[\textbf{w/o OT} fitting result]{
    \begin{minipage}[t]{0.3\linewidth}
    \centering
    \includegraphics[width=5cm]{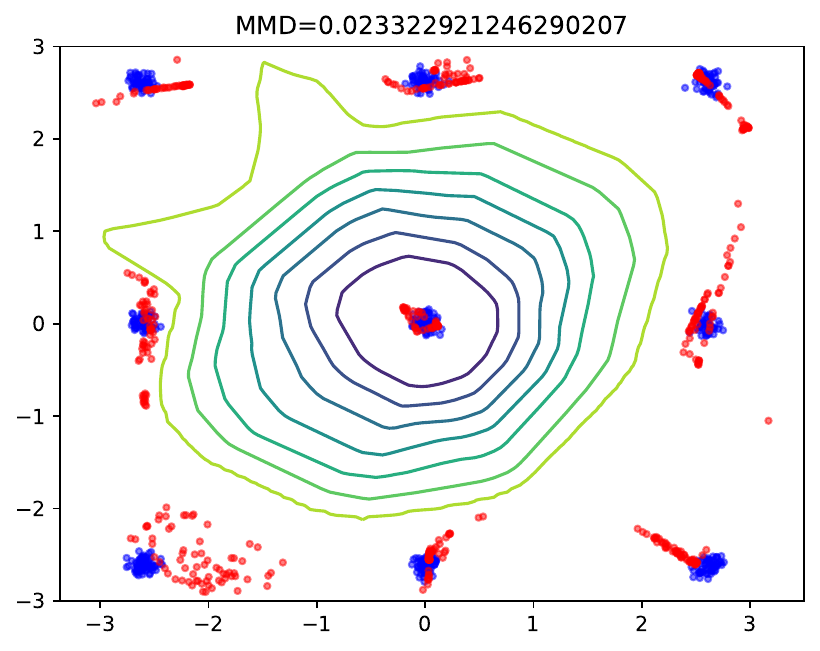}
    \end{minipage}
    }
  \subfigure[gradient quivers]{
    \begin{minipage}[t]{0.3\linewidth}
    \centering
    \includegraphics[width=5cm]{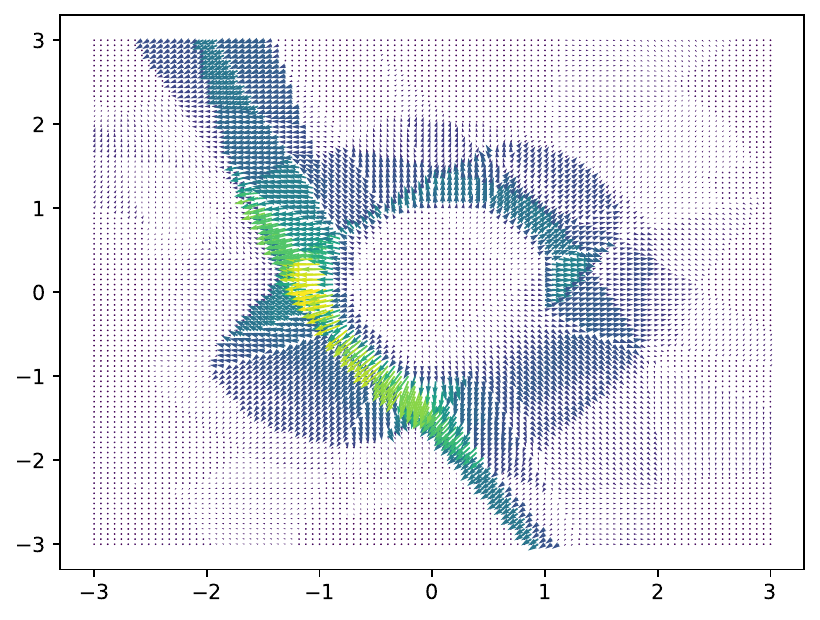}
    \end{minipage}
  }
  \subfigure[weight analysis]{
    \begin{minipage}[t]{0.3\linewidth}
    \centering
    \includegraphics[width=5cm]{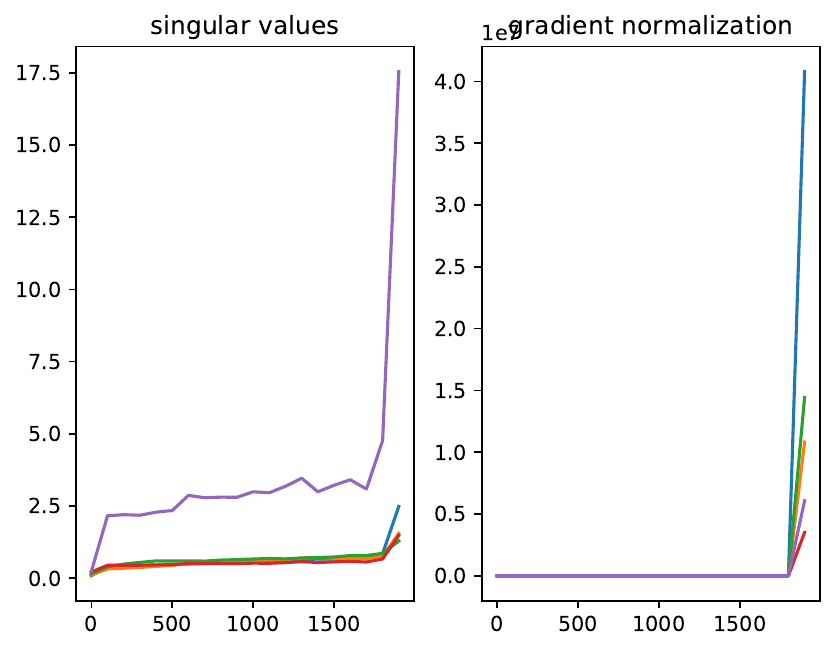}
    \end{minipage}
  }
  \subfigure[\textbf{w OT} fitting results]{
    \begin{minipage}[t]{0.3\linewidth}
    \centering
    \includegraphics[width=5cm]{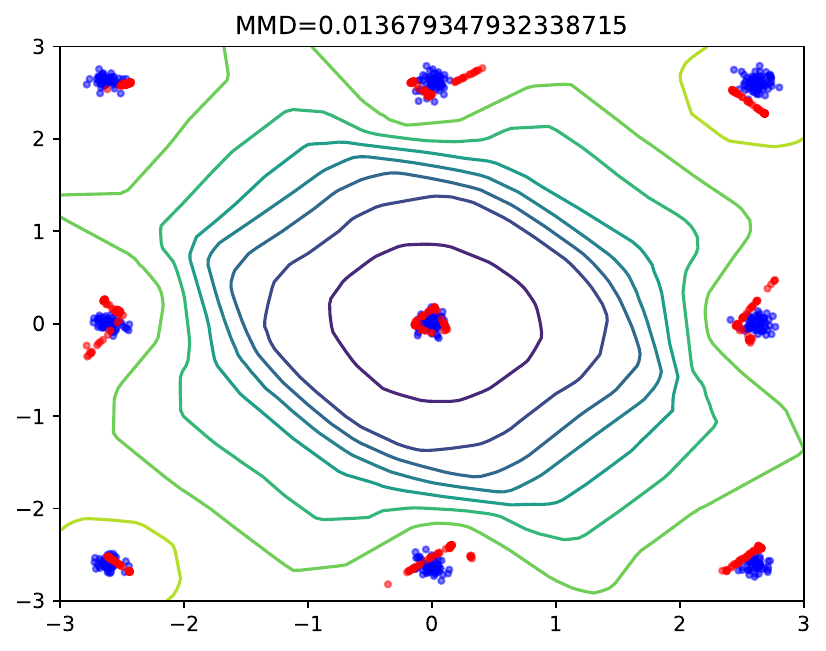}
    \end{minipage}
    }
  \subfigure[gradient quivers]{
    \begin{minipage}[t]{0.3\linewidth}
    \centering
    \includegraphics[width=5cm]{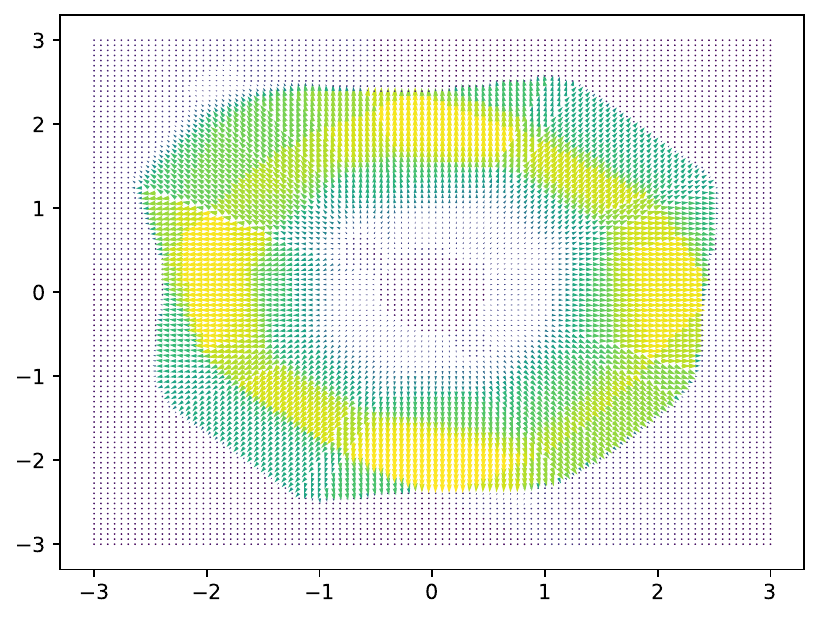}
    \end{minipage}
  }
  \subfigure[weight analysis]{
    \begin{minipage}[t]{0.3\linewidth}
    \centering
    \includegraphics[width=5.5cm]{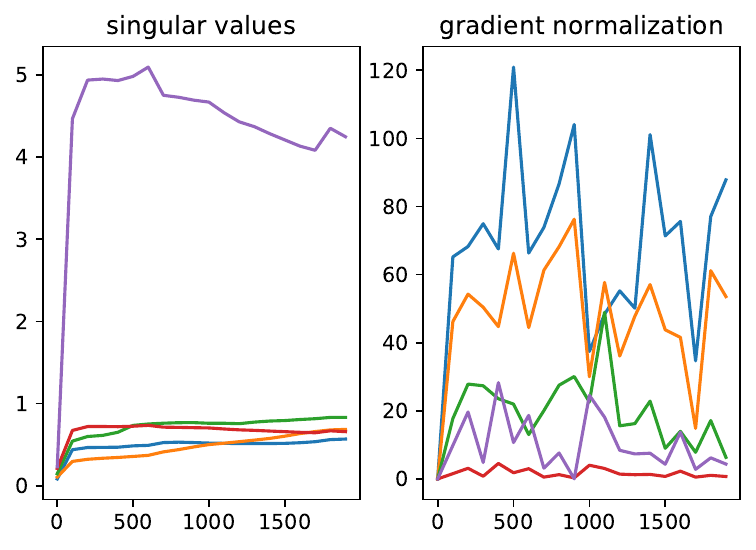}
    \end{minipage}
  }
  \caption{Toy example modeling results on Gaussian mixture model. The method without OT (\textbf{w/o OT}) has collapsed after 1800 iterations of training, which can be identified through the weight analysis (c). The best fitting results and the network output contour is selected from 2000 iterations of learning and synthesizing (in (a) \textbf{w OT} is the result of iteration 1760, while \textbf{w/o OT} from 1880 in (d)). Lines in different colors in (c) and (f) stand for gradient normalization of 5 different layers in EBM. \label{fig:toyexample_gaussian_ebm}}
  \vspace{-0.5cm}
  \end{figure*}
  \begin{figure*}[t]
    \subfigure[\textbf{w/o OT} best fitting]{
      \begin{minipage}[t]{0.3\linewidth}
      \centering
      \includegraphics[width=5cm]{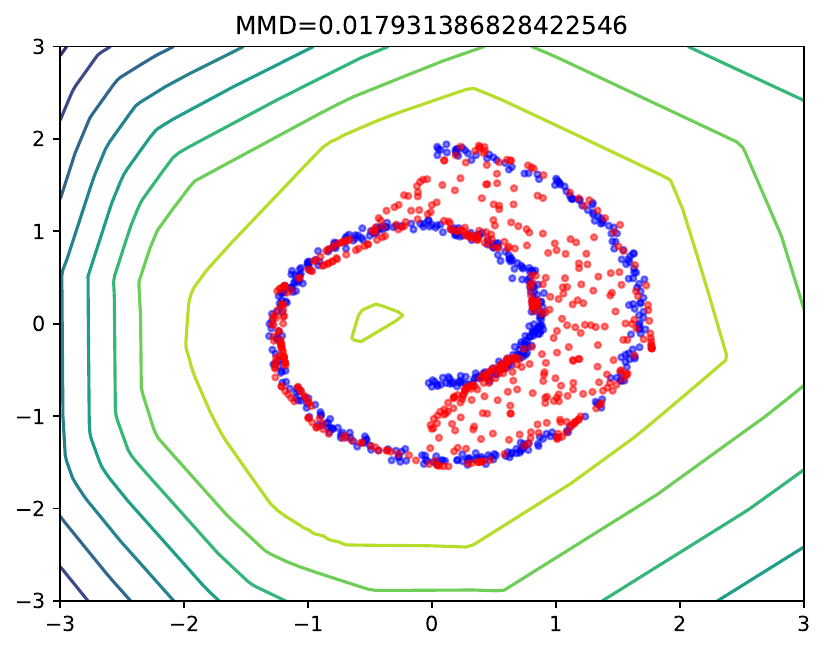}
      \end{minipage}
      }
    \subfigure[gradient quivers]{
      \begin{minipage}[t]{0.3\linewidth}
      \centering
      \includegraphics[width=5cm]{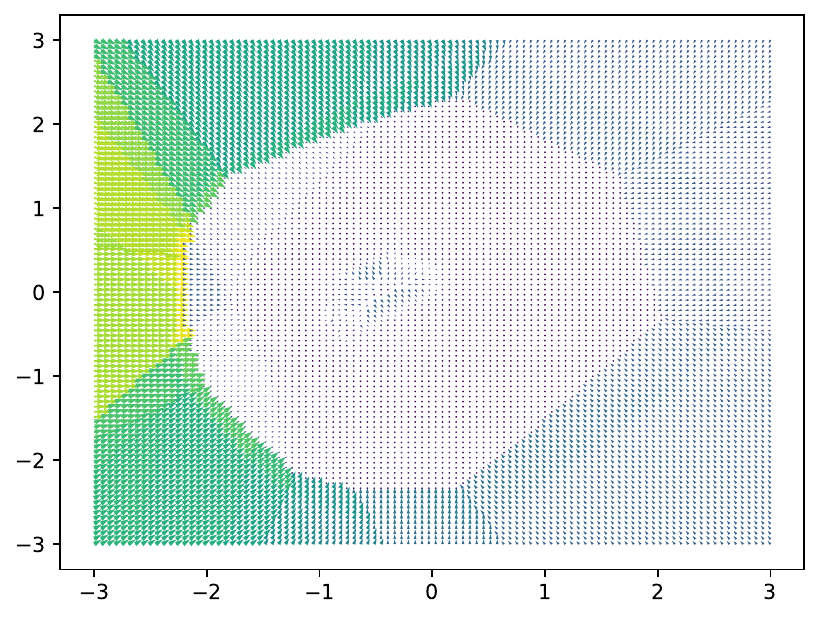}
      \end{minipage}
    }
    \subfigure[weight analysis]{
      \begin{minipage}[t]{0.3\linewidth}
      \centering
      \includegraphics[width=5cm]{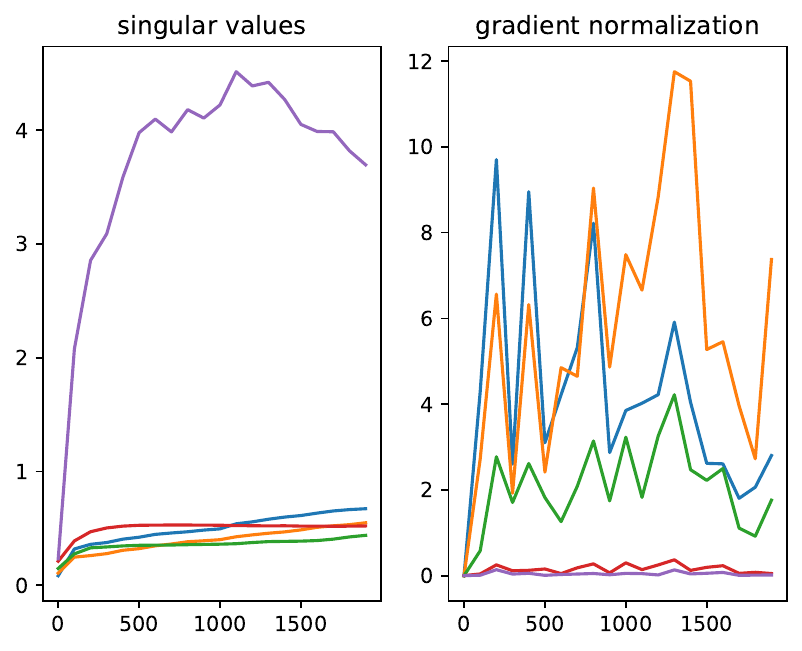}
      \end{minipage}
    }
    \subfigure[\textbf{w OT} best fitting]{
      \begin{minipage}[t]{0.302\linewidth}
      \centering
      \includegraphics[width=5cm]{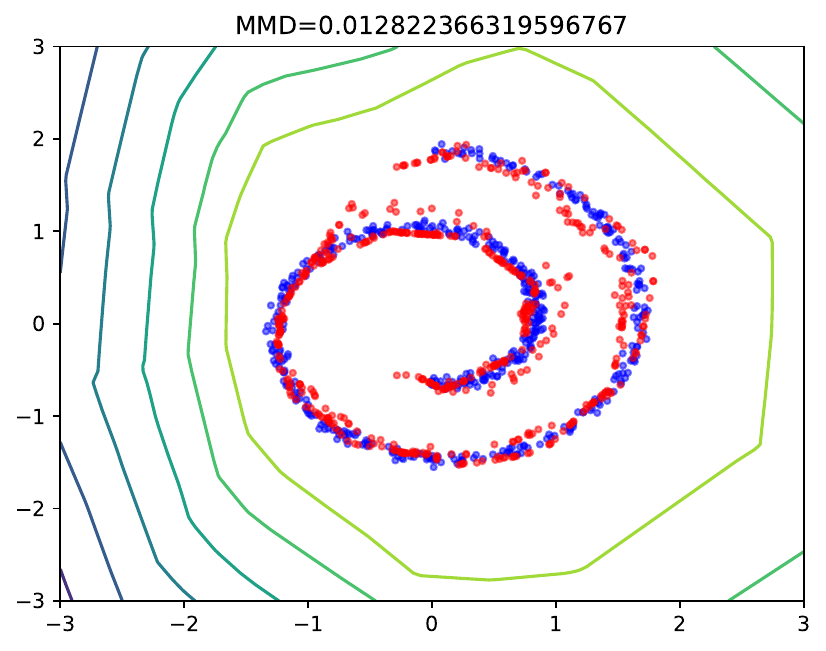}
      \end{minipage}
      }
    \subfigure[gradient quivers]{
      \begin{minipage}[t]{0.302\linewidth}
      \centering
      \includegraphics[width=5cm]{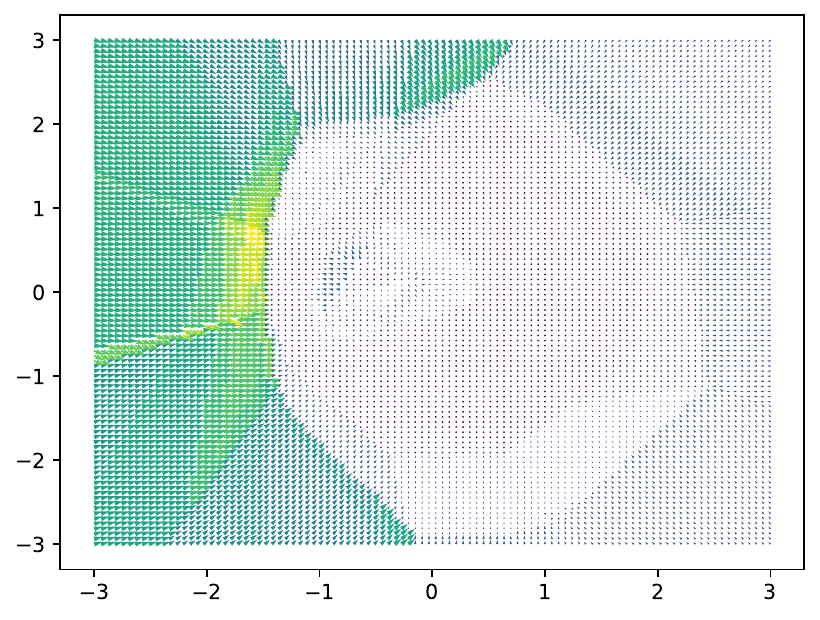}
      \end{minipage}
    }
    \subfigure[weight analysis]{
      \begin{minipage}[t]{0.35\linewidth}
      \centering
      \includegraphics[width=5.1cm]{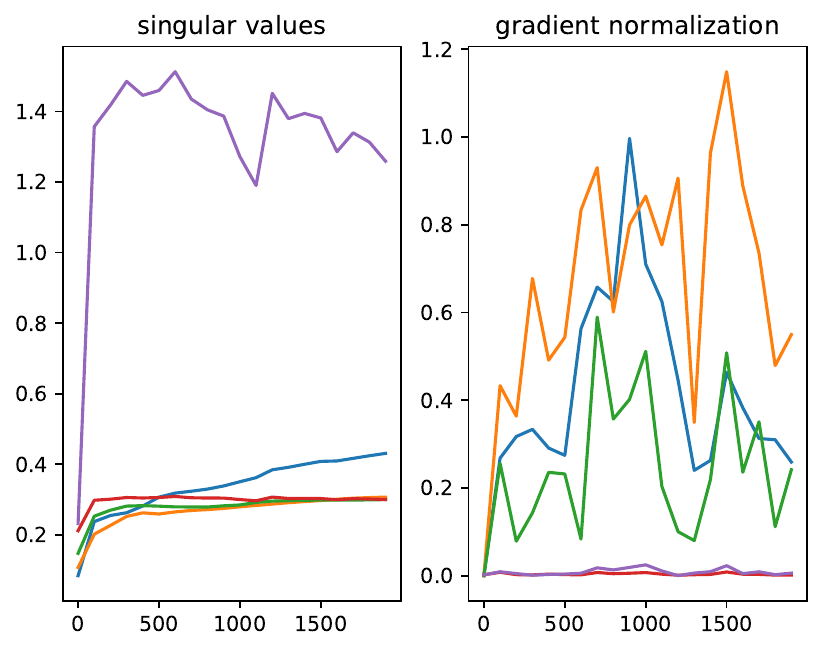}
      \end{minipage}
    }
    \caption{Toy example modeling results of Swissroll. The picked iteration index of subfigure (a) and (d) are respectively 1960 and 1840.\label{fig:toyexample_swissroll_ebm}}
    \end{figure*}
\noindent\textbf{Better Generalization and Faster Convergence} \\
The learned model data manifold or vector field reflects the model generalization ability. Thus we provide the vector field contours of the estimation on 2D 9-Gaussian mixture model in Figure~\ref{fig:toyexmaple_gaussian}(a), where the contours visualize the learned data manifold. The selected fitting results are the best over all learning epochs. We also include gradient quiver illustration ((b)(f) of Figure~\ref{fig:toyexample_gaussian_ebm} and  \ref{fig:toyexample_swissroll_ebm}) and the weight analysis ((c)(f) of Figure~\ref{fig:toyexample_gaussian_ebm} and  \ref{fig:toyexample_swissroll_ebm}) of the whole learning process.
In quiver illustrations, the arrows with the larger length indicate a greater pushing force.

\begin{wraptable}{c}{0.45\textwidth}
\begin{minipage}{\linewidth}
    \centering
    \begin{tabular}{ccc}
    \toprule
         Model& GMM &Swissroll  \\
         \midrule
         EBM& 82.2& 78.1 \\
         OT-EBM&89.6& 85.3 \\
         \bottomrule
    \end{tabular}
\end{minipage}
\caption{The accuracy (\%) of two compared methods on recovery modes in toy examples. \label{tab:recovery}}
\vspace{-20pt}
 \end{wraptable}
 
According to the presented results of Figure~\ref{fig:toyexmaple_gaussian}, we can conclude that {\em OT-EBM converges faster than EBM}: at the same learning time point, the transporting particles of OT-EBM are fitted closer than those from EBM; {\em OT-EBM learns a better data manifold}: at each slide of Figure~\ref{fig:toyexmaple_gaussian}, OT-EBM demonstrates the better-learned data manifolds. Particularly, in sub-figure (b), the Swissroll contours of OT-EBM manifest a concave at the left upper part, which results in the particles being able to correctly transfer to around the contour of the observed locations. Besides, OT-EBM visually achieves better fitting on both Gaussian mixture and Swissroll distributions with much lower mean maximum discrepancy (MMD), measuring lower divergence between the estimated distribution and the observed distribution. We also compare the percentage of recovery modes, which is listed in Table~\ref{tab:recovery}.\\

\noindent\textbf{Local Minimal and Vector Field}\\
\noindent As shown in Figure \ref{fig:toyexample_gaussian_ebm}(a), the sample particles are distributed in chaos, the weights and gradients (c) boom at this epoch. While OT-EBM maintains a steady improvement during the learning. When transporting a large amount of particles~(in this experiment we set 200) to learn Swissroll distribution, the data samples from EBM are trapped at a poor local minimal and failed to jump out, as shown in Figure~\ref{fig:toyexample_swissroll_ebm}(a). Together with Figure~\ref{fig:toyexample_swissroll_ebm}(b), without OT, lengths of gradient arrows are constant, especially those distributed around the output manifold; therefore, the samples in the gap sense no force and stay in this location. The gradients of EBM with OT method in the Figure~\ref{fig:toyexample_swissroll_ebm}(e) are much better and reasonably distributed along the manifold, where the particles are pushed to the exact locations by the gradient force. The above observations can draw the conclusion that OT-EBM not only fits random particles closer to the observed samples, but also learns a better vector field in terms of direction and magnitude.\\  
\\

\noindent\textbf{Better Learned Weights and Model Collapse} \\
To verify that the neural energy function in OT-EBM is $1$-Lipschitz continuous, we calculate the gradient norm of learned parameters for different layers \emph{w.r.t} all steps in the learning process. The statistics are shown in Figure~\ref{fig:toyexample_gaussian_ebm} and Figure~\ref{fig:toyexample_swissroll_ebm}(c)(f). With identical learning environments, greater gradient usually means the much unstable learning and being easy to collapse. We measure the singular values and the gradient normalization as the basis of weight analysis. In the Gaussian mixture modeling, the model collapses at the 1890-th iteration, which is manifest in Figure~\ref{fig:toyexample_gaussian_ebm}(c) that the weight matrix singular values and the gradient normalization boom at the time point. Compared with original EBM, in Figure~\ref{fig:toyexample_gaussian_ebm}(f), the singular values and the gradient normalization value of OT-EBM maintain stable during the whole learning process. 

\subsection{Few Sample Real Image Modeling}
Next we verify the effectiveness of OT on real images by implementing EBM and OT-EBM on several widely used datasets, including SUN~\citep{xiao2010sun}, CIFAR-10~\citep{krizhevsky2009learning}, and CelebA~\citep{liu2015deep}. Real images commonly have higher dimensions, and thus the neural energy $\Phi_{\theta}$ is formulated into the deep convolution neural network (CNN)\citep{krizhevsky2012imagenet} with a larger parameter capacity to capture non-linear visual features, namely $h(x)$ in $\Phi_{\theta}$ consists of convolutional component operators such as down-sampling and convolution operators. To fully compare EBM and OT-EBM, we start by modeling on small scale images sets to reduce the potential negative effects of dataset size, and then evaluate larger scale datasets, which pose a greater model generalization challenge.\\

 
\noindent\textbf{Robustness Comparison}\\
We use a VGG-16~\citep{simonyan2014very} to model images selected from the SUN dataset and resized to 64$\times$64 pixels. For a fair comparison, both EBM and OT-EBM are under the identical neural network and hyperparameter settings.
\begin{figure}
     \begin{minipage}[t]{0.55\linewidth}
    \centering
    \subfigure[EBM]{
      \begin{minipage}{0.45\linewidth}
      \centering
      \includegraphics[width=3.4cm]{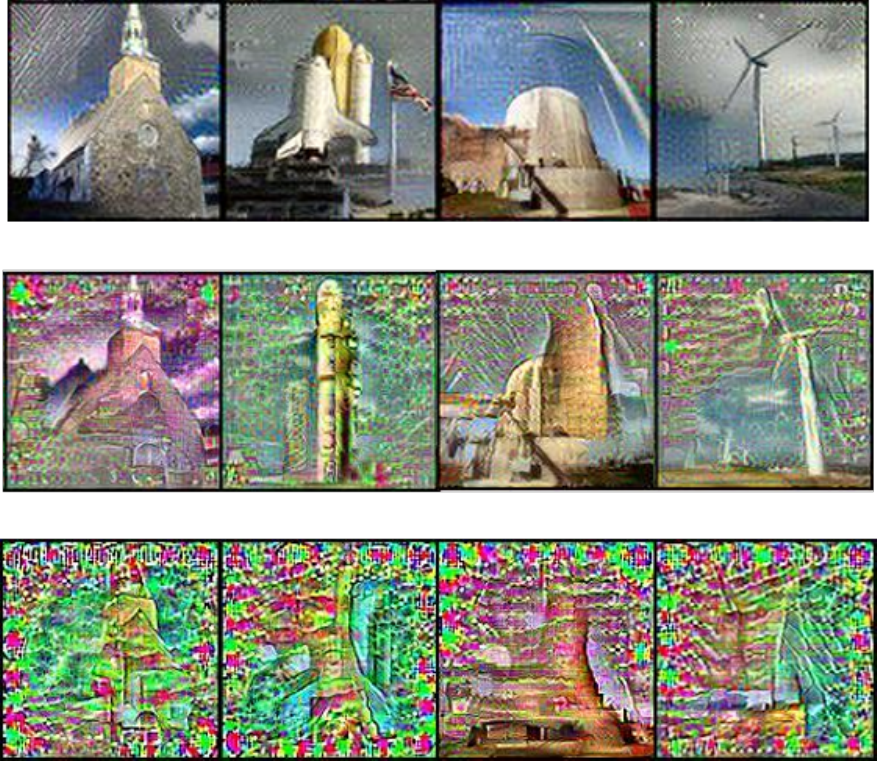}
      \end{minipage}
      }
    \subfigure[OT-EBM]{
      \begin{minipage}{0.45\linewidth}
      \centering
      \includegraphics[width=3.4cm]{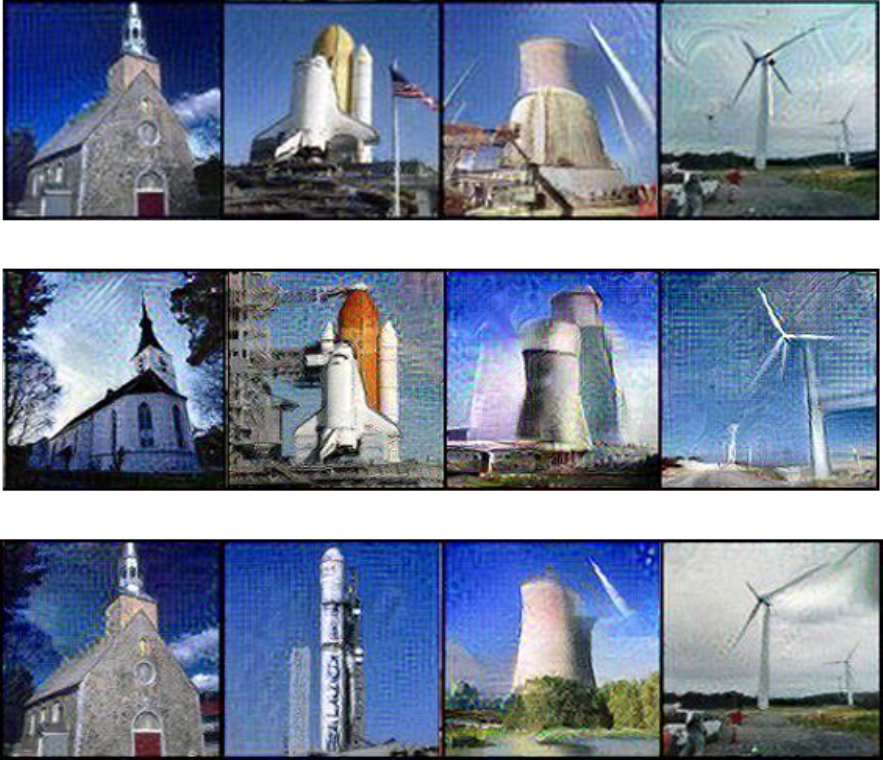}
      \end{minipage}
    }
    \caption{Learning stability improvement on high dimensional image modeling. Images in the first row of each group are generated with the default learning rate 0.112. The second and third row images are generated with the learning rates 0.3 and 0.5 respectively. \label{fig:imagerobustness}}
  \end{minipage}
  \begin{minipage}[t]{0.4\linewidth}
  \begin{center}
   \vspace{-5pt}
    \includegraphics[width=\textwidth, height=3.8cm]{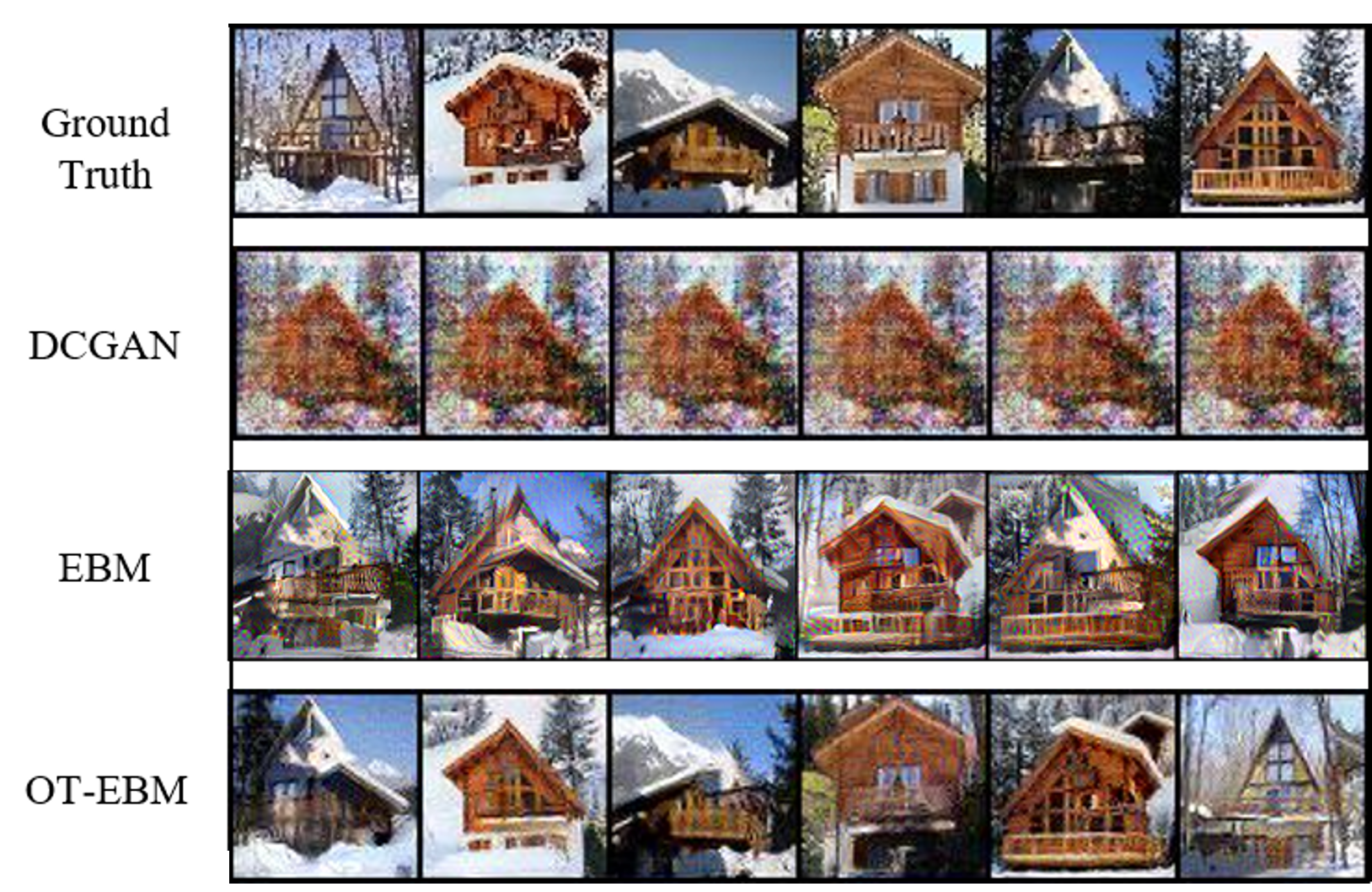}
  \end{center}
  \vspace{-16pt}
  \caption{Few sample learning comparison on dataset SUN, we choose 6 images from scene category ``cabins'' as inputs, shown in the Ground-Truth row. \label{fig:few_sample}}
  \end{minipage}
\end{figure}
 
 According to the results shown in Figure~\ref{fig:imagerobustness}, our method is robust in terms of being adaptable to high dimensional data and adventurous hyperparameters tuning, indicating that OT-EBM is more practical that can reduce the workload of hyperparameter adjustments. 
Specifically, in the first row of Figure~\ref{fig:imagerobustness}, we apply both groups with default settings: the learning rate of $\theta$ is $0.001$ and the learning rate of sampling $\mathbf{x}$ is $0.112$.
In the second row, we slightly increase the learning rate of $\theta$ to $0.01$ and find that the quality of both methods has decreased but the deterioration of the compared group with OT is lighter than the original EBM. In the third row, compared with the default settings, we increase the learning rate of $\mathbf{x}$ to $0.3$ and EBM completely collapses  while OT-EBM still generates relatively high quality images.
\\

\noindent\textbf{Comparison with GAN-based Model}\\
Different from recent popular generative models such as GANs, which rely on a large number of data to acquire knowledge of hidden space layouts, the energy-based models are naturally built with the learning from a few samples. This is mainly due to that before the suggestion of stochastic approximation by Robbins and Monro~\citep{robbins1951stochastic}, the estimation of the empirical mean of sampling-based methods has to traverse the whole dataset. This means that smaller datasets will lead to more efficient and accurate estimates. Although recently there are researches that deal with the single-image generation using GANs, such as~\citep{shocher2018zero,gandelsman2019double}, but they require the further model constructing.

While the generation with few samples can be applied to all sorts of EBMs with superior good performance. To verify that OT also improves EBM on this task, we compare a typical GAN model, the DCGAN~\citep{radford2015unsupervised}, with EBM and OT-EBM on 6 images randomly selected from the SUN dataset.
For a fair comparison, we duplicate the input images several times to a total of 10,000 to match the similar training setting of DCGAN. All the displayed generated images are the best we achieve on each method in Figure~\ref{fig:few_sample}, where DCGAN is already in its stable state and has overfitted the input data. The learning of both EBM and OT-EBM starts from a Gaussian noise $q$ initialization and generatively learns the neural energy. It is noted that OT-EBM yields more clear images than EBM, since the sample dynamics of OT-EBM have the determined force.

\subsection{Image Generation on Large Scale Datasets}
Next we evaluate the OT-EBM models with larger datasets (on both size and quantity) for further measuring the advantages of the OT learning strategy.
\begin{figure}[t]
  \centering
  \subfigure[]{
    \begin{minipage}[t]{0.1\linewidth}
    \centering
    \includegraphics[width=1.04cm]{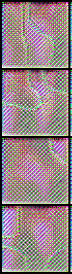}
  \end{minipage}
  }
  \subfigure[]{
  \begin{minipage}[t]{0.25\linewidth}
  \centering
  \includegraphics[width=4cm]{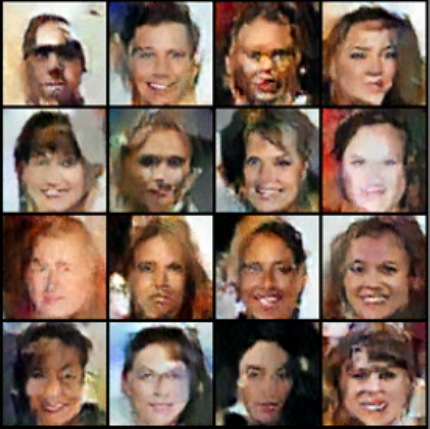}
  \end{minipage}
  }
  \subfigure[]{
  \begin{minipage}[t]{0.25\linewidth}
  \centering
 \includegraphics[width=4cm]{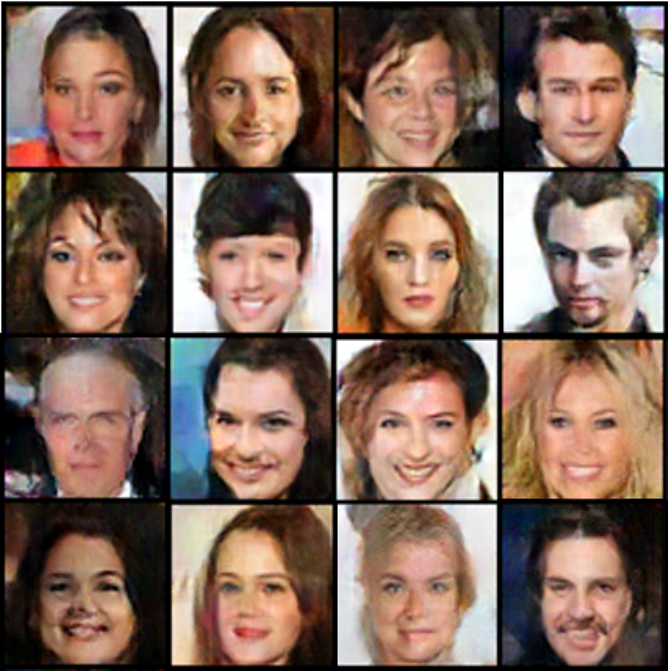}
  \end{minipage}
  }
  \subfigure[Learning curves]{
  \begin{minipage}[t]{0.3\linewidth}
  \centering
   \includegraphics[width=5.5cm]{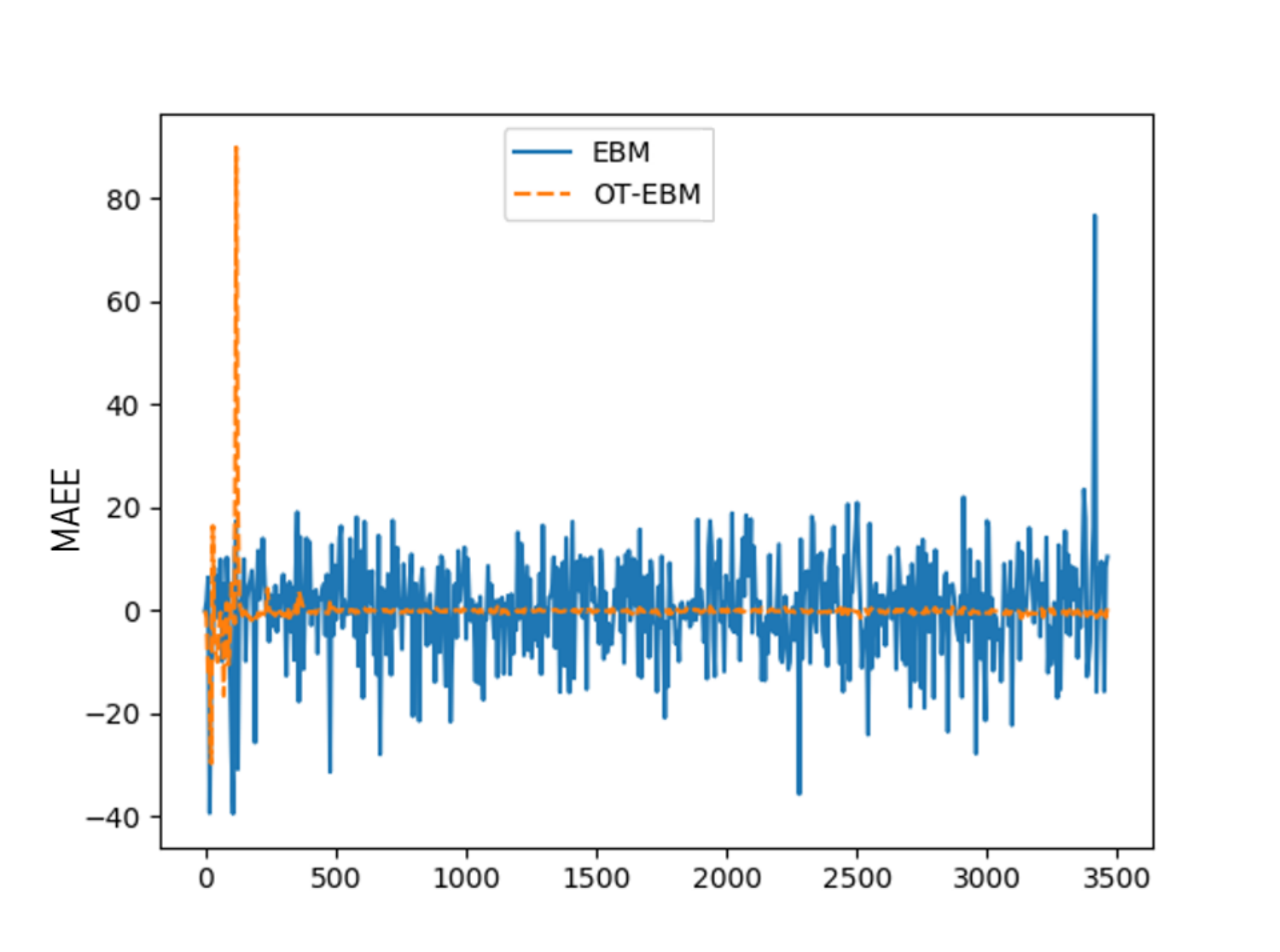}
  \end{minipage}
  }
  \caption{CelebA Human face generation results from EBM and OT-EBM. The EBM results in (b) shows the best generation before its model collapse, and (a) shows its examples of collapse results. (c) shows results of OT-EBM. (d) shows MAEE curves of EBM and OT-EBM on CelebA human face generation.\label{fig:celeba}}
\end{figure}
\subsubsection{Single Class Generation}
In this section we evaluate the near-optimal method by an implementation on large scale datasets that contain mere one class. The data are in a highly identical visual structure, for example, human faces.\\

\noindent\textbf{Human Face Modeling}\\
We show results on the public human face benchmark dataset, CelebA~\citep{liu2015deep}, which is cropped to face-centered.  The neural energy $\Phi_{\theta}$ in \eqref{eq:neuralphi} is implemented with a simple bottom-up CNN with 5 convolution layers and the non-linear activation function after each convolution layer is Leaky ReLU. The learning rates of $\theta$ and $\mathbf{x}$ are respectively $1\times 10^{-4}$ and $1$; $\tau=120$, $L=28$.

EBM and OT-EBM share the same settings and are learned without any other tricks. The dynamical samples of EBM in the early stage will transform reasonably toward the hidden human face distribution, but will always unpredictably collapse to meaningless noise during the intermediate stage. This is observed visually through Figure~\ref{fig:celeba}(a) and statistically in Figure~\ref{fig:celeba}(d). Faces generated by OT-EBM are distinctly closer to real human faces than the best results of EBM. The MAEE in Eq. \eqref{eq:maee_distance} recorded during the learning process indicates that OT-EBM is stable during a long period of training and converges in advance than EBM. These reported results can also be regarded as the evidence that OT-EBM provides a better gradient flow for learning neural parameters and persistent bounded dynamics for samples. We provide FID scores of EBMs in different structures applied with and without OT methods, listed in Table~\ref{tab:fid_celebA}.\\
\begin{table}[ht!]
 
  \centering
     \centering
  \setlength\tabcolsep{8pt}
  
  \resizebox{\columnwidth}{!}{
    \begin{tabular}{ccccccc} 
      
      \toprule
      \multirow{2}{*}{\textbf{Models}}&\multicolumn{2}{c}{\textbf{CelebA}}&\multicolumn{2}{c}{\textbf{LSUN-Bedroom}}&\multicolumn{2}{c}{\textbf{LSUN-Church}}\\
      \cmidrule(lr){2-3}\cmidrule(lr){4-5}\cmidrule(lr){6-7}
        & \textbf{w/o OT}& \textbf{w OT} & \textbf{w/o OT}& \textbf{w OT} & \textbf{w/o OT}& \textbf{w OT} \\
      \midrule
        Deep FRAME~\citep{cai2019frame}& 73.43 & \cellcolor{Gray}67.21& 100.95 & \cellcolor{Gray}70.34 &85.12&\cellcolor{Gray}72.12\\
        EBM~\citep{cai2019frame} &64.76 & \cellcolor{Gray}52.11 &65.21 & \cellcolor{Gray}56.76 &79.23& \cellcolor{Gray}61.23\\
        Stacked EBM~\citep{gao2018learning} & 53.06 & \cellcolor{Gray}44.21 &  58.23 & \cellcolor{Gray}46.13 &43.99&\cellcolor{Gray}39.2\\
        CoopNet~\citep{xie2018cooperative}& 35.22 & \cellcolor{Gray}28.17 & 34.21 & \cellcolor{Gray}30.45 &37.84& \cellcolor{Gray}30.41\\
        NCSN v2~\citep{song2020improved} &25.12&\cellcolor{Gray}22.34&23.32& \cellcolor{Gray}16.44 &17.31&\cellcolor{Gray}15.23\\
      \bottomrule
    \end{tabular}
  }
  \caption{The FID comparison among EBMs with (\textbf{w}) and without (\textbf{w/o}) applying OT method on one-class dataset benchmarks.
  \label{tab:fid_celebA}}
\end{table}

\noindent\textbf{Scene Image Generation}\\
We also provide an evaluation on a scene dataset benchmark LSUN~\citep{yu15lsun}, where the experiments are conducted on two categories `bedroom' and `church' (for indoor and outdoor scenes). The output image size is 128$\times$128 for `bedroom', and 64$\times$64 for the `church'. The backbone learning is CoopNet (ResNet100-based), and the Lipschitz constrain $L=256$. The learning rate of $\theta$ is $2\times10^{-4}$, and the learning rate is $0.15$. 

\begin{figure}[h]
  \begin{center}
    \includegraphics[width=15.25cm]{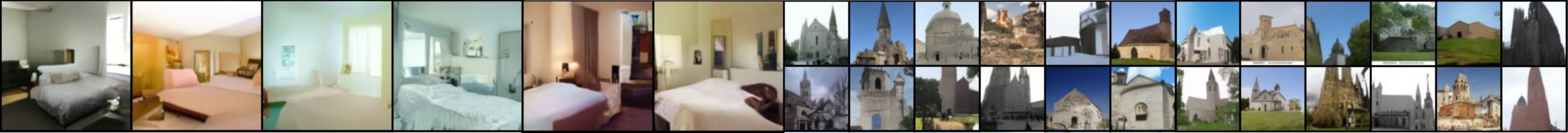}
    \end{center}
    \vspace{-5pt}
    \caption{Visualizations of LSUN-Bedroom and LSUN-Church.}
    \label{fig:buildings}
\end{figure}
We take FID as a quantity measure for on single class image generation task. Larger FID means less structural similarity beween the estimated distribution and the real distribution.  The scores are measured by 10,000 generated samples, and the results are presented in Table~\ref{tab:fid_celebA} and Figure~\ref{fig:buildings}. We observe obvious improvements of OT-EBM over EBM.

\subsubsection{Multi-category Image Modeling}
Herein we provide a capability estimation of OT on multi-class large scale datasets.\\

\begin{figure}[t]
  \centering
  \subfigure[]{
    \begin{minipage}[t]{0.26\linewidth}
    \centering
    \includegraphics[width=4.cm]{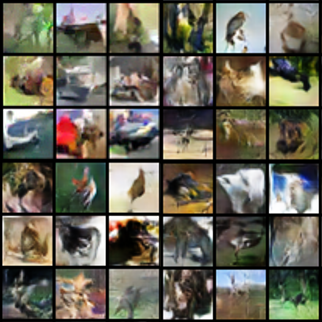}
  \end{minipage}
  }
  \subfigure[]{
  \begin{minipage}[t]{0.26\linewidth}
  \centering
  \includegraphics[width=4.cm]{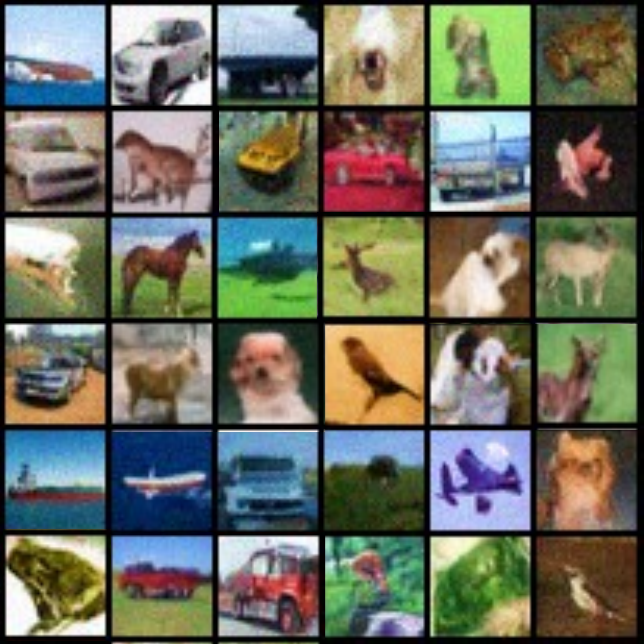}
  \end{minipage}
  }
  \subfigure[Learning curves]{
  \begin{minipage}[t]{0.4\linewidth}
  \centering
  \includegraphics[width=6cm]{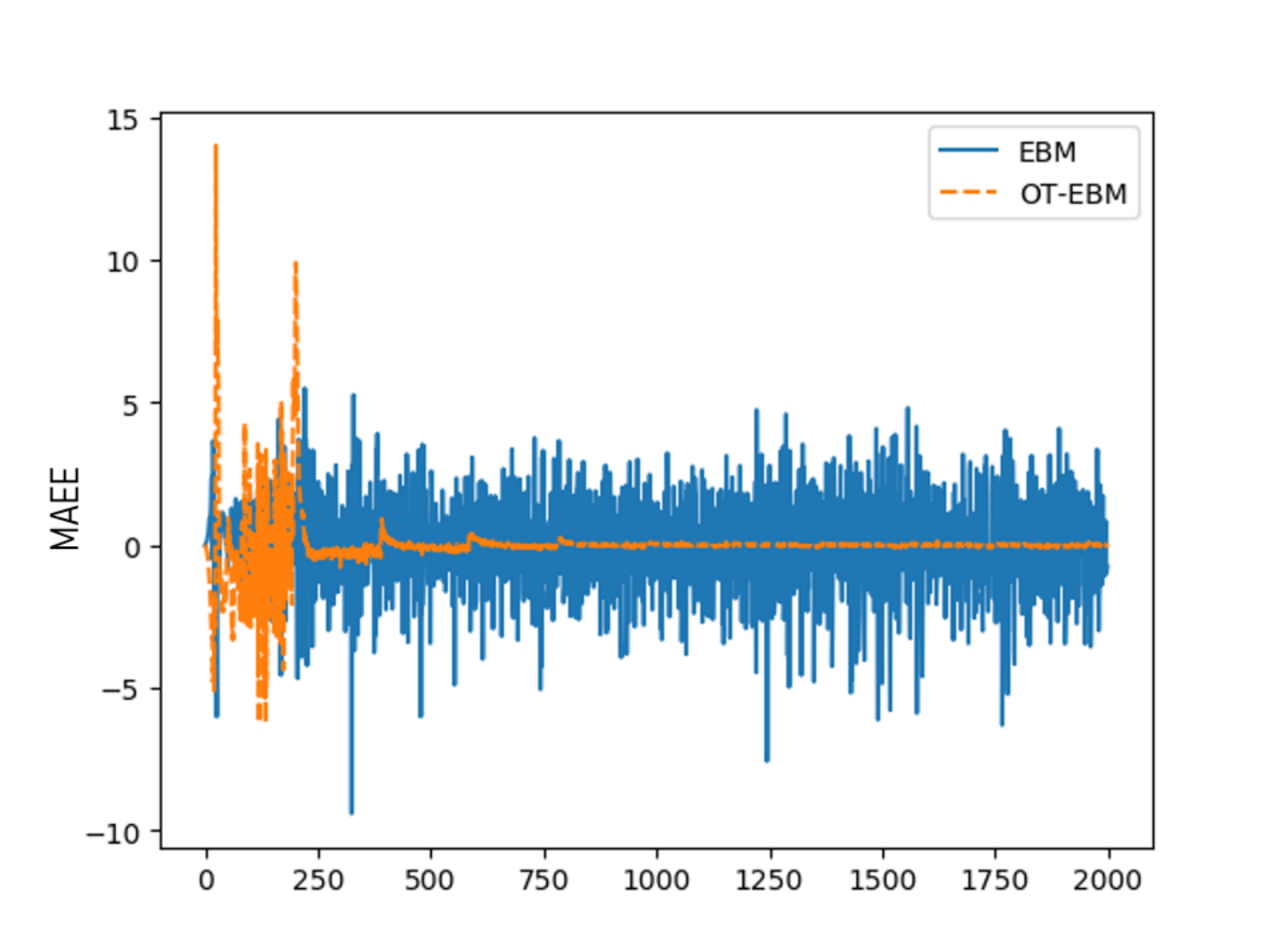}
  \end{minipage}
  }
  \caption{We present some representative generated results of EBM (a) and OT-EBM~(b) on CIFAR-10 dataset. (c) shows the MAEE curves of the mentioned two models on CIFAR-10.\label{fig:converge_speed}}
\end{figure}

\noindent\textbf{CIFAR-10 Benchmark}\\
Unlike CelebA that only contains human faces, CIFAR-10 is a well-studied dataset that consists of 10 categories of $32\times 32$ pixels natural images with a high diversity. We present the image generation results of EBM and OT-EBM in Figure~\ref{fig:converge_speed}(a)(b), respectively. Apparently, our method can greatly improve the visual sample quality as well as the convergence speed, according to the values of MAEE shown in Figure~\ref{fig:converge_speed}(c).

In this section, we evaluate the OT method with recent state-of-the-art methods on CIFAR-10, particularly on energy-based models. Apparently, when FID and IS get better, the effect of OT learning method gets gradually weaker, as shown in Table~\ref{tab:inception_score}. Considering the learned manifold or distribution is pretty close to the real one, the improving space for OT method becomes narrow. However, the synthesis quality can also be improved through OT, according to the FID records in Table~\ref{tab:inception_score}. For simplicity, we will bypass the application details of typical EBMs such as FRAME and IGEBM, and will discuss the special EBM later, such as CoopNet. Next we will particularly discuss the recent state-of-the-art (SOTA) EBM.  
\begin{figure}[ht!]
  \centering
    \subfigure[EBM]{
  \begin{minipage}[ht]{0.35\linewidth}
  \centering
  \includegraphics[width=5.5cm]{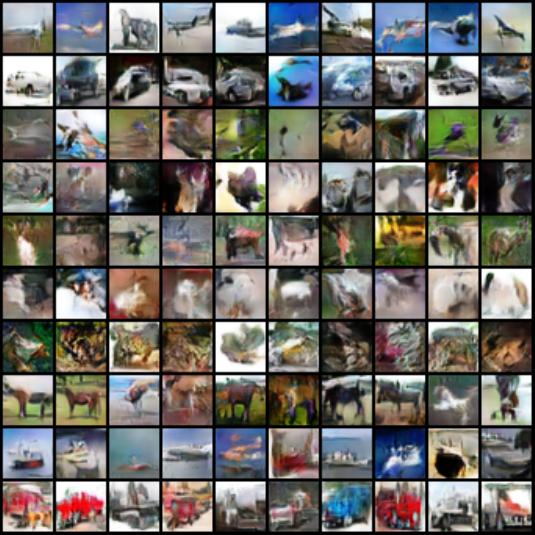}
  \end{minipage}
  }
  \subfigure[OT-EBM]{
  \begin{minipage}[ht]{0.56\linewidth}
  \centering
  \includegraphics[width=7.5cm]{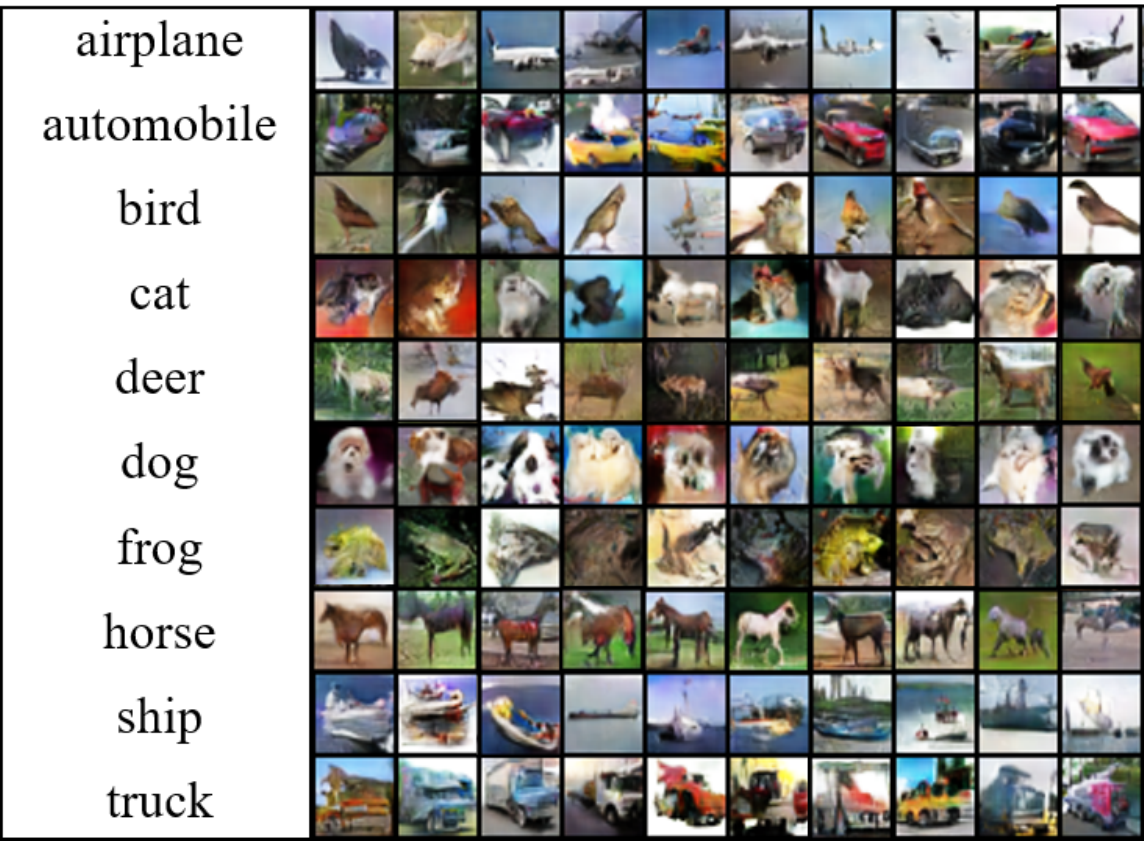}
  \end{minipage}
  }
  \caption{The generated results of conditional learning from OT-EBM and EBM, both of which are constructed with ResNet32 identically.\label{fig:cifar10-cond}}
   \vspace{-10pt}
\end{figure}
\begin{table}[t]
 \setlength\tabcolsep{6pt}
\resizebox{\columnwidth}{!}{
\begin{tabular}{@{}lllll@{}}

\toprule
 \multicolumn{1}{c}{}& \multicolumn{1}{c}{IS$\uparrow$} & \multicolumn{1}{c}{FID$\downarrow$} &\multicolumn{1}{c}{IS} &\multicolumn{1}{c}{FID} 
 \\ \cmidrule(l){2-5} 
 \multicolumn{1}{l}{\multirow{1}{*}{\textbf{Energy-based Models}}}                                    & \multicolumn{2}{l}{\textbf{Unconditional Generation}}     & \multicolumn{2}{l}{\textbf{Conditional Generation}} \\ 
\midrule
Deep FRAME~\citep{cai2019frame}                                            & 4.28$\pm$0.05                           & 76.34                   & 4.95$\pm$0.05                      & 67.20                \\
w-Deep FRAME~\citep{cai2019frame}                                          & 5.58$\pm$0.13                           & 49.94                   & 6.05$\pm$0.13                      & 42.09                \\
\rowcolor{Gray}
Deep FRAME\textbf{+OT}                                          & 5.62$\pm$0.11                           & 50.41                   & 6.12$\pm$0.11                      & 40.19 \\
\cmidrule(r){1-1}
EBM(or GCN)~\citep{cai2019frame}                                             & 4.74$\pm$0.07                           & 50.26                   & 5.05$\pm$0.07                      & 68.37                \\
w-EBM                                              & 5.37$\pm$0.05                           & 55.12                   & 6.23$\pm$0.08                      & 50.78                \\
\rowcolor{Gray}
EBM\textbf{+OT}                                              & 5.78$\pm$0.09                           & 51.84                   & 6.76$\pm$0.09                      & 43.52                \\
\cmidrule(r){1-1}
stacked EBM~\citep{gao2018learning}                                              & 5.02$\pm$0.05                           & 45.60                   & 6.57$\pm$0.05                      & 57.56                \\
\rowcolor{Gray}
stacked EBM\textbf{+OT}                                        & 5.91$\pm$0.11                           & 49.93                   & 6.89$\pm$0.05                      & 41.23                \\
\cmidrule(r){1-1}
IGEBM~\citep{du2019implicit}                                                    & 6.02                               & 40.58                   & 6.78                           & 38.2                 \\
\rowcolor{Gray}
IGEBM\textbf{+OT}                                               & 6.37$\pm$0.06                           & 36.27                   & 7.02$\pm$0.07                      & 35.98                \\

\cmidrule(r){1-1}
CoopNet~\citep{xie2018cooperative}                                                  & 6.55$\pm$0.08                           & 26.22                   & 7.10$\pm$0.05                      & 34.49                \\
CoopNet\textbf{\textbf{+OT}}                                           & 6.69$\pm$0.07                           & 28.98                   & 7.22$\pm$0.11                      & 27.42                 \\
\rowcolor{Gray}
CoopNet\textbf{\textbf{+OT}} + Atten                                           & 7.01$\pm$0.04                           & 20.23                   & 7.92$\pm$0.13                      & 23.98                 \\
\cmidrule(r){1-1}
NSCN~\citep{song2019generative}                                                     & 8.87$\pm$0.12                  & 25.32                   & -                              & -                    \\
\rowcolor{Gray}
NSCN\textbf{+OT}                                                & 8.98$\pm$0.15                           & 19.78                   & -                              & -                    \\
NSCN v2~\citep{song2020improved}                                                  & 8.40$\pm$0.07                           & 10.97          & -                              & -                    \\
\rowcolor{Gray}
NSCN v2\textbf{+OT}                                            & 8.68$\pm$0.10                           & 9.01                   & -                              & -                    \\ 

\midrule
\multicolumn{5}{l}{\textbf{GAN-based Models}}                                                                                                                                      \\
\midrule
SN-GANs~\citep{miyato2018spectral} &-&-&8.22$\pm$0.05&21.7\\
BiGAN~\citep{brock2018large}                                                    & -                                   & -                       & 9.22                           & 14.3                 \\
StyleGAN v2 ADA~\citep{karras2020training}                                               & \textbf{9.83}                                  & \textbf{2.92}                       &\textbf{10.14}                          & \textbf{2.42} \\
\bottomrule
\end{tabular}
}
 \caption{The IS comparison on CIFAR-10 benchmark, especially among EBMs with (\textbf{+OT} or OT-EBM) and without applying OT (dataset CIFAR-10). `w-' stands for learning with the JKO discrete flow $\mathcal{F}_{w^2}$. `+Atten' means adding up self-attention mechanism to the neural EBM.
  \label{tab:inception_score}}
\end{table}

As for the SOTA energy-based generative model NCSN~\citep{song2019generative,song2020improved,song2021scorebased}, the claimed performance on CIFAR-10 is difficult to reproduce, especially for IS. To achieve the baseline performance, we apply a sampling-by-learning strategy, namely using generated samples from several well-trained (after 10,000 iterations learning and 1,000 iterations of Langevin dynamics each batch) NCSN models for the evaluation of IS. With this strategy, the evaluation process will become much more time-efficient and the obtained IS results are more reasonable. Despite of this, the best IS of NSCN we obtain can only reach 8.72, and after applying OT method, we achieve 0.3 improvement over our baseline. And thus for FID, the baseline is 10.97 and after applying OT we achieve FID 9.01. Better IS and FID comparing with the record in public  result from estimated gradients, rather than accurate gradients computed by normal EBM.

The application of OT on NCSN is quite simple. We maintain the anneal langevin dynamic sampling strategy; however, we also reserve the gaussian noise item in the original SGLD (Eq.~\eqref{eq:sgld}), which has been abandoned in OT, since the removal of this item will cause bias accumulation, according to the sampling mechanism of annealing. And the main modification happens on the parameter estimation \eqref{eq:new_learning}b. The settings stays the same with the NCSN default settings. We also notice OT only achieves unstable and limited improvements on NCSN models, due to the approximated gradients NCSN computes. 
\\

\begin{figure}[h]
  \begin{center}
    \includegraphics[width=15.25cm]{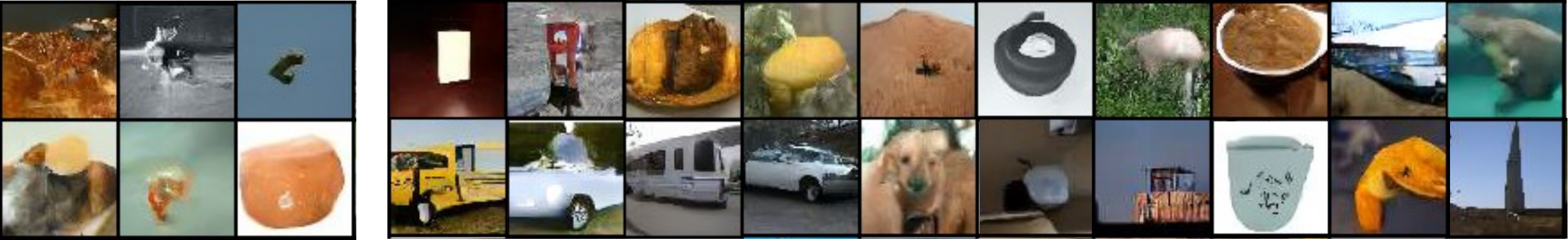}
    \end{center}
    \vspace{-5pt}
    \caption{Visualization of unconditional ImageNet generation. The left block is from original EBM, the right is from model learned with OT method. }
    \label{fig:imagenet}
    \vspace{-5pt}
\end{figure}

\noindent\textbf{ImageNet Benchmark}\\
To explore the capability of OT method on EBM for large scaling image modeling, we also conduct generation experiments on ImageNet dataset. The synthesized image size is 64$\times$64. We visualize the testing results in Figure~\ref{fig:imagenet}, presenting the whole sampling process from noises to identifiable images and it turns out the final generation quality stable. The base energy-based model is NCSNv2. Considering the image and object class amount of ImageNet is much larger than the other datasets, the hyper-parameter settings are set highly different from the default ones. As a results, $L=$128 and we have anneal sampling step learning rate $5.12\times10^{-6}$. It can be observed that synthesized objects from EBM with OT are more identifiable with clear edges than those from the original EBM. Note that this generation is unconditional.
We achieve IS 45.4 and FID 52.3 on this benchmark dataset. 


\subsection{Extended Experiments}
\label{sec:capacity}
In this section, we will show that our near-optimal learning method maintains a superior performance over EBM on the extended tasks, including the image inpainting, cooperative learning, high resolution generation and graph inference, to
 demonstrate the capacity of addressing image reconstruction with missing information,
application on cooperative training, high-dimensional samples and even structural data. \\

\begin{figure}[ht]
  
  \centering
  \subfigure[Square Masked]{
  \begin{minipage}[t]{\linewidth}
  \centering
  \includegraphics[width=15cm]{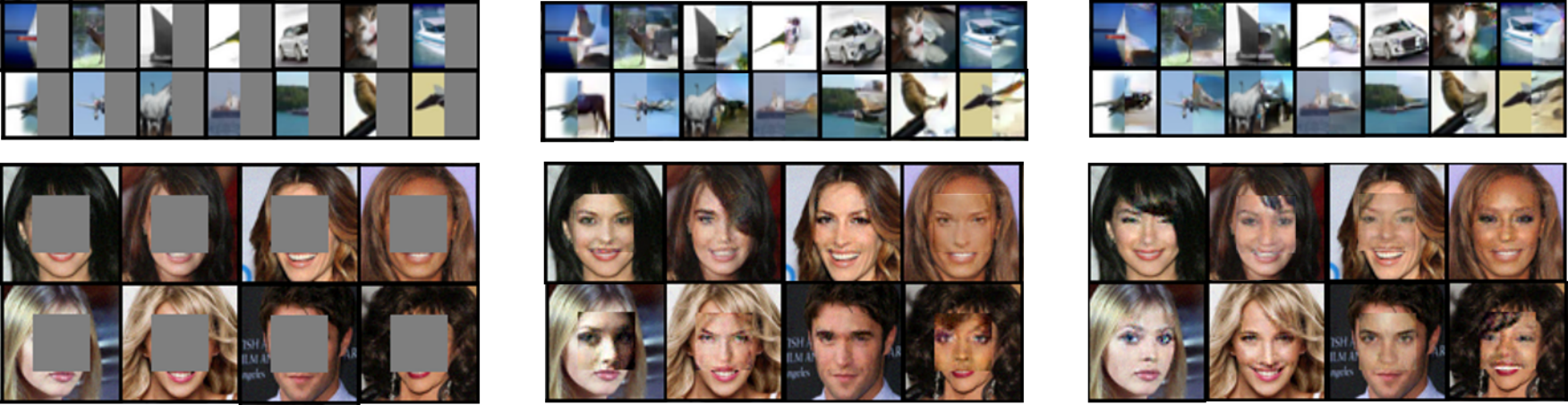}
  \end{minipage}
  }
  \subfigure[Spot Masked]{
  \begin{minipage}[t]{\linewidth}
  \centering
  \includegraphics[width=15cm]{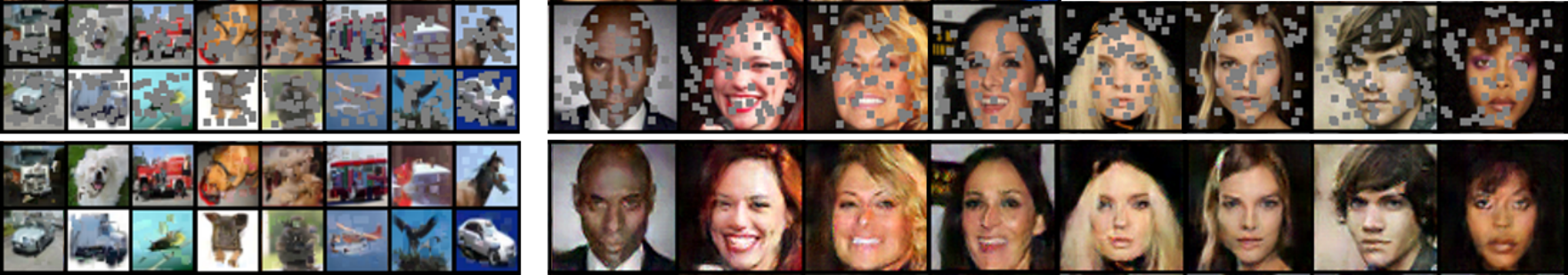}
  \end{minipage}
  }
  \caption{Image inpainting examples of OT-EBM for learning images from masked CIFAR-10 and CelebA. (a) are square mask results for an evaluation on mode coverage, where the middle and the last column respectively represent the outcomes of OT-EBM on different stages, and the left column presents the input. (b) are the spot masks for examining image recovery ability, where the first row is the input masked images, and the second row shows the restored results.\label{fig:inpainting}}
\end{figure}

\noindent\textbf{Image Inpainting}\\
Generative models are expected to approximate the data distribution moderately well under a global statistical distance measure, and to find a good data manifold that covers nearly all the data points and every mode. A common test for the mode coverage and the generalization is image inpainting~\citep{bertalmio2000image}. In Figure~\ref{fig:inpainting}, we apply two types of masks to test the ability to sample the masked pixels, and run two fully trained OT-EBMs on these images. To test the mode coverage and the overfitting, we use the large square masks to cover the half of the pixels; to test the model restoration ability, we use random small spots as masks. Results show that OT-EBM possesses the ability of diverse completions on human faces and natural images, and infers reasonable contents from the blank. This indicates that OT-EBM has characterized diverse modes of synthesis.\\

\noindent\textbf{Cooperatively Learning with a Generator}\\
 The generator network is usually based on the latent variable model in machine learning, and seeks to establish a mapping from low-dimensional latent space to high-dimensional data space. Learning this mapping directly is painful since one has to deal with many intractable probabilistic computations that might arise in maximum likelihood estimation and related strategies. GAN is proposed to learn the generator adversarially and implicitly with a discriminator. Noting that EBM can either be a generator or a discriminator, ~\citep{xie2018cooperative} proposes CoopNet where they treat EBM as a discriminator (or a synthesizor as they name it) and train the networks cooperatively; namely, the latent variable model (the generator) can speed up the time-consuming sampling in EBM, while EBM distills its knowledge into the generator. To boost the performance of cooperative learning models, we attempt to manipulate the network structure to see the improving potential. And it proves that not only deeper CNN with more general learned representations will benefit the generation process, but the self-attention mechanism~\citep{vaswani2017attention} also helps the final quality of the sampling, as shown in Table~\ref{tab:inception_score}.

To examine whether or not learning EBM in CoopNet with our near-optimal OT approach can also maintain the positive effectiveness on the latent variable generator, we evaluate the CoopNet model learning with our near-optimal OT scheme visually in Figure~\ref{fig:cifar10-cond}, and present comparisons on IS in Table~\ref{tab:inception_score}. Details of this learning can be found in Appendix B. In conclusion, the results show that our approach also works on CoopNet.\\

\begin{figure}[t]
  \centering
  
  \subfigure[EBM]{
  \begin{minipage}[ht]{\linewidth}
  \centering
  \includegraphics[width=14cm]{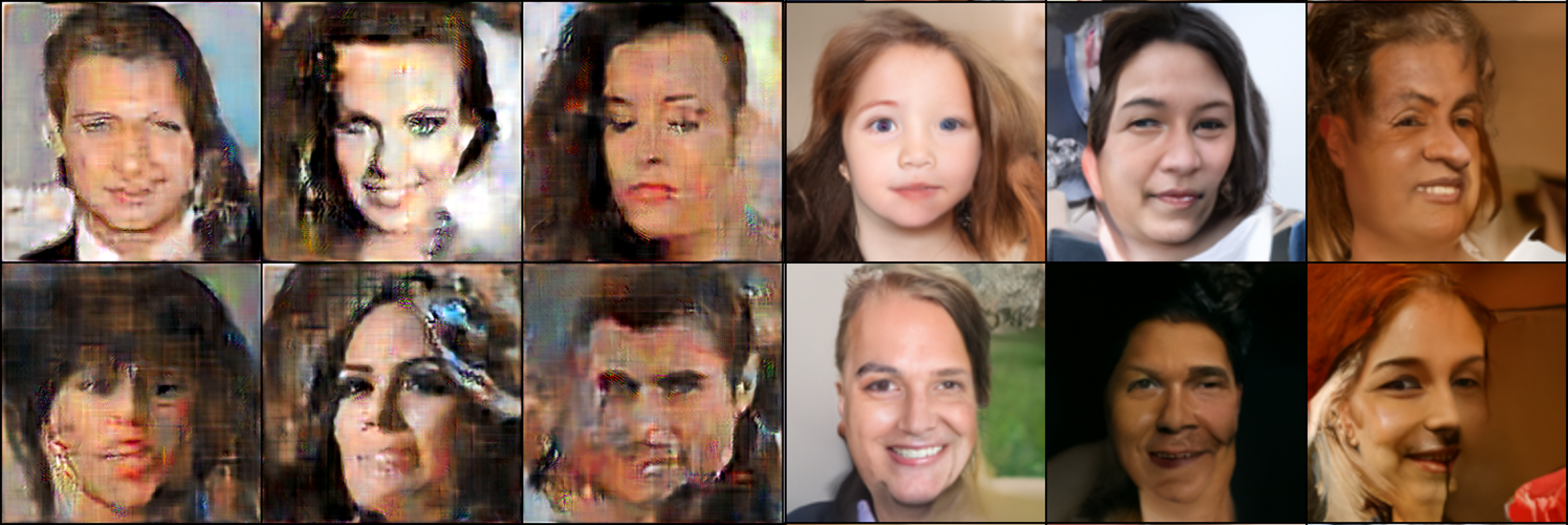}
  \end{minipage}
  }
  \subfigure[OT-EBM]{
  \begin{minipage}[ht]{\linewidth}
  \centering
  \includegraphics[width=14cm]{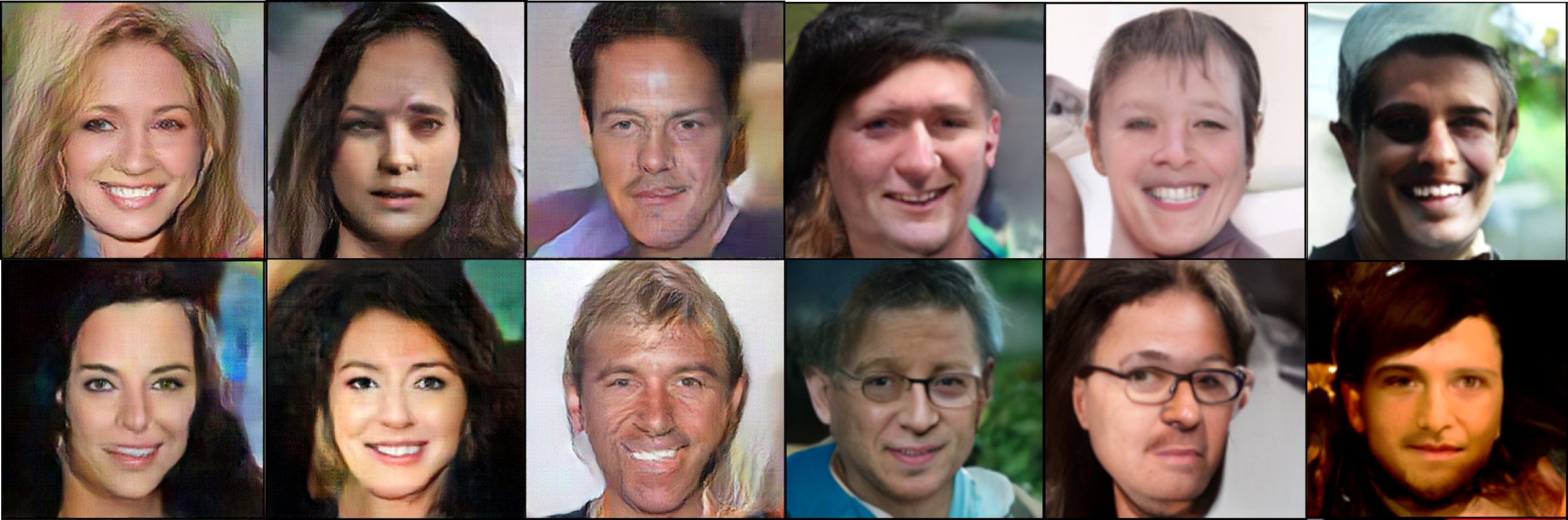}
  \end{minipage}
  }
  
  \caption{High resolution human face generation results comparison. The base networks are both Residual CNN. Left half pictures are from CelebA, the other half are from FFHQ.\label{fig:celebabig}}
\end{figure} 
\noindent\textbf{High Resolution Face Generation}\\
In this section, we further implement OT-EBM on high-resolution generation task.
Implicit generative models show an extraordinary ability on high-resolution image generation~\citep{brock2018large}. However, the dimensionality is a curse in sampling-based models including EBM, because the sampling chain will grow exponentially with the dimensionality. Unsurprisingly, we will face the dilemma of when to early stop the sampling algorithm. If the stopping time is too early, the sampler can barely sample anything or will be under-fitted; if the stopping time is too late, the algorithm can easily go wrong due to overfitting or collapsing. Our OT-based EBM can solve this problem to a certain extent, crediting to the bounded gradient flow and deterministic dynamics. In practice, we can simply set the sampling time long enough, and the model can avoid under-fitting and will remain stable in terms of learning parameters and generating samples.

To test the EBM generation on high dimensional data, we use a high-resolution version of CelebA dataset, CelebA-HQ~\citep{Karras2018} and Flickr-Faces-HQ Dataset(FFHQ)~\citep{karras2019style}, where each face image has $1024\times 1024$ pixels. We use a deeper residual CNN and resize the input to $256\times 256$, and the results are provided in Figure~\ref{fig:celebabig}. On the CelebA benchmark, where conventional EBM performs poorly, the OT method can effectively improve the generation quality. In this generation situation, EBM has already achieved relatively high-quality generated results, while OT promotes the generation with more smooth edges and reasonable shapes. We provide more high resolution generation in the Appendix~\ref{fig:high_resolution}.
\\

\begin{figure}[t]
  \centering
  
  \subfigure[EBM]{
  \begin{minipage}[t]{0.31\linewidth}
  \centering
  \includegraphics[width=5cm]{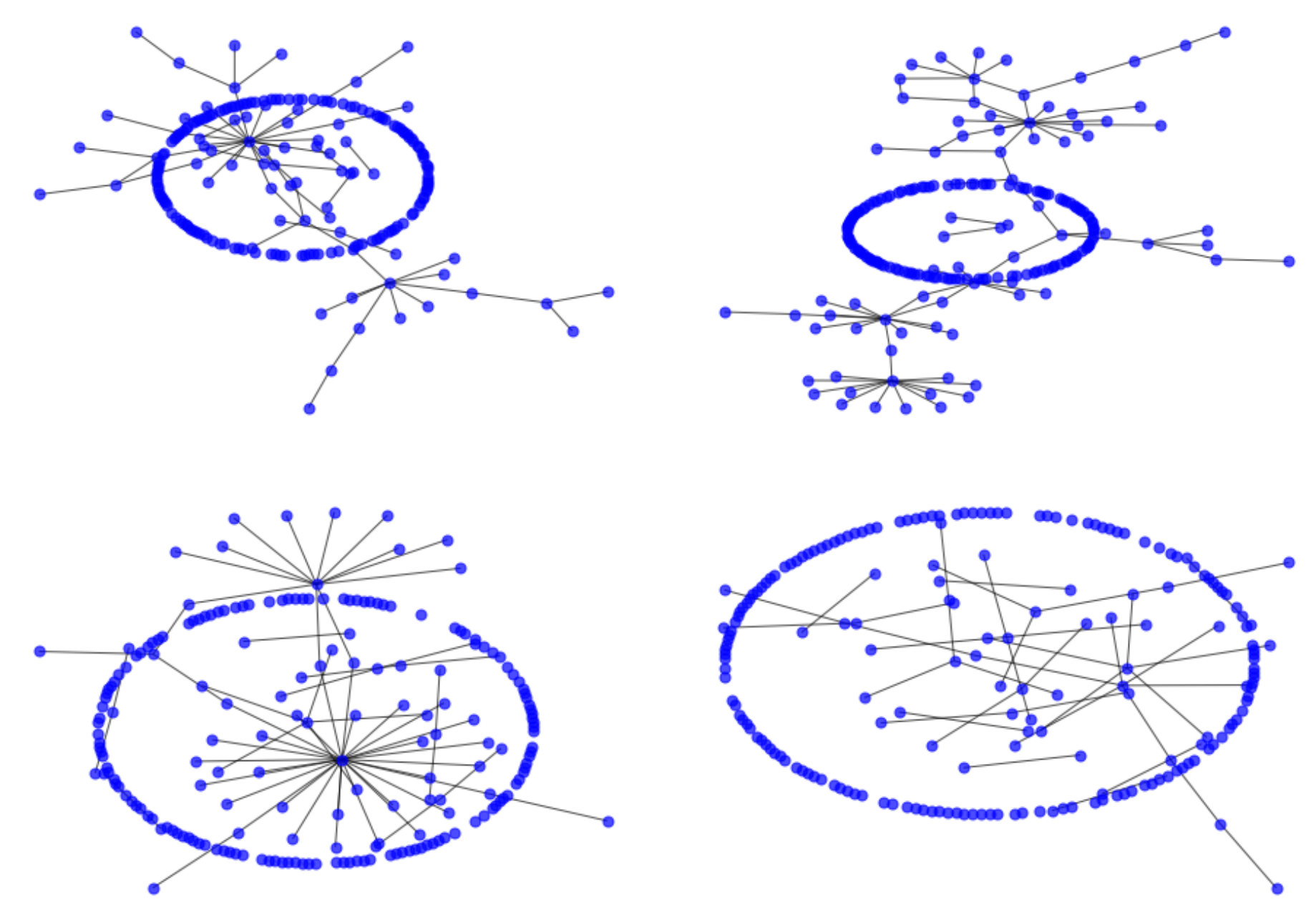}
  \end{minipage}
  }
  \subfigure[OT-EBM]{
  \begin{minipage}[t]{0.31\linewidth}
  \centering
  \includegraphics[width=5cm]{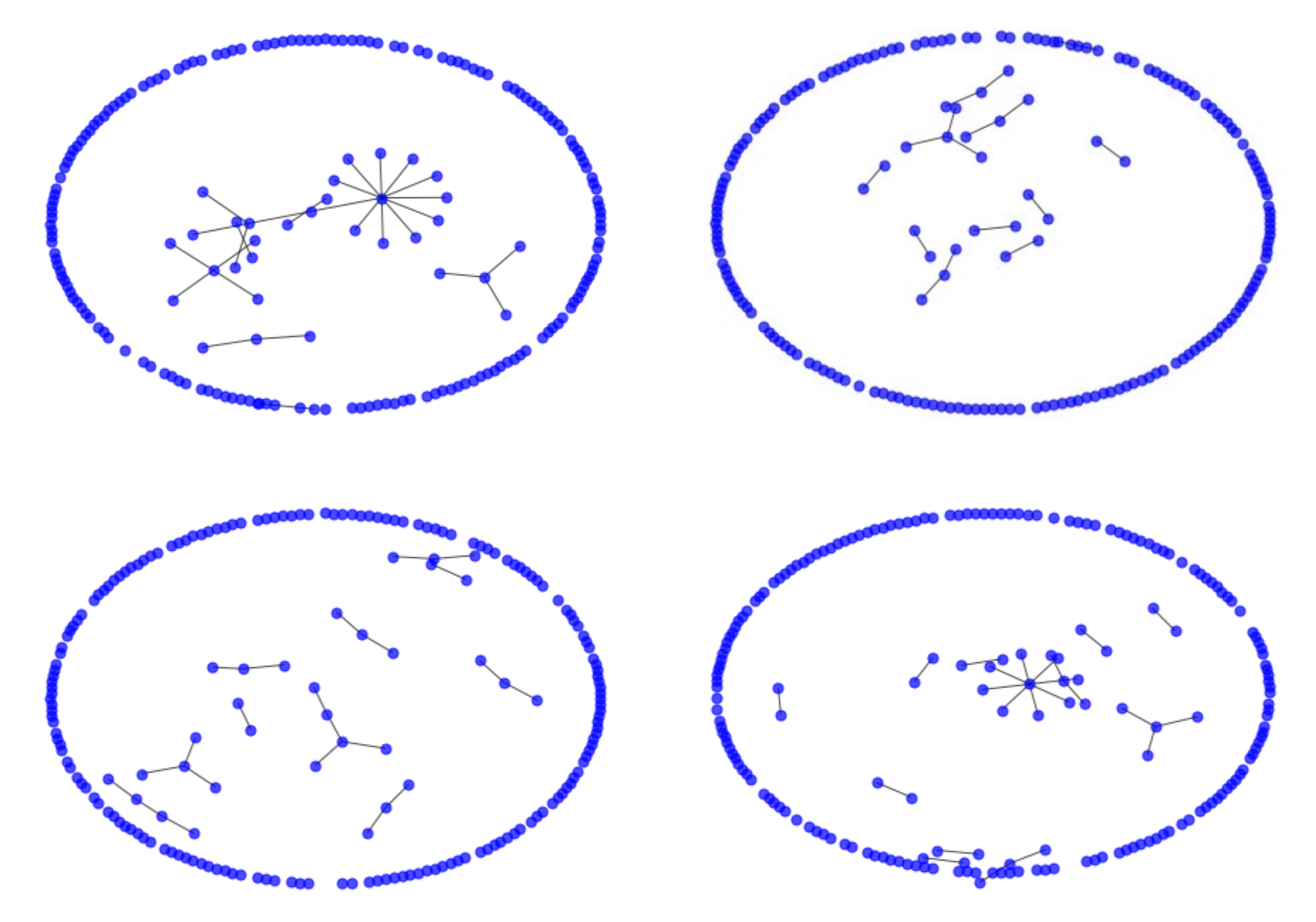}
  \end{minipage}
  }
  \subfigure[Ground-Truth]{
    \begin{minipage}[t]{0.31\linewidth}
    \centering
    \includegraphics[width=5cm]{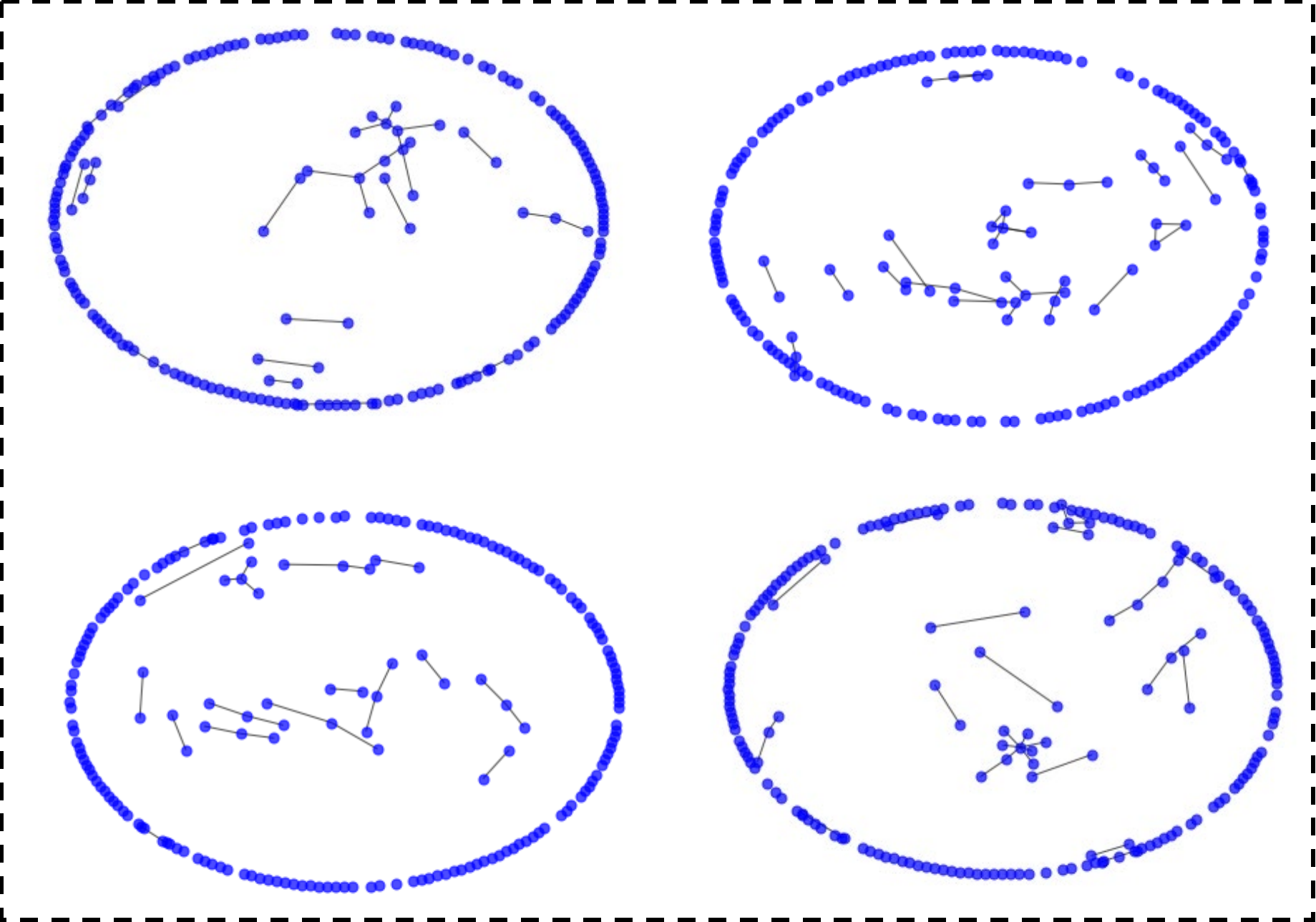}
    \end{minipage}
    }
  \caption{Visualization of generated graph relations. The first column depicts the inferenced graph from EBM (a), the middle column graphs are from OT-EBM (b), and (c) is the ground truth. \label{fig:graph}}
\end{figure}
\noindent\textbf{Discrete Graph Inference}\\
To prove that our method can also improve the modeling ability of degenerate distributions, we consider the problem of rebuilding graphical relationships~(for example the social networks). This is done by implementing a classification graph convolution network~(GCN)~\citep{kipf2016semi} with a multi-layer perceptron~(MLP) that operates directly on the graphic related data. We can then treat it as an EBM and generate embedding vectors of nodes based on properties of the neighborhoods.

As an instance of this problem, we use the Cora dataset~\citep{Sen2008}, which is a text dataset including 2708 scientific publications classified into seven classes. Its citation network has a total of 5000+ links and each word dictionary contains 1433 unique words. We pick several groups of subgraphs, each of which contains 200 randomly selected nodes to measure the internal connection instances in the generated graph. We visualize the generated graph results in Figure~\ref{fig:graph}, where (a)~(b)~(c) are graphs respectively from the EBM, OT-EBM, and the training data (the ground-truth).

We find that the estimated connection of OT-EBM is highly similar to the ground-truth in the sense of graphical structure, while EBM produces too dissimilar structural and excessively heavy-weighted sub-graphs. 
Through the visualization, we can observe that the graphs synthesized from EBM have nearly no small subgraphs, which are rare in the ground-truth graphs. On the contrary, our OT method shows the advantage on more delicate graph synthesis, which explains why OT-EBM achieves higher classification performance in Table~\ref{tab:graph}, where for quantitative evaluation, we measure three types of the accuracy of generated graphs:
\begin{itemize}
  \item \textbf{Training Accuracy.} This metric is the highest reaching accuracy during model training or learning.
  \item \textbf{Synthesis Accuracy.} 
  This is used to measure the quality of synthesized graphs by a pre-trained GCN with graph training set. Synthesized graphs would have a similar structure with the training set; namely, better synthesis will obtain higher accuracy on the normally trained GCN, since a well pre-trained GCN can maturely capture the structural features. 
  \item \textbf{Testing Accuracy.} It is the highest testing accuracy on different learning modes. 
\end{itemize} 

\begin{table}[ht!]
 
  \centering
  \resizebox{0.65\columnwidth}{!}{
  \begin{tabular}{cccc}
  \toprule
       \textbf{Method}& \textbf{Training Acc.} &\textbf{Synthesis Acc.} &\textbf{ Testing Acc.}  \\
       \midrule

        GCN &1.0& - & 0.81\\
        \midrule
        EBM&0.34 &0.16 & 0.15\\
        OT-EBM&\textbf{0.64} &\textbf{0.45} & \textbf{0.31}  \\
       
  \bottomrule
  \end{tabular}
  }
  \caption{The accuracy of the generated graphs in the graph relation inference scenario. \label{tab:graph}}
\end{table}
As shown in Table~\ref{tab:graph}, all the evaluation results show that our OT strategy trained GCN achieves higher accuracies. On the one hand, our OT method improves the learning GCN progress such that the learned GCN obtains a better representative capacity; on the other hand, OT benefits the sampling process that the synthesized relations form a more similar graph structure than those from EBM. Namely, the proposed OT learning scheme presents an appealing potential to generate visual samples that simultaneously match both the real data distribution and the global graphic relation distribution.

\section{Conclusion}
In this paper, we revisit EBMs from the perspective of optimal transport and discover that the unstable gradient flow may lead to sub-optimal parameter learning and data sampling. Based on this discovery, we propose OT-EBM by reformulating the consecutive time KL discrete flow into the Benamou-Brenier JKO discrete flow, which considers the neural energy to be Lipschitz continuous. OT-EBM is proved to be a near-optimal discrete time scheme in transporting particles along the optimal transmission path with the second-order approximation. Extensive experimental results demonstrate that our proposed method has a superior modeling performance and stability across a variety of datasets.
Our investigation on the generalization with EBMs also provides an impetus for future researches on EBMs, which shows that 
OT-EBM is more stable and can benefit almost all existing EBMs and their applications.

\newpage

\appendix
\section*{Appendix A. Theory Proof}

\subsection*{Proof of Theorem~\ref{thm:1}.}
\label{append:proof}
\begin{proof}
  Under the Gaussian reference measure, to simplify the proof, we study the neural energy $\Phi_{\theta}$ in \eqref{eq:neuralphi} with only one linear operation (convolution), and presume the nonlinear activation function $h(\cdot)$ to be ReLU functio,
  \begin{equation}
    \label{eq:relu}
    h(\cdot)=\max(\cdot,\ 0).
  \end{equation}
  Namely, $\Phi_{\theta}$ is reduced to $\Phi_{\theta} = h(\omega\mathbf{x} + b)$. Our results can be easily extended to multilayer neural energy models. \citep{xie2016theory} shows that the corresponding EBM has the piecewise Gaussian property, as summarized in the following Theorem~\ref{thm:piecegauss}.
  \begin{theorem}
    \label{thm:piecegauss}
    \textnormal{\citep[Theorem 1]{xie2016theory}}
    If the nonlinear activation in \eqref{eq:neuralphi} is ReLU, then the distribution of EBM in \eqref{eq:ebm} is piecewise Gaussian, and on each piece the probability density can be written as
    \begin{equation}
      \label{eq:piecegauss}
      p_{\theta}(\mathbf{x}) \propto \exp\left[-\frac{\left\Vert \mathbf{x}-\mathbf{y} \right\Vert^{2}}{2} \right],
    \end{equation}
    where $\mathbf{y}=\sum_{k=1}^{K}\mathcal{I}_k\theta_{k}$ is an approximate reconstruction of $\mathbf{x}$ in one part of the data space by a linear transformation. Here $\theta=(\omega,b)$ is $K$-dimensional and $\mathcal{I}_k = \mathcal{I}_k(\mathbf{x};\theta) = \mathds{1}(\omega_k \mathbf{x}+b_k>0)$ is an indicator function.
  \end{theorem}
  The above theorem implies an implicit encoding and decoding process: the encoding process is bottom-up and infers $\mathcal{I}_k = \mathcal{I}_k(\mathbf{x};\theta) = \mathds{1}(\omega_k \mathbf{x}+b_k>0)$, where $\omega_k$ plays a role of filter; the decoding process is top-down and reconstructs $\mathbf{x}=\sum_{k=1}^K \mathcal{I}_k\theta_k$, where $\omega_k$ plays the role of basis function. Thus we can represent $\mathbf{x}$ by a $K$-dimensional activation vector $\mathcal{I}=(\mathcal{I}_1\ldots\mathcal{I}_K)$, where $\mathcal{I}_k \in\{0,1\}$, and the data space is divided into $2^{K}$ pieces.
  
  Theorem~\ref{thm:piecegauss} also implies that the aforementioned encoding process can be regarded as the evolution of Brownian particles with Gaussian transition kernels, where data samples are equivalent to Brownian particles: a Brownian particle is a particle whose position in $\mathbb{R}^d$ is given by a Wiener process, for which the probability of a particle moving from $\mathbf{x}\in\mathbb{R}^d$ to $\mathbf{y}\in\mathbb{R}^d$ is given by the probability density in Gaussian form, as in \eqref{eq:piecegauss}.
  
  Denote the data space that consists of data sharing the same activation pattern $\mathcal{I}$ as $A(\mathcal{I};\theta)=\{\mathbf{x}:\mathcal{I}(\mathbf{x};\theta)=\mathcal{I}\}$, then the probability density on this piece is $\mathcal{N}(\sum_{k=1}^K \mathcal{I}_k\theta_k, I)$. The evolution of the system is prescribed by the transition probability of particles, which is equivalent to being driven by the reconstruction error $\mathbf{x} - \sum_{k=1}^K \mathcal{I}_k\theta_k$ with additive noise. The reconstruction error is formulated as
  \begin{equation}
    \begin{split}
    \label{eq:sgld_2}
    \mathbf{x}_{t+\tau} - \mathbf{x}_{t} &= -\tau(\mathbf{x}_t - \mathbf{y}) + \mathcal{N}(0,I) \\
    &= -\tau\left(\mathbf{x}_t - \sum_{k=1}^K \mathcal{I}_k(\mathbf{y};\theta)\theta_k\right) + \mathcal{N}(0,I) \\
    &= -\tau\left(\sum_{k=1}^K\mathcal{I}_k(\mathbf{x};\theta)\theta_k -\sum_{k=1}^K\mathcal{I}_k(\mathbf{y};\theta)\theta_k\right) + \mathcal{N}(0,I) \\
    &= -\tau\sum_{k=1}^K\mathcal{I}_k^{\prime}\theta_k + \mathcal{N}(0,I).
    \end{split}
  \end{equation}
  We observe that $\nabla_{\mathbf{x}}\Phi_{\theta}=\mathcal{I}^{\prime}\theta$, which means the above decoding process \eqref{eq:sgld_2} is actually the same as sampling from $p_{\theta}$ using SGLD \eqref{eq:sgld}. Now that we have established a bridge between the EBM density and transporting particles. For the discrete diffusion equation \eqref{eq:sgld_2}, considering that each $\mathbf{x}_t$ is a random process and for a fixed time step $\tau>0$, a large deviation rate functional characterizes the behaviour (empirical measure) of the particle system at $t=\tau$ in terms of the initial distribution at $t=0$. And also for \eqref{eq:sgld_2}, a single step in the time-discretized transportation can always be characterized by the minimization of a gradient flow functional, which in our case should be the same as the LDP rate functional.
  
  Next we show that the LDP rate functional is in the exact form of KL gradient flow $\mathcal{F}_{\text{KL}}$ in \eqref{eq:kldflow}. Before we proceed the proof, we first introduce some preknowledge and notations. Let $\{\mathbf{x}^i\}_{i=1}^{n}$ be \emph{i.i.d.} random variables with finite values and with distribution $p$. We regard the $\{\mathbf{x}_t^i\}_{i=1}^{n}$ as random processes of positions of particles in the space, so that their concentration is given by their empirical measure $\rho_t=\frac{1}{n}\sum_{i=1}^{n}\delta_{\mathbf{x}_t^i}$. The key tool for LDP is Sanov's theorem, and its Boltzmann expression is as follows.
  
  \begin{theorem}
    \label{thm:sanov}
    \textnormal{(Boltzmann-Sanov)}
    Let $S$ be closed subset such that $S$ is the closure of its interior, then
    \begin{equation}
    \label{eq:sanov}
    \lim_{n\rightarrow\infty}\frac{1}{n}\log\mathbb{P}[\rho\in S]=-\min_{\mu\in S}D_{\rm{KL}}(\mu||\nu).
    \end{equation}
  \end{theorem}
  
  We consider that the behaviour of empirical measure $\rho_{\tau}$ is under the condition of a given initial distribution $\rho_0=\frac{1}{n}\sum q_{\mathbf{x}^i}$. The Sanov theorem~\ref{thm:sanov} expresses the probability of $\rho_{\tau}$ being close to $\rho$ as
  \begin{equation}
    \label{eq:our_sanov}
    \mathbb{P}(\rho_{\tau}\approx \rho | \rho_0) \approx \exp\left[-\min_{\mu\in \mathcal{T}(\rho,\rho_0)}D_{\text{KL}}(\mu||\rho_0)\right].
  \end{equation}

  Since each Brownian particle is high-dimensional and the samples we consider generally have structural information, we can treat samples as a connected graph, where each node is a subparticle. However, Sanov's theorem can only capture the global particle movements without reckoning the position of internal subparticles. To solve this, we can assign subparticles to cliques and assume that the subparticles in the same clique have similar dynamics, while different cliques are independent (due to the Markov property). Let $C$ denote the number of cliques, and denote $\mathbf{x}^{i,j}$ to be the $j$-th clique in particle $\mathbf{x}^{i}$, then $\{\mathbf{x}^{i,j}\}_{j=1}^{C}$ are i.i.d. Also note that we can treat each $\mathbf{x}^{i,j}$ as $d$-dimension if we set the value of irrelevant dimensions of $\mathbf{x}^{i}$ to $0$. Naturally, we can write
  \begin{equation}
    \label{eq:x_sum}
    \mathbf{x}^{i}=\mathbf{x}^{i,1}+\cdots+\mathbf{x}^{i,C}.
  \end{equation}
  For a $d$-dimensional vector $\theta$, it can be partitioned into $C$ blocks, where each block corresponds to a clique. We denote the partitioned $\theta$ as $\beta$, to a constant coefficient. We can calculate moment generating function $M_{\beta}=\mathbb{E}[e^{\beta\mathbf{x}}]$ for evaluation, since its logarithm $\log M_{\beta}$ happens to be a free energy function. We set 
  \begin{equation}
    \label{eq:beta}
    \beta=\frac{\theta\tau}{n}
  \end{equation}
  for derivation simplicity. Given the mean of $C$ subparticles of particle $\mathbf{x}^{i}$ as $\rho_{\tau}^{i}=\frac{1}{C}\sum_{j=1}^{C}\mathbf{x}^{i,j}$, the law of large numbers states that $\rho_{\tau}^{i}\rightarrow p$ almost surely as $n\rightarrow\infty$. The following Cram{\'e}r's theorem shows how fast this approximation is.
  
  \begin{theorem}
    \label{thm:cramer}
    \textnormal{(Cram{\'e}r)}
    Given a sequence of i.i.d. real valued random variables $\{\mathbf{x}^i\}_{i=1}^{n}$ with a common moment generating function $M_{\beta}=\mathbb{E}\left[e^{\beta\mathbf{x}^{1}}\right]<\infty$, then the following holds
    \begin{equation}
      \label{eq:cramer}
      \lim_{n\rightarrow\infty}\frac{1}{n}\log\mathbb{P}[\rho\geq a]=-I(a) \quad (a\in\mathbb{R}^d),
    \end{equation}
    where $I(a)=\max_{\beta\in\mathbb{R}^d}\left(\beta a - \log\mathbb{E}\left[e^{\beta\mathbf{x}^{1}}\right]\right)$ is the Legendre transformation of $\mathbb{E}\left[e^{\beta\mathbf{x}^{1}}\right]$.
  \end{theorem}
  \vspace{-10pt}
  Now since $\{\mathbf{x}^{i,j}\}_{j=1}^{C}$ are independent and identically distributed, with \eqref{eq:x_sum}, we have
  \begin{equation*}
    \begin{split}
    \mathbb{E}\left[e^{\beta\mathbf{x}^{i}}\right] &= \mathbb{E}\left[e^{\beta\left(\mathbf{x}^{i,1}+\cdots+\mathbf{x}^{i,C}\right)}\right] \\
    &= \mathbb{E}\left[e^{\beta\left(\mathbf{x}^{i,1}\right)} \cdots e^{\beta\left(\mathbf{x}^{i,C}\right)}\right]\\
    &= \mathbb{E}\left[e^{\beta\left(\mathbf{x}^{i,1}\right)}\right] \cdots \mathbb{E}\left[e^{\beta\left(\mathbf{x}^{i,C}\right)}\right]\\
    &= \mathbb{E}\left[e^{\beta\mathbf{x}^{1}}\right]^C.
    \end{split}
  \end{equation*}
  So that we can let their common moment generating function be written as $M_{\beta}=\mathbb{E}\left[e^{\beta\mathbf{x}^{i,1}}\right]=\mathbb{E}\left[e^{\beta\mathbf{x}^{i}}\right]^{\frac{1}{C}}$, and let $a^i$ of $\mathbf{x}^i$ be
  \begin{equation}
    \label{eq:a_x}
    a^{i} = \left(1+\frac{1}{C}\right)\mathbf{x}^{i}.
  \end{equation}
  Then Cram{\'e}r's theorem~\ref{thm:cramer} states that their empirical mean satisfies LDP with rate functional $I(a)$; \emph{i.e.}, by \eqref{eq:cramer}, we have
  \begin{equation}
    \label{eq:clique_cramer}
    \mathbb{P}(\rho_{\tau}^{i}\approx\rho^i) \approx \exp\left[-\max_{\beta\in\mathbb{R}^d}\left(\beta a^i - \frac{1}{C}\log\mathbb{E}\left[e^{\beta\mathbf{x}^{i}}\right]\right)\right].
  \end{equation}
  The empirical measure over all particles is simply the empirical mean of the Dirac measure, which can be written as
  \begin{align}
      \mathbb{P}(\rho_{\tau}\approx\rho\ \nonumber,  \rho_{\tau}^{1}\approx\rho^1,\ldots,\rho_{\tau}^{n}\approx\rho^n|\rho_0) &= \mathbb{P}(\rho_{\tau}\approx\rho|\rho_0)\prod_{i=1}^{n}\mathbb{P}(\rho_{\tau}^{i}\approx\rho^{i})\\ \nonumber 
      & \label{eq:final_empirical_subcramer} \approx \exp\left[-\min_{\mu\in \mathcal{T}(\rho,\rho_0)}D_{\text{KL}}(\mu||\rho_0)\right] \cdot \\ 
      & \quad\quad\quad\quad\quad \prod_{i=1}^{n}\exp\left[-\max_{\beta\in\mathbb{R}^d}\left(\beta a^i - \frac{1}{C}\log\mathbb{E}[e^{\beta\mathbf{x}^i}]\right)\right] \\\nonumber 
      & \label{eq:final_empirical_prodtosum} =\exp\left[-\min_{\mu\in \mathcal{T}(\rho,\rho_0)}D_{\text{KL}}(\mu||\rho_0) \right. \\
      &\left. \quad\quad\quad\quad\quad\quad - \sum_{i=1}^{n}\max_{\beta\in\mathbb{R}^d}\left(\beta a^i - \frac{1}{C}\mathbb{E}[\beta\mathbf{x}^{i}]\right) \right]\\\nonumber 
      & \label{eq:final_empirical_maxtorelu} =\exp\left[-\min_{\mu\in \mathcal{T}(\rho,\rho_0)}D_{\text{KL}}(\mu||\rho_0) \right. \\
      &\left. \quad\quad\quad\quad\quad\quad\quad\quad - \sum_{i=1}^{n}h(\beta(a^i - \frac{1}{C}\mathbf{x}^i))\right]\\\nonumber 
      & \label{eq:final_empirical_suba} =\exp\left[-\min_{\mu\in \mathcal{T}(\rho,\rho_0)}D_{\text{KL}}(\mu||\rho_0) \right. \\
      &\left. \quad\quad\quad\quad\quad\quad\quad\quad - \frac{\tau}{n}\sum_{i=1}^{n}h(W\mathbf{x}^i+b)\right]\\
      & \label{eq:final_empirical_meantoexpectation} \approx \exp\left[- \left(\min_{\mu\in \mathcal{T}(\rho,\rho_0)}D_{\text{KL}}(\mu||\rho_0)  
      + \tau\mathbb{E}_{\mathbf{x}\sim\rho}\left[\Phi_{\theta}(\mathbf{x})\right] \right)\right].
  \end{align}
  In the above derivation, \eqref{eq:final_empirical_subcramer} is derived by substituting \eqref{eq:our_sanov} and \eqref{eq:clique_cramer}, and \eqref{eq:final_empirical_prodtosum} is by the properties of exponents. In \eqref{eq:cramer}, the logarithm of probability must be less than or equal to $0$, so $I(a)$ must be greater than or equal to $0$, giving \eqref{eq:final_empirical_maxtorelu}: $\max\left(\beta a^i - \frac{1}{C}\mathbb{E}[\beta\mathbf{x}^{i}]\right) = \max\left(\beta a^i - \frac{1}{C}\mathbb{E}[\beta\mathbf{x}^{i}], 0\right) \stackrel{\eqref{eq:relu}}{=} h\left(\beta a^i - \frac{1}{C}\mathbb{E}[\beta\mathbf{x}^{i}]\right)$. \eqref{eq:final_empirical_suba} is given by substituting \eqref{eq:beta} and \eqref{eq:a_x}. The derived exponent \eqref{eq:final_empirical_meantoexpectation} is exactly the KL discrete flow $\mathcal{F}_{\text{KL}}$ in \eqref{eq:klproximal}, which concludes the proof that the empirical measure of all particles satisfies LDP with rate functional $\mathcal{F}_{\text{KL}}$ in discrete time.\\
  \end{proof}

\subsection*{Borel Product}
\label{append:product}
In fact, it is possible for computing the product of two measurable spaces. Referring to the Sec. 8.2  of~\citep{loeve2017probability} and the Chapter 5 of~\citep{axler2020measure}, if there are two measurable spaces $(X_1, \Sigma_1)$ and $(X_2, \Sigma_2)$ in consideration, where $\Sigma_1$ and $\Sigma_2$ are the $\sigma$-algebras on $X_1$ and $X_2$ respectively and let $q$ and $f$ be two measures on these spaces. The product of two Borel measures $q$ and $f$ is defined as 
\begin{equation*}
    fq(E)= \int_{X_2}  q(E^y)df(y)=\int_{X_1}f(E^x)dq(x),
\end{equation*}
 where $E$ is a set of $(x,y)$, $x\in X_1$, $y\in X_2$, $E^x =\{y\in X_2| (x,y)\in E\}$ and $E^y =\{x\in X_1| (x,y)\in E\}$. Clearly, in the original definition, if we let $p=fq$, $p$ cannot be in $\mathcal{P}$, since $p$ is the Borel measure over $E$, namely it is in the mode of $|\mathbb{R}^d|\times|\mathbb{R}^d|$. However, this is not an error, we will redefine the measure product in our settings.

 \noindent In our settings, let both $X_1$ and $X_2$ be  $\mathbb{R}^d$. The product definition is:
 \begin{equation*}
     p(\mathbf{x})=fq(\mathbf{x})= \int_{\mathbb{R}^d}  q(\mathbf{x})df(\mathbf{x})=\int_{\mathbb{R}^d}f(\mathbf{x})dq(\mathbf{x}).
 \end{equation*}
 Since two measurable spaces of the defined production are identical, we then merely consider $E'=\{(\mathbf{x}, \mathbf{x})|\forall \mathbf{x}\in\mathbb{R}^d\}$, the subspace of $E=\{(\mathbf{x}, \mathbf{y})| \forall \mathbf{x},\mathbf{y}\in\mathbb{R}^d\}$. With this method, $E, E^x, E^y$ in the original definition are one-by-one aligned and reduced to $\mathbb{R}^d$. And in this sense, $p=fq$  belongs to $\mathcal{P}$.
 
\subsection*{Proof of Fokker-Planck equation}
\label{append:fp_proof}

The conditions we have from Eq. ~\eqref{eq:continuity} to \eqref{eq:fp} are
\begin{equation*}
    \left\{
    \begin{aligned}
    &\partial_{t} \rho + \div(\rho\cdot\nu) =0 \quad                                  & \text{(Continuity condition)}                     \\
      &\nu=-\nabla\frac{\delta F}{\delta\rho} =-(\nabla\log\rho + \nabla\Phi) \quad & \text{(Variational condition)} 
    \end{aligned}
    \right.
\end{equation*}
\begin{proof}
Proving $\partial_{t} \rho = \div(\rho \nabla \Phi ) + \Delta \rho$ is equivalent to prove:
\begin{equation*}
    \begin{aligned}
    \Rightarrow &\quad\partial_{t} \rho - \div[\rho(\nabla\log\rho + \nabla\Phi)] = 0 \\
    \Rightarrow &\quad\div[\rho(\nabla\log\rho + \nabla\Phi)] = \div(\rho \nabla \Phi ) + \Delta \rho \\
     \Rightarrow &\quad\div(\rho\nabla\log\rho) = \Delta\rho.
    \end{aligned}
\end{equation*}
After we expand the LHS above, we have
\begin{equation*}
    \begin{aligned}
        \div(\rho\nabla\log\rho) &=  \div(\rho\cdot\frac{\partial\log\rho}{\partial\mathbf{x}}) = \div(\rho\cdot\frac{1}{\rho}\cdot\frac{\partial\rho}{\partial\mathbf{x}}) \\
        & =\frac{\partial\frac{\partial\rho}{\partial\mathbf{x}}}{\partial\mathbf{x}} + \frac{\partial\frac{\partial\rho}{\partial\mathbf{x}}}{\partial t} = \Delta\rho + \frac{\partial\frac{\partial\rho}{\partial\mathbf{x}}}{\partial t}.
    \end{aligned}
\end{equation*}
Since $\frac{\partial\rho}{\partial\mathbf{x}}$ is static when time interval $dt\rightarrow 0$, therefore the last term $\frac{\partial\frac{\partial\rho}{\partial\mathbf{x}}}{\partial t}$ above equals zero, and to this end we have done proving the LHS equals to the RHS.
\end{proof}
\subsection*{Deduction of Equation \eqref{eq:gammaconverge}}
\label{append:gamma}
With Eq. \eqref{eq:klproximal}, \eqref{eq:jkoproximal} and \eqref{eq:f}, we can prove this approximation Eq.  \eqref{eq:gammaconverge} by proving $\mathcal{F}_{\text{KL}}-\frac{1}{2} \mathcal{F}_{w^2}\rightarrow 0$ when $\tau\rightarrow0$.
\begin{proof}
\begin{equation*}
    \begin{aligned}
        \mathcal{F}_{\text{KL}}(\rho, \rho_t) &- \frac{1}{2}\mathcal{F}_{w^2}(\rho, \rho_t) =  \tau\min_{\rho\in\mathcal{T}(\rho_{t},\rho_{t+\tau})} \frac{1}{\tau}D_{\mathrm{KL}}(\rho\|\rho_t)
        -\frac{1}{4\tau}w{2}(\rho, \rho_t) 
        + \tau\mathbb{E}_{\rho_t}[\Phi] 
        - \frac{1}{2}\tau F(\rho_t) \\
        &=   \tau\min_{\rho\in\mathcal{T}(\rho_{t},\rho_{t+\tau})} \frac{1}{\tau}D_{\mathrm{KL}}(\rho\|\rho_t)
        -\frac{1}{4\tau}w^{2}(\rho, \rho_t) 
        + \tau\mathbb{E}_{\rho_t}[\Phi] 
        - \frac{1}{2}\tau[-\mathcal{H}(\rho_t)+\mathbb{E}_{\rho_t}[\Phi]] \\
        & = \tau\min_{\rho\in\mathcal{T}(\rho_{t},\rho_{t+\tau})} \frac{1}{\tau} D_{\mathrm{KL}}(\rho\|\rho_t)
        -\frac{1}{4\tau}w^{2}(\rho, \rho_t) 
        + \frac{1}{2}\tau\mathbb{E}_{\rho_t}[\Phi]
        + \frac{1}{2}\tau\mathcal{H}(\rho_t)\\
        &\xrightarrow{\text{Theorem \ref{thm:klkjolink}}} \frac{1}{2}\tau F(\rho_{t+\tau}) 
        - \frac{1}{2}\tau F(\rho_t)
        + \frac{1}{2}\tau\mathbb{E}_{\rho_t}[\Phi]
        + \frac{1}{2}\tau\mathcal{H}(\rho_t)\xrightarrow{\tau\rightarrow 0}0
    \end{aligned}
\end{equation*}
\end{proof}
\section*{Appendix B. Details of OT-embeded CoopNet}
\label{append:coopnet}
\begin{figure}[t]
  \centering
  \centerline{\includegraphics[width=\linewidth]{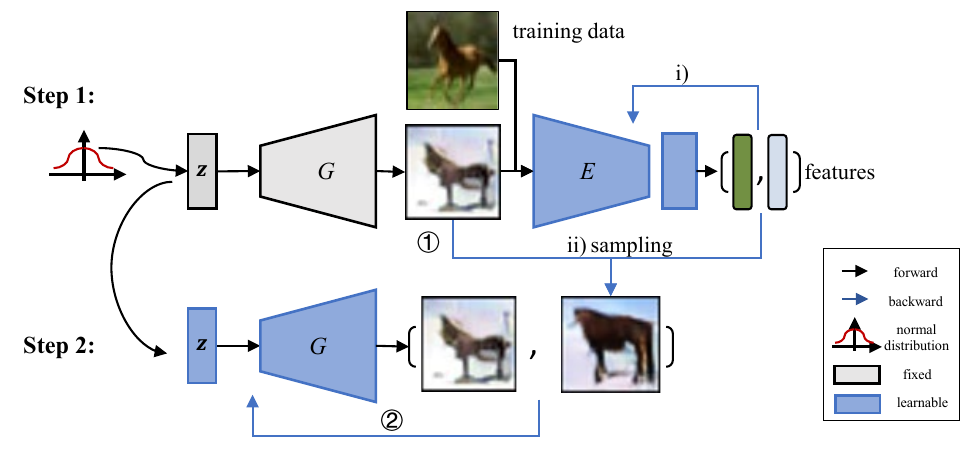}}
  \caption{Illustration of cooperative learning during each iteration step. The blue colored part of each step is the under-training model parameters instantly. In the first step, the EBM will be learned by applying SGLD to minimize the KL flow (or using the dynamic JKO discrete flow, the OT-embeded method) of the generator represented distribution and the training data distribution. Meanwhile, the synthesized new sample (by updating the generated sample) will be applied to the second step, to estimate the generator and the latent space vector $\mathbf{z}$.\label{pipeline}}
\end{figure}

Herein we briefly present the model that combines EBM and GAN: the cooperative learning neural network~(CoopNet)~\citep{xie2018cooperative} and its OT-embeded form. CoopNet treats GAN's generator $g_{w}:\mathbb{R}^k \rightarrow \mathbb{R}^d$ as a latent variable model which seeks to explain $d$-dimensional samples $\{\mathbf{x}^i\}_{i=1}^{n}$ synthesized from EBM via a set of $k$-dimensional random vectors $\{\mathbf{z}^i\}_{i=1}^{n}$ (typically $k\ll d$). Parameter $w$ is learned by solving the reconstruction loss
\begin{equation}
  \label{eq:coopnetg}
  \min_{w}\frac{1}{n}\sum_{i=1}^{n}\min_{\mathbf{z}^i}\left\|g_w(\mathbf{z}^i) - \mathbf{x}^i\right\|^2.
\end{equation}

Regarding to the equation \eqref{eq:coopnetg}, after initializing $\mathbf{x}$, $\mathbf{z}$, $\theta$ and $w$, the learning algorithm of CoopNet during each iteration can be summarized in two stages:
\begin{enumerate}[label=\protect\circled{\arabic*}]
  \item For a batch of $\mathbf{z}$, compute the $g_w(\mathbf{z})$, and take it as an intermediate state of EBM or OT-EBM, update $\theta$ and $\mathbf{x}$ via \eqref{eq:standard_learning} or the OT learning cross-updating equation \eqref{eq:new_learning}, which depends on whether we use EBM or OT-EBM.
  Note that the cross-updating process is an alternate process that includes i) the EBM or OT-EBM $E$ being parameter-estimated by approximating the generated sample with the real data sample; and ii) sampling a new one as the learning target of the generator.
  \item Update $\mathbf{z}$ and $w$ via the gradients of \eqref{eq:coopnetg} \emph{w.r.t} $\mathbf{z}$ and $w$ respectively. Before the $\mathbf{z}$ sampling process, $\mathbf{z}$ is initialized from the last step.
\end{enumerate}
\begin{figure}
     \centerline{\includegraphics[width=0.8\linewidth]{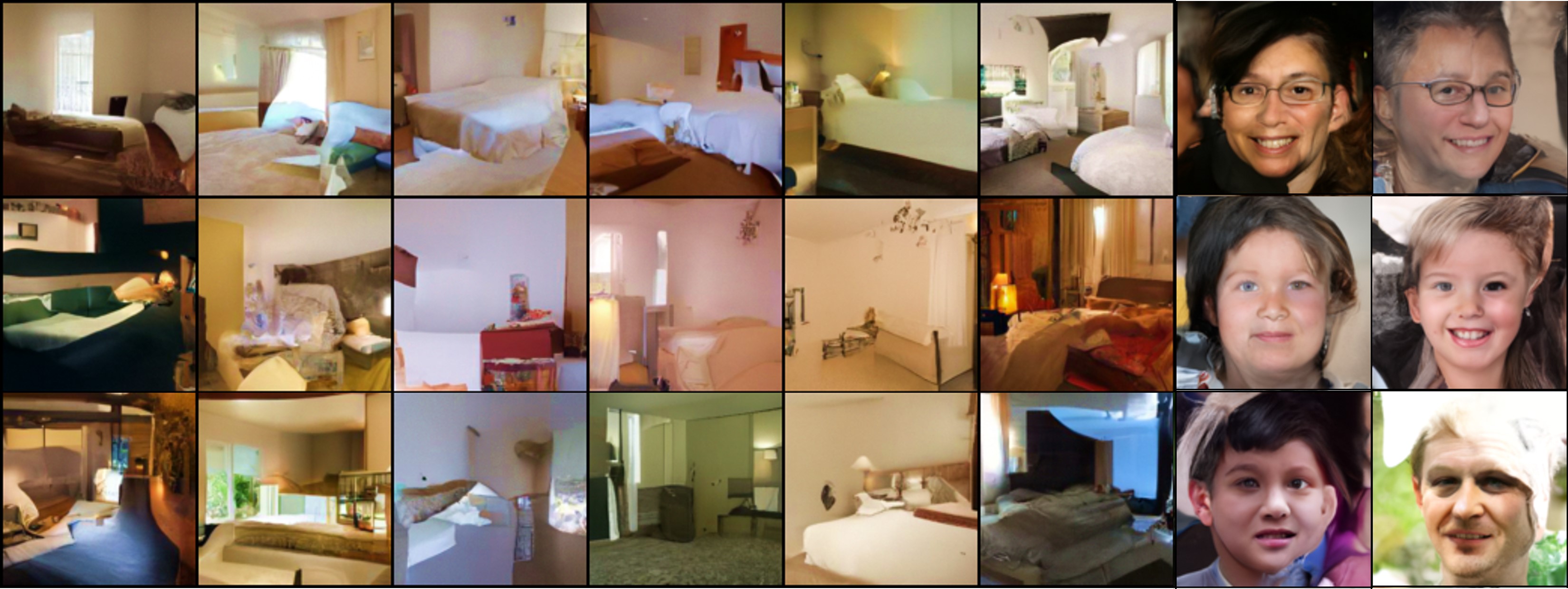}}
    \caption{High resolution generated results as complementary}
    \label{fig:high_resolution}
\end{figure}

\section*{Appendix C. Further Discussion with Weight Normalization}
The weight normalization~\citep{salimans2016weight} in neural networks usually optimizes the learning gradient during the training. Namely, those different normalization approaches, such as weight clipping and spectral normalization, factually share identical learning goal with the proposed OT method. To explore the relationships with OT, we try to add up normalization layers to the EBM network. However, the fine-tuning process seems difficult to converge even with every attempts. The cause behind this might be: (1) layers of weight normalization result in complicated function $\Phi_{\theta}$ thus a high expense on gradient computation; (2) the distributed gradients are disturbed and will introduce bias accumulation in SGLD; (3) theoretically, simultaneously applying weight normalization with OT is a repetition, especially the spectral normalization, which also aims at normalizing the energy function (or the discriminator in GAN) to Lipschitz -continuous. But the weight normalization is so far inapplicable to EBM network, which might due to the SGLD and needs further exploration.


\bibliography{otebm}

\end{document}